\documentclass[10pt,twocolumn,letterpaper]{article}

\usepackage{cvpr}              

\usepackage{amsmath}
\usepackage{amssymb}
\usepackage{booktabs}
\usepackage{times}
\usepackage{epsfig}
\usepackage{array,multirow, booktabs, verbatim}
\usepackage{dblfloatfix, url}
\usepackage{graphicx} 
\usepackage[pagebackref,breaklinks,colorlinks]{hyperref}
\def\eqnvspace{{\vspace{-0mm}}}
\def\figvspaceB{{\vspace{-4.0mm}}}
\def\figvspace{{\vspace{-5.0mm}}}
\def\tabvspace{{\vspace{-8mm}}}
\newcommand{\Paragraph}[1]{\vspace{-0mm} \noindent \textbf{#1} \hspace{0mm}}
\newcommand{\Section}[1]{\vspace{-1.0mm} \section{#1} \vspace{-1.5mm}}
\newcommand{\SubSection}[1]{\vspace{-1.0mm} \subsection{#1} \vspace{-1mm}}

\usepackage[capitalize]{cleveref}
\crefname{section}{Sec.}{Secs.}
\Crefname{section}{Section}{Sections}
\Crefname{table}{Table}{Tables}
\crefname{table}{Tab.}{Tabs.}

\def\cvprPaperID{6513} 
\def\confName{CVPR}
\def\confYear{2022}

\begin{document}

\title{Robust Egocentric Photo-realistic Facial Expression Transfer for Virtual Reality}

\author{Amin Jourabloo\textsuperscript{$\dagger$} \ \ \ Baris Gecer\textsuperscript{$\ddag$} \ \ \ Fernando De la Torre\textsuperscript{$\S$}\textsuperscript{$\dagger$} \ \ \ Jason Saragih\textsuperscript{$\dagger$} \ \ \ \ Shih-En Wei\textsuperscript{$\dagger$} \\ Stephen Lombardi\textsuperscript{$\dagger$}\ \ \ \ Te-Li Wang\textsuperscript{$\dagger$} \ \ \ \ Danielle Belko\textsuperscript{$\dagger$} \ \ \ \ Autumn Trimble\textsuperscript{$\dagger$} \ \ \ \ Hernan Badino\textsuperscript{$\dagger$}\\
\textsuperscript{$\dagger$}Facebook Reality Labs, Pittsburgh, PA \\
\textsuperscript{$\ddag$}Imperial College London\\
\textsuperscript{$\S$}Robotics Institute, Carnegie Mellon University
}


\vspace{-0.2in}
\maketitle
\vspace{-0.2in}
\thispagestyle{empty}
\vspace{-0.2in}

\begin{abstract}
    Social presence, the feeling of being there with a “real” person, will fuel the  next generation of communication systems driven by digital humans in virtual reality (VR). The best 3D video-realistic VR avatars that minimize the uncanny effect rely on person-specific (PS) models. However, these PS models are time-consuming to build and are typically trained with limited data variability, which results in poor generalization and robustness.  Major sources of variability that affects the accuracy of facial expression transfer algorithms include using different VR headsets (e.g., camera configuration, slop of the headset), facial appearance changes over time (e.g., beard, make-up), and environmental factors (e.g., lighting, backgrounds). This is a major drawback for the scalability of these models in VR.

This paper makes progress in overcoming these limitations by proposing an end-to-end multi-identity architecture (MIA) trained with specialized augmentation strategies. MIA drives the shape component of the avatar from three cameras in the VR headset (two eyes, one mouth), in untrained subjects, using minimal personalized information (i.e., neutral 3D mesh shape).  Similarly, if the PS texture decoder is available, MIA is able to drive the full avatar (shape+texture) robustly outperforming PS models in challenging scenarios. {\bf Our key contribution to improve robustness and generalization, is that our method implicitly decouples, in an unsupervised manner, the facial expression from nuisance factors (e.g., headset, environment, facial appearance)}. We demonstrate the superior performance and robustness of the proposed method versus state-of-the-art PS approaches in a variety of experiments. 
\vspace{-2mm}

\end{abstract}

\Section{Introduction}


Our experience with communication systems is two-dimensional,  mostly via  video teleconferencing (e.g., Messenger), that includes both audio and video transmissions.  Recent studies on videoconferencing have shown that the more closely technology can simulate a face-to-face interaction, the more participants are able to focus, engage, and retain information~\cite{Bolle09}. A more advanced level of communication with virtual reality (VR) via telepresence~\cite{ orts2016holoportation, tewari2019fml, thies2016face2face, elgharib2019egoface, nagano2018pagan, hu2021egorenderer, chandran2020semantic, chen2021high} will allow virtual presence at a distant location and a more authentic interaction.  If successful, this new form of face-to-face interaction can reduce the time and financial commitments of travel, make sales meetings or family meetings more immersive, with a huge impact for the environment and use of personal time.  


\begin{figure}[t]
\begin{center}
    \includegraphics[width=\linewidth]{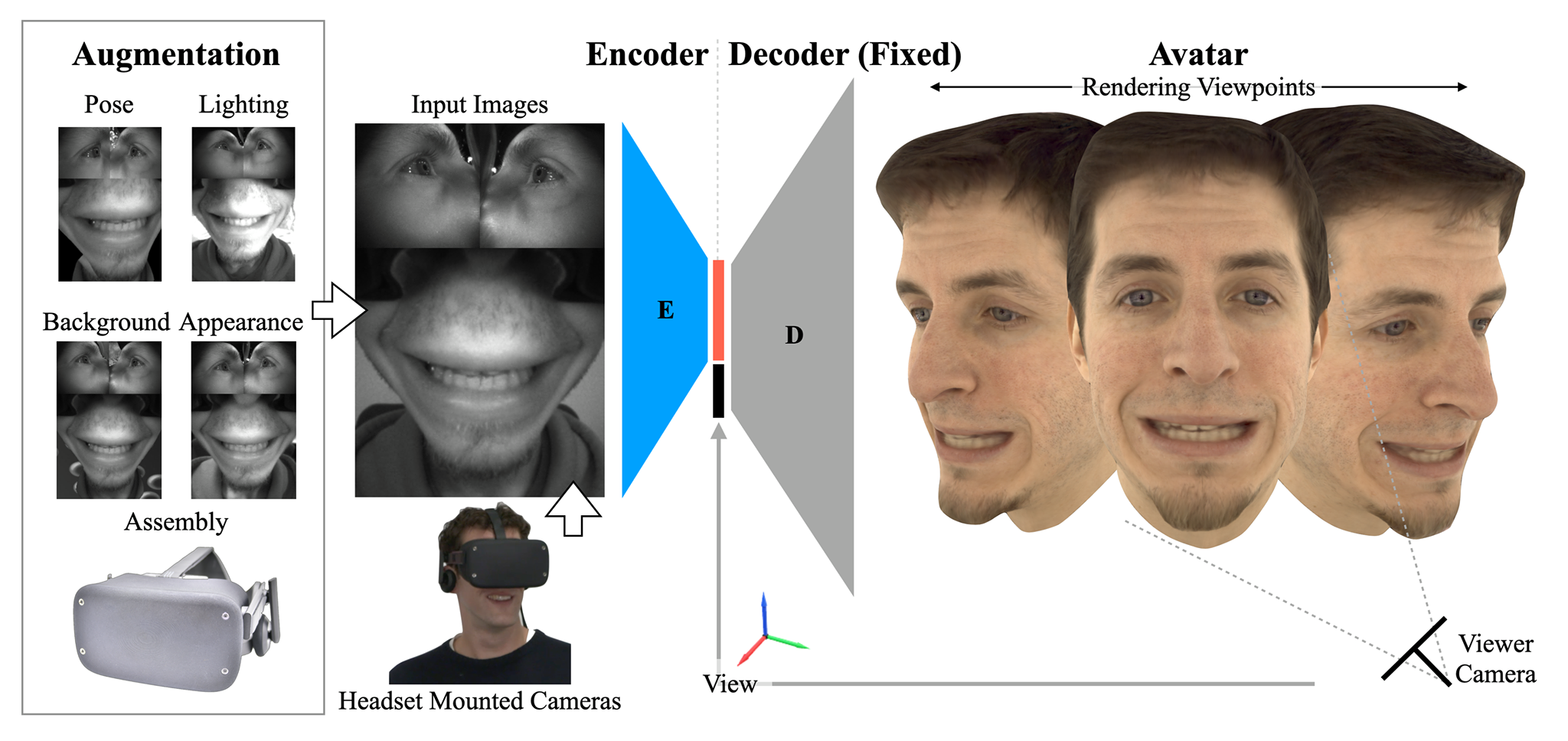}
\end{center}
\figvspaceB
   \caption{3D Photo-realistic avatar driven by three headset-mounted camera (HMC) images in a VR headset. This paper presents a system to drive photo-realistic avatars robustly with variability in headsets, lighting,  environmental background, head pose, and facial appearance. }
\label{fig:intro_fig1}
\figvspace
\end{figure}

Today most real-time systems for avatars in AR/VR are cartoon-like (e.g., Hyprsense, Loom AI); on the other hand, Hollywood has animated nearly uncanny digital humans as virtual avatars using advanced computer graphics technology and person-specific models (e.g., Siren). While some of these avatars can be driven in real-time from cameras, building the PS model is an extremely time-consuming and hand-tuned process that prevents democratization of this technology. 
This paper makes progress in this direction by generating video-realistic avatars by transferring subtle facial expressions from the headset mounted cameras (HMC) images in a VR headset to a 3D talking head (see Fig.~\ref{fig:intro_fig1}). 


\begin{figure*}[t]
\begin{center}
\setlength\tabcolsep{1.25pt}
\scriptsize
\begin{tabular}{ c c c c c c}
 \includegraphics[width=18mm]{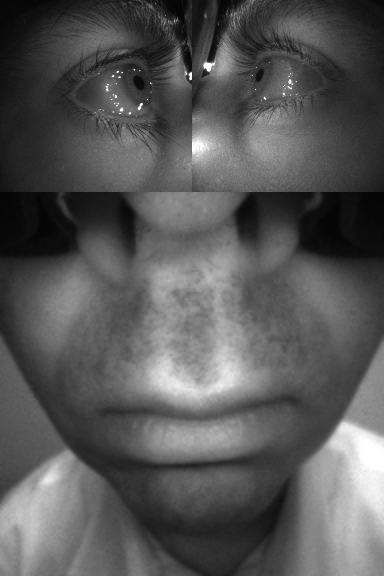} & \includegraphics[width=18mm]{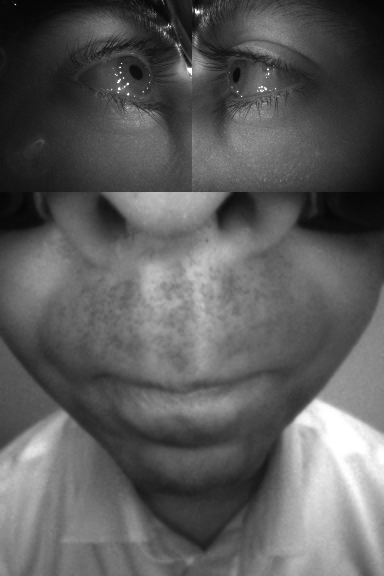} & \includegraphics[width=18mm]{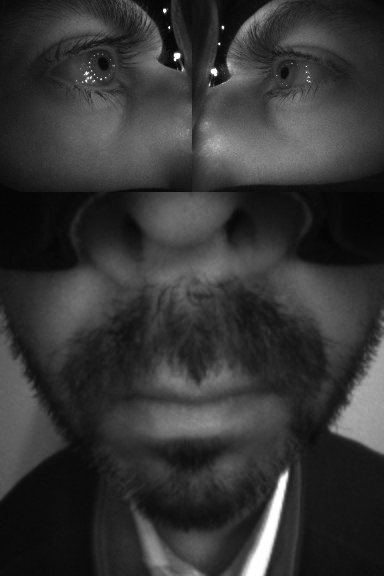} &
 \includegraphics[width=18mm]{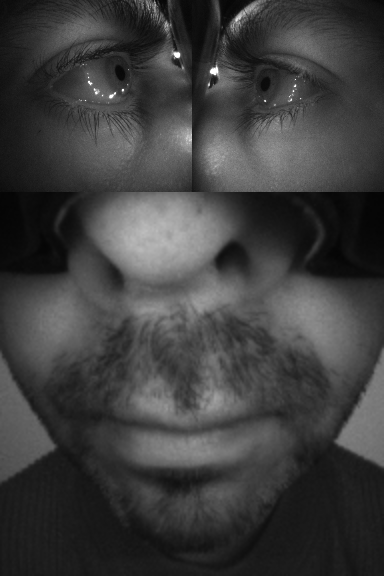} &
 \includegraphics[width=18mm]{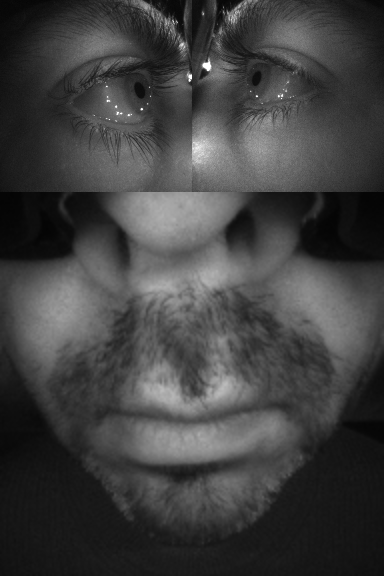} &
 \includegraphics[width=35mm]{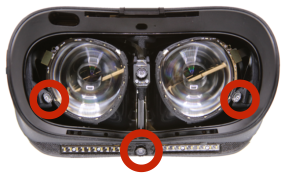} \\
 (a) & (b) & (c) & (d) & (e) & (f) \\
 Headset & \textbf{\textcolor{blue}{Headset}} & \textbf{\textcolor{blue}{Headset}} & \textbf{\textcolor{blue}{Headset}} & Headset &\\
 Environment & Environment & Environment & Environment & \textbf{\textcolor{blue}{Environment}} &\\
 Facial appearance & Facial appearance & \textbf{\textcolor{blue}{Facial appearance}} & \textbf{\textcolor{blue}{Facial appearance}} & \textbf{\textcolor{blue}{Facial appearance}} & 
 
\end{tabular}
\end{center}
\figvspaceB
\caption{Comparing the HMC images of a subject in multiple HMC captures with variations in headset, environment and facial appearance. (a) The training HMC capture, (b-e) the testing HMC captures. The \textcolor{blue}{blue} \textbf{bold} font shows the variations respect to the training capture (a). The red circles in (f) show the locations of the cameras inside the headset.}
\label{fig:test_samples}
\figvspace
\end{figure*}

\begin{figure*}[b!]
\begin{center}
\includegraphics[width=140mm]{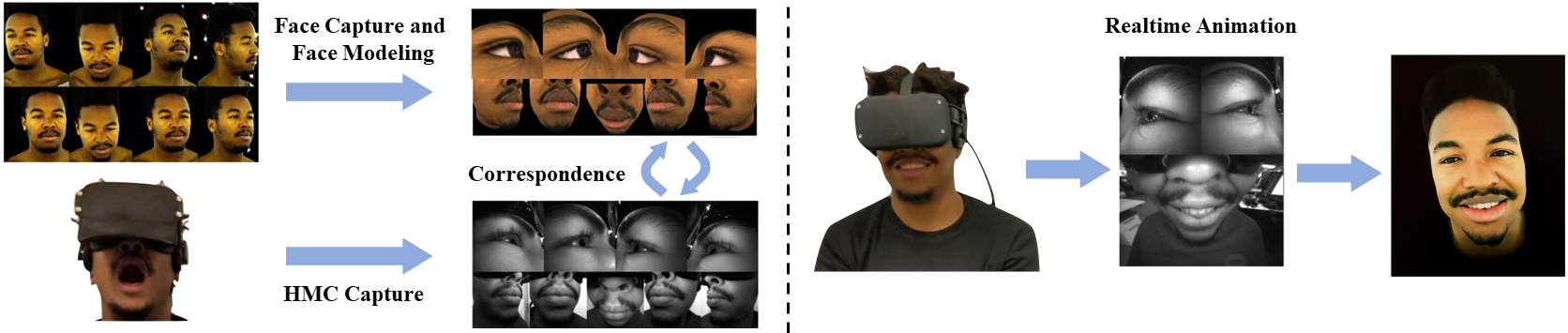}
\end{center}
\figvspaceB
\caption{Training and testing pipeline for animating the face codec avatar. In the data collection stage, we perform a face capture to generate the codec avatar of the subject~\cite{lombardi2018deep} and a HMC capture. We utilize~\cite{wei2019vr} to find the correspondence between the avatar and the HMC capture. Finally, we can train a model to animate the codec avatar (CA) from the HMC images in real-time.}
\label{fig:pipline}
\figvspace
\end{figure*}

We build on recent work on codec avatars (CA)~\cite{lombardi2018deep} that learn a PS model from a Plenoptic study. Recall that driving an avatar from HMC cameras is typically more challenging than driving it from regular cameras (e.g., iPhone)~\cite{roth2016adaptive, lin2020towards, zhu2020reda}, due to the domain difference between IR cameras and the texture/shape of the avatar, variability in HMC images due to headset variability (e.g., camera location, IR LED illumination), high-distortion introduced by the near-camera views, and partial visibility of the face (Fig.~\ref{fig:test_samples}). Wei \etal~\cite{wei2019vr} proposed an end-to-end deep learning network for learning the mapping between the HMC images and the parameterized avatar. First, this model solves the unknown correspondence between HMC images and the avatar parameters in an unsupervised manner using an eleven-view HMC headset. Second, to animate the CA in real time from three HMC images(i.e., inference), Wei \etal~\cite{wei2019vr} learns an encoder network to regress from 3-view HMC images to CA's parameters (Fig.~\ref{fig:pipline}).

While previous work has reported compelling photo-realistic facial expression transfer results, the existing method has limitations due to the PS nature of the approach. It is time-consuming, expensive and error-prone to capture sufficient statistical variability when collecting PS samples to learn a robust model. It will typically require recording several sessions with variability across lighting, headsets and iconic changes (e.g., makeup, beard), which limits its scalability. To build generic models (conditioned to the neutral shape), {\bf the most import contribution of this paper is to propose multi-identity architecture (MIA), an architecture that factorizes nuisance parameters such as camera parameters, facial aesthetic changes (e.g., beard, makeup) and environmental factors (e.g., lighting) from the facial motion (i.e., facial expression)}. This is critical because the encoder is able to extract from the HMC images {\em only} the information that is relevant to the final task, which is transferring subtle facial expressions, and it is able to marginalize information that is not relevant (headset, facial appearance, environment). Implicitly, this results in an algorithm that aligns facial expressions (3D shape + texture) {\bf across users in an unsupervised manner}. Recall that is a very  difficult  problem  to  align  subtle  facial behavior (using both 3D shape + texture) across users in a supervised or unsupervised manner.   That is,  how can we  find  the  correspondence  of  expression  across  subjects? Even if done manually, it is an extremely challenging problem and MIA (to the best of our knowledge) is the first algorithm that solves this problem in an unsupervised and discriminative manner (see subsection 4.3). MIA results in an algorithm for facial expression transfer for VR, that improves upon PS models in realistic scenarios.




\Section{Prior Work}


\begin{figure*}[t]
\small
\begin{center}
\includegraphics[width=120mm]{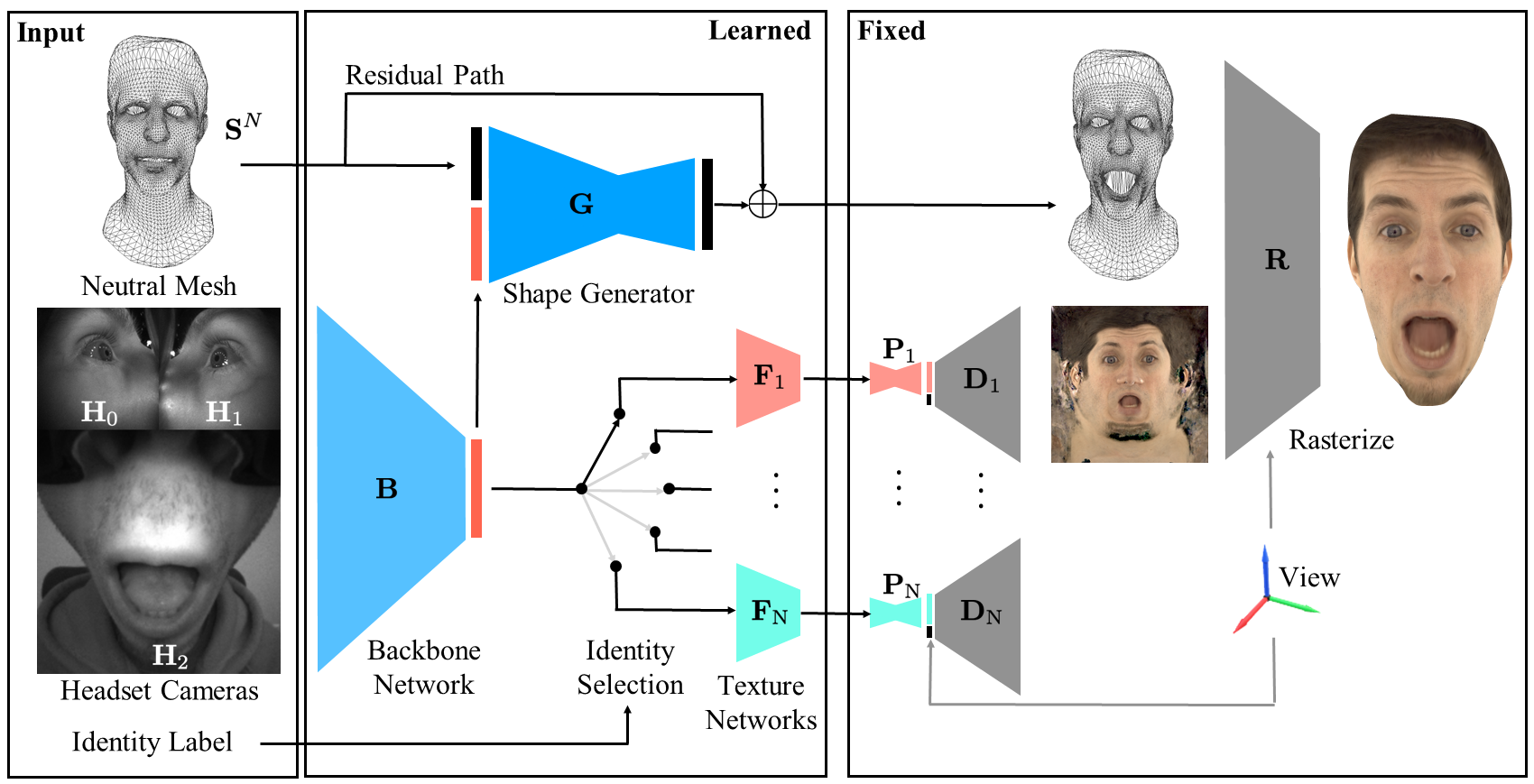}
\end{center}
\figvspaceB
   \caption{The proposed multi-identity architecture (MIA). It consists of three main parts: the backbone network $\textbf{B}$, the $3$D shape network $\textbf{G}$, and the texture networks $\textbf{F}_i$. The identity selector module pass the features to the corresponding texture network.}
\label{fig:intro_fig2}
\figvspace
\end{figure*}

\SubSection{Animating Stylized and Codec Avatars} 

Animating stylized avatars from video has a long history,  for instance~\cite{chaudhuri2019joint} fits a generic $3$DMM to the face and use it to retarget the facial motion to a $3$D characters. To improve the accuracy,  Chaudhuri \etal~\cite{chaudhuri2020personalized} proposed to learn person-specific expression blendshapes and dynamic albedo maps from the input video of subjects. In~\cite{song2020unsupervised}, facial action unit intensity is estimated in a self-supervise manner by utilizing a differentiable rendering layer for fitting the expression and to retarget the expression to the character. In contrast, expression transfer from a VR headset~\cite{olszewski2016high, elgharib2019egoface, hickson2019eyemotion, lou2019realistic} is more challenging due to partial visibility of face in HMC images, the specific hardware, and limited existing data.

CAs animate avatars by estimating the parameters of a PS shape and texture model from HMC~\cite{lombardi2018deep, wei2019vr, chu2020expressive, schwartz2020eyes}, see Fig.~\ref{fig:pipline}. In~\cite{lombardi2018deep}, combination of real and synthetic HMC images are utilized for reducing the domain gap between real HMC images in IR spectrum and rendered images for training encoder and reducing the HMC-avatar domain gap. Wei \etal~\cite{wei2019vr} utilize a cycle-GAN to achieve accurate cycle consistency between 11-view HMC images and CA. Then, they train a person specific regressor from 3-view HMC images to the CA's parameter. Chu \etal~\cite{chu2020expressive} propose to use modular CA to have more freedom for animating the eyes and mouth.
In a different approach, Richard \etal~\cite{richard2020audio} animate the CA based on the gaze direction and audio inputs. The aforementioned methods rely on PS models, are typically not robust to variations in headsets and environments.

\SubSection{3D Shape Estimation} 
Early approaches for model-based shape and texture estimation are based on active shape model~\cite{cootes1994active} (ASM) and Active Appearance Model~\cite{cootes1998active, matthews2004active} (AAM). The AAM methods learn a joint holistic model of shape and appearance. $3$D Morphable Model ($3$DMM) provides a dense $3$D representation for faces \eg the Basel Face Model~\cite{paysan20093d} and the FaceWarehouse~\cite{cao2013facewarehouse}. In~\cite{jourabloo2017poseICCV, guo2020towards}, $3$DMM is incorporated in an end-to-end CNN training to dicriminatively estimate the $3$D shape of faces given single input image. Tran \etal~\cite{tran2019learning} propose to learn a nonlinear $3$DMM via deep neural network from in-the-wild images, and in this way $3$DMMs are capable of representing non-linear facial expressions.  
The proposed method in~\cite{feng2021learning} can extract expression-dependent details of the 3D shape from a single image.~\cite{gecer2019ganfit} proposed to use the GAN generator for 3DMM fitting and estimating high-fidelity UV texture. Similarly,~\cite{jackson2017large} proposed to utilize the volumetric representation of face instead of using 3DMM. An unsupervised method proposed in~\cite{genova2018unsupervised} for, identity 3DMM fitting, regressing the 3D shape and texture. Also, in~\cite{sanyal2019learning} the identity constraints utilized among the images of the same subject.
Similar to~\cite{tran2019learning} we learn a non-linear discriminative $3$DMM, but we extend it to learn the model from HMC images given a neutral $3$D shape, and align the expressions across subjects in an unsupervised manner.  To the best of our knowledge, this is the first work that solves the correspondence of expression across subjects in unsupervised manner.

\vspace{-2mm}
\Section{Multi-Identity Model}

This section describes the proposed multi-identity architecture (MIA) and augmentation techniques to robustify and generalize existing encoder models for driving CAs.  

\SubSection{Multi-Identity Architecture (MIA)}

Given 3-view HMC images of the eyes and mouth (see Fig.~\ref{fig:intro_fig1}), our goal is to estimate the facial expression of a CA (shape+texture), and render it in an arbitrary view in VR. The MIA  has three main parts (see Fig.~\ref{fig:intro_fig2}):  the backbone network, the $3$D shape network, and the texture branch.  

\Paragraph{Backbone network:} The backbone network, $\textbf{B}_\psi$ in Fig.~\ref{fig:intro_fig2},  is shared among subjects. Its goal is to factorize the expression from other nuisance factors such as lighting, background, or camera views, and build an internal representation that is invariant to those factors. As we will show in the experiments section, MIA naturally finds that the best way to encode HMC images across subjects, is by marginalizing out person-specific factors in addition to the nuisance factors mentioned. This results in learning an
embedding that only preserves expression without the need of solving for correspondence across expression among subjects.

\Paragraph{$3$D shape network:}  
MIA assumes that the neutral shape of the test subject,  $\textbf{S}^N  \in \mathbb{R}^{7306\times3}$, in given\footnote{Extracting neutral face from a single or few-shot phone-captured images is a well studied problem~\cite{roth2016adaptive, lin2020towards, zhu2020reda}, and there are a number of commercial solutions available~\cite{FacePlusPlus, KeenTools}. }. This is the only information MIA needs to generalize the shape component of the network to untrained subjects. A  network  $\textbf{G}_\gamma$ is trained to estimate $3$D shape, $\hat{\textbf{S}} \in \mathbb{R}^{7306\times3}$ from HMC images. The network $\textbf{G}_\gamma$ takes both of the output of the backbone network $\mathbf{B}_{\psi}$ and $\textbf{S}^N$ to estimate the person specific $3$D shape expression residual. The neutral $3$D shape is used to re-inject person-specific information that was factored out in $\mathbf{B}_{\psi}$. For instance, eye openness, which varies across identities, can be extracted from the neutral $3$D shape $\textbf{S}^N$ of each subject. With this, we reconstruct the $3$D shape of subject $i$ as: 
\begin{equation}
\hat{\textbf{S}_i}=\textbf{S}_i^N + \textbf{G}_\gamma (\textbf{B}_\psi(\textbf{H}^0_i, \textbf{H}^1_i, \textbf{H}^2_i), \textbf{S}_i^N).
\eqnvspace
\end{equation}
The network $\textbf{G}_\gamma$ is trained by minimizing the Euclidean distance between the target $\textbf{S}_i$ and estimated $\hat{\textbf{S}_i}$ $3$D shapes, 
\begin{equation}
\textbf{L}_S^i=\|\textbf{W}^S \odot (\textbf{S}_i -  \hat{\textbf{S}_i}) \|_2^2,
\label{equ:loss_shape}
\eqnvspace
\end{equation}
where $\textbf{W}^S$ is the weight mask for the visible areas.

\begin{figure}[t]
\begin{center}
\setlength\tabcolsep{1.05pt}
\begin{tabular}{ c c c }
 \includegraphics[width=18mm]{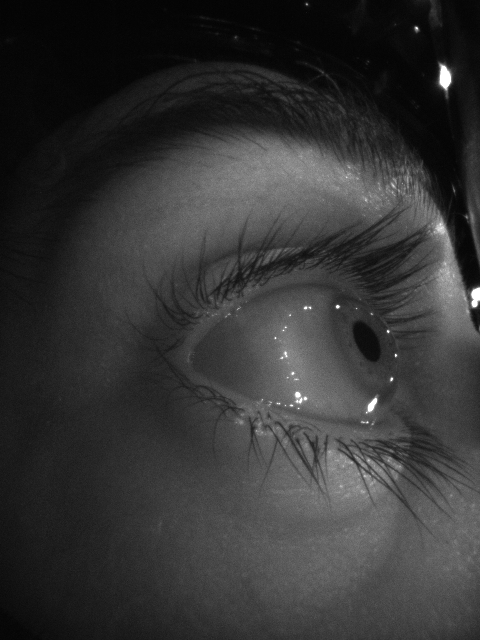} & \includegraphics[width=18mm]{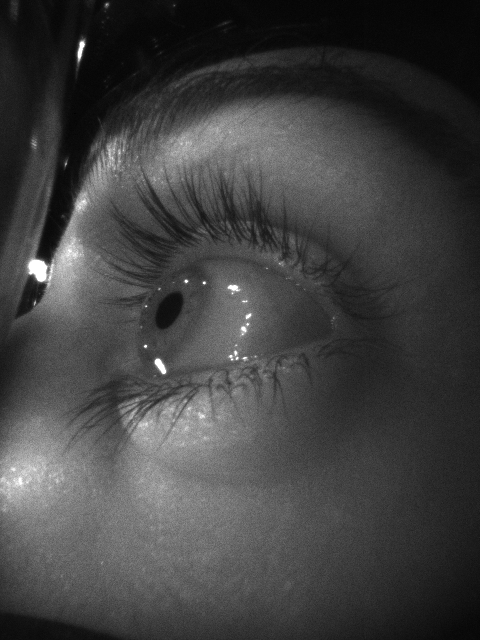} & \includegraphics[width=32mm]{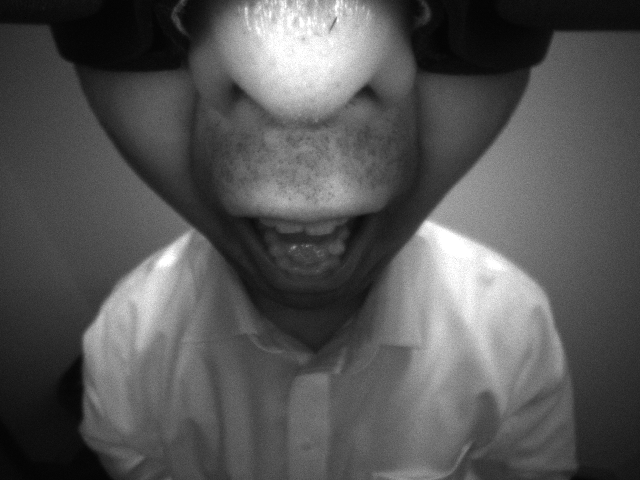}  \\ 
 \includegraphics[width=18mm]{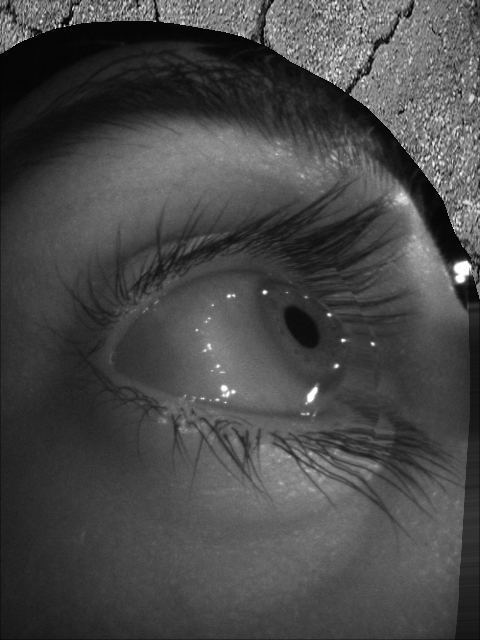} & \includegraphics[width=18mm]{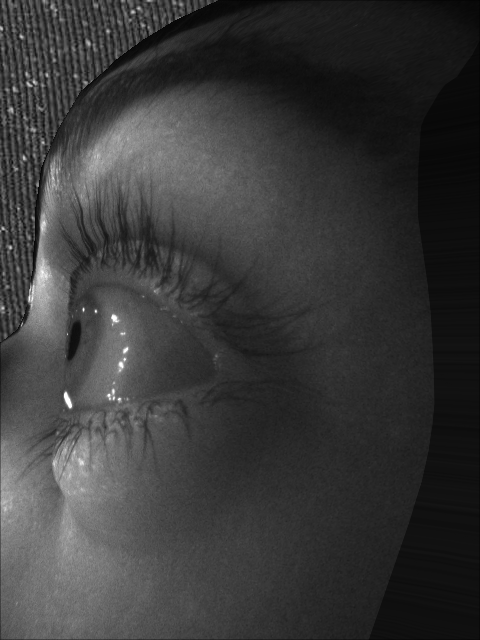} & \includegraphics[width=32mm]{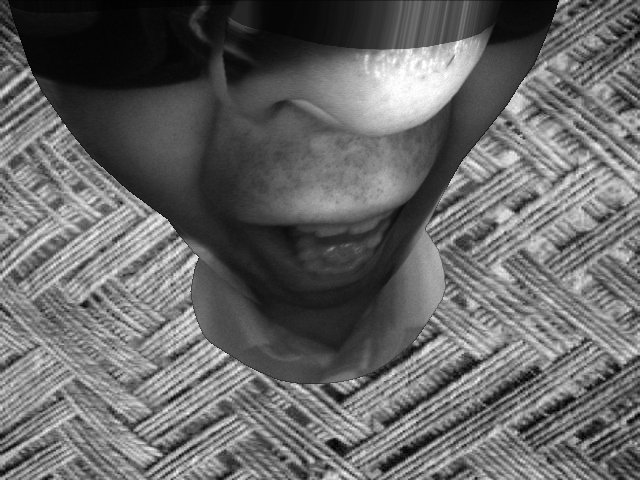}     
\end{tabular}
\end{center}
\figvspaceB
\caption{Examples of applying $3$D augmentation layer to HMC images. First row: Real HMC images, second row: Augmented images by changing $3$D pose, focal length and background. }
\label{fig:augmentation}
\figvspace
\end{figure}

\Paragraph{Texture network:} When the pre-trained PS texture decoder is available for each identity, our goal is to be able to animate the CA from HMC images robustly and with minimal adaptation effort. In this paper, we presume pre-trained decoders $\textbf{D}_\phi$ from~\cite{lombardi2018deep} are available, but our work can be similarly applied to other PS models as well (e.g.~\cite{tran2019towards,lee2020uncertainty}). The network $\textbf{D}_\phi$ takes, as input, an expression parameter $\textbf{z} \in \mathbb{R}^{256}$ and a view vector $\textbf{v} \in \mathbb{R}^{3}$, and generates person-specific and view-specific texture $\textbf{T}^\textbf{v} \in \mathbb{R}^{1024\times1024\times3}$ that, together with shape, can be used to render the avatar,
\begin{equation}
\textbf{T}^\textbf{v}  =\textbf{D}_\phi(\textbf{z},\textbf{v}). 
\label{equ:equ_decoder}
\end{equation}

However, since each PS model is trained independently of all others, the structure of the latent space, $\mathbf{z}$, is not consistent across identities. We would like to utilize the shared backbone encoder $\mathbf{B}_{\psi}$ across identities to encourage robustness via joint training. Inspired by multi-task learning techniques~\cite{naruniec2020high,cao2018partially}, we additionally learn person-specific adaptation layers, $\mathbf{F}_{\theta}$, that transform the identity-consistent expression embedding produced by $\mathbf{B}_{\psi}$ to each identity's personalized latent space. Finally, to eliminate unnecessary dimensions in $\mathbf{z}$, non-informative dimensions, we apply PCA dimensionality reduction, denoted $\mathbf{P} \in \mathbb{R}^{256\times 80}$ to each identity's latent space and fix it during training. Together, these components are used to generate PS expression parameters as follows:
\begin{equation}
\hat{\textbf{z}}_i=\textbf{P}_i(\textbf{F}_{\theta_{i}} (\textbf{B}_\psi(\textbf{H}^0_i, \textbf{H}^1_i, \textbf{H}^2_i)))+\overline{\textbf{z}}_i,
\eqnvspace
\end{equation}
where $i$ is subject index, and $\overline{\textbf{z}}_i$ is the average expression parameter for subject $i$. Then, we use Eqn.~\ref{equ:equ_decoder} to generate the estimated texture $\textbf{T}_i^\textbf{v}$ from view $\textbf{v}$. To guide the network, we minimize the Euclidean loss between the estimated and the target expression parameters and textures:   
\begin{equation}
\textbf{L}_T^i=\|\textbf{z}_i - \hat{\textbf{z}_i} \|_2^2 + \lambda_\textbf{T} \|\textbf{W}^T \odot (\textbf{T}_i^\textbf{v} -  \hat{\textbf{T}_i^\textbf{v}} ) \|_2^2,
\eqnvspace
\end{equation}
where $\textbf{W}^T$ is the weight mask for the visible areas from the HMC images and $\lambda_T$ is the weight for the texture loss.

\begin{figure*}[t]
\begin{center}
\setlength\tabcolsep{0.2pt}
\small
\begin{tabular}{ccccccccccc}
 \includegraphics[width=15.32mm]{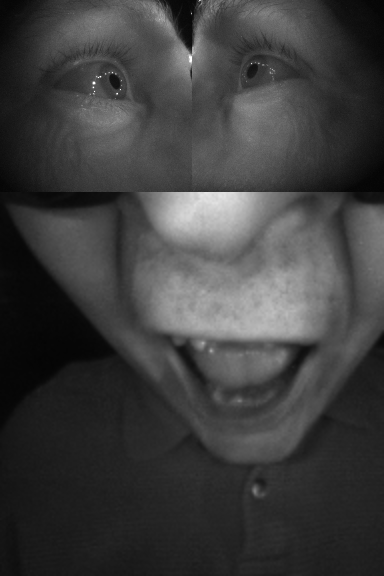} & 
 \includegraphics[trim=90 100 100 70,clip, width=14mm]{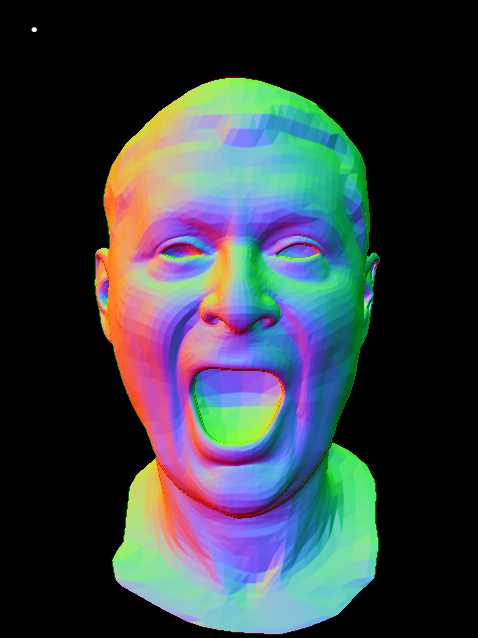} &
 \includegraphics[trim=90 100 100 70,clip, width=14mm]{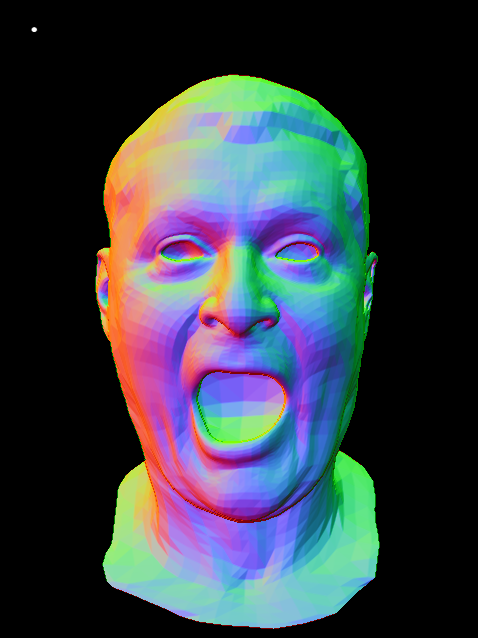} & \ \ &
 
 \includegraphics[width=15.32mm]{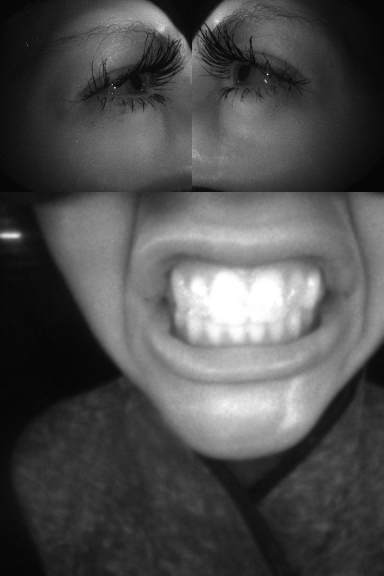} & 
 \includegraphics[trim=90 100 100 70,clip, width=14mm]{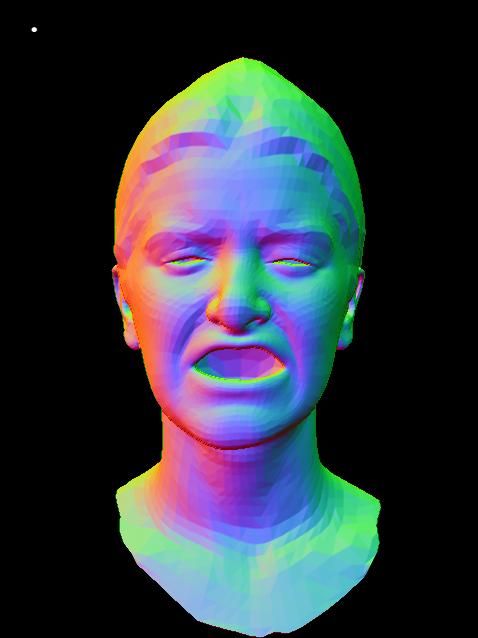} &
 \includegraphics[trim=90 100 100 70,clip, width=14mm]{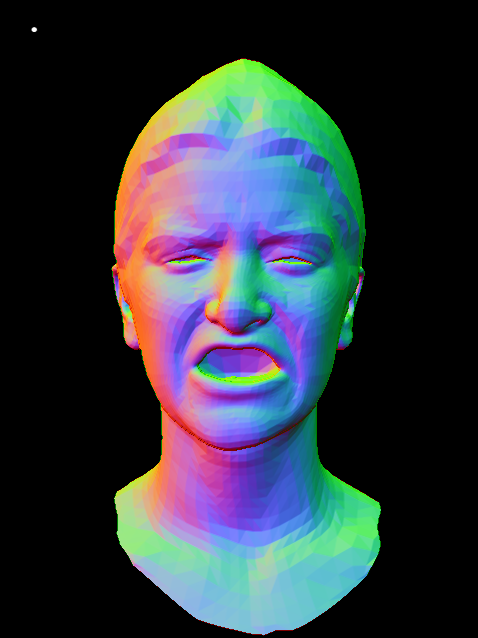} & \ \ &
 
 \includegraphics[width=15.32mm]{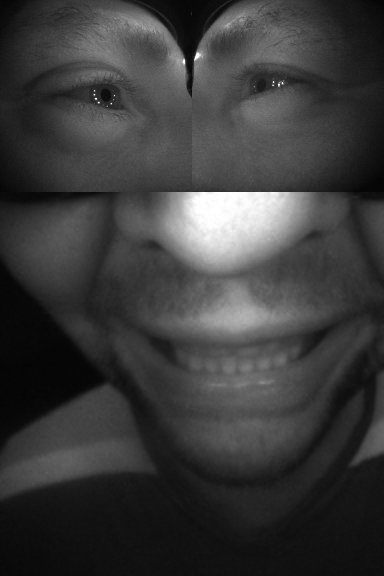} & 
 \includegraphics[trim=90 100 100 70,clip, width=14mm]{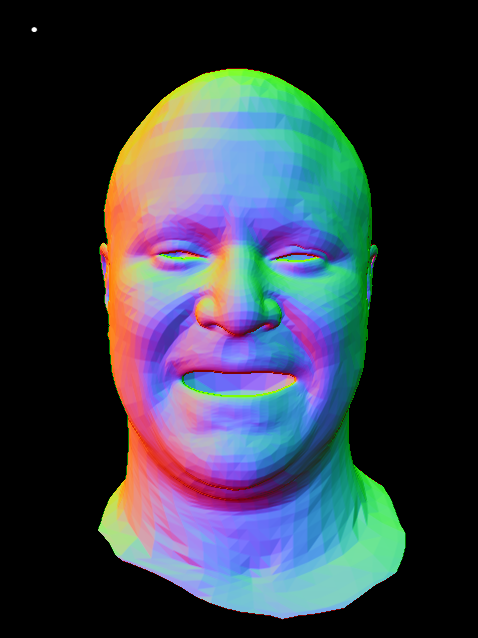} &
 \includegraphics[trim=90 100 100 70,clip, width=14mm]{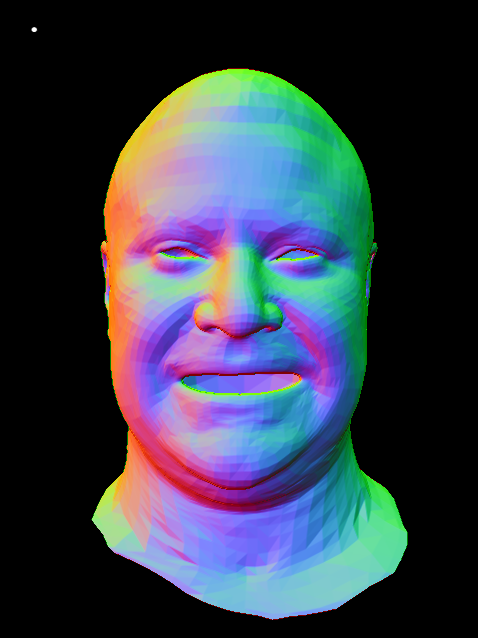} \\
 
  \includegraphics[width=15.32mm]{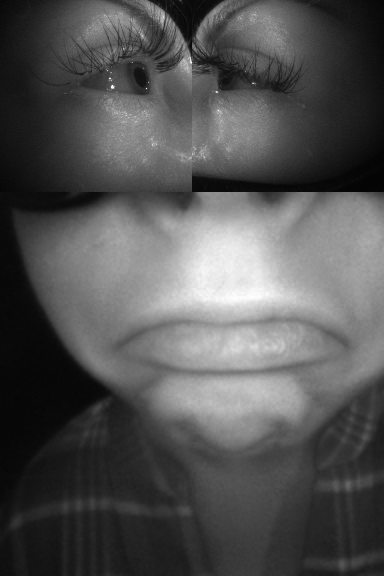} & 
 \includegraphics[trim=90 100 100 70,clip, width=14mm]{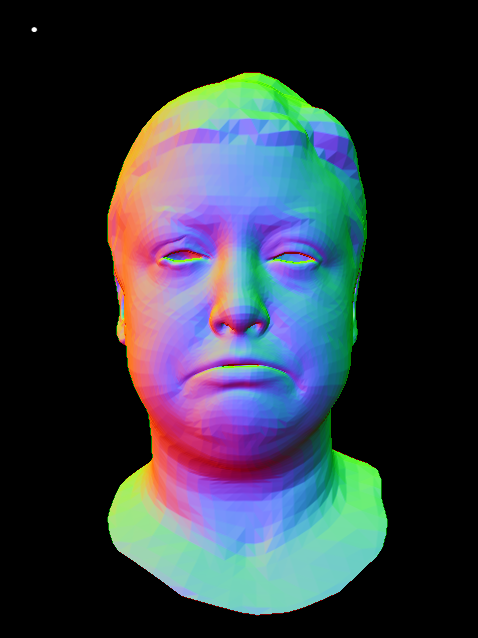} &
 \includegraphics[trim=90 100 100 70,clip, width=14mm]{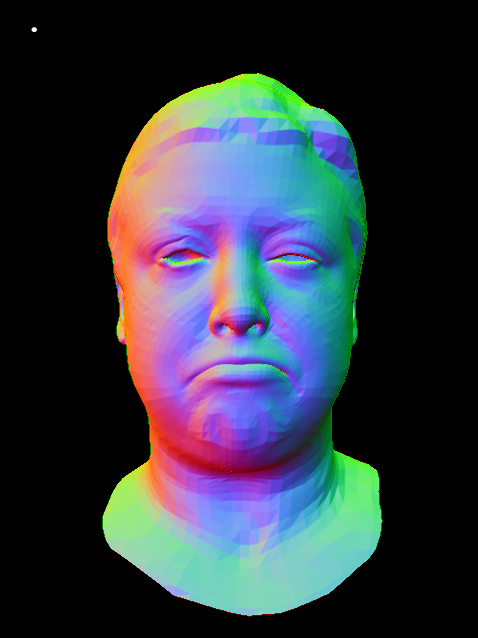} & \ \ &
 
 \includegraphics[width=15.32mm]{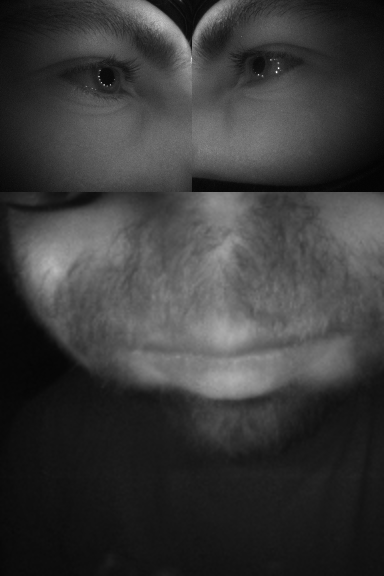} & 
 \includegraphics[trim=90 100 100 70,clip, width=14mm]{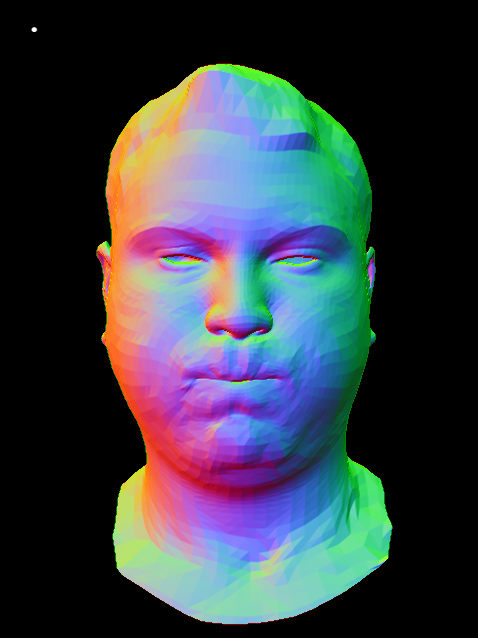} &
 \includegraphics[trim=90 100 100 70,clip, width=14mm]{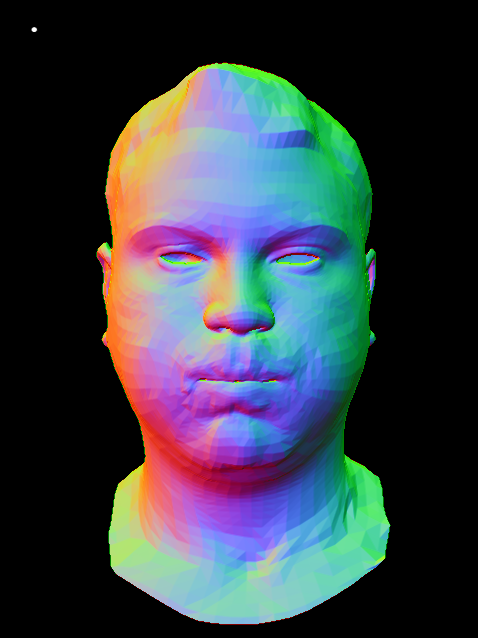} & \ \ &
 
 \includegraphics[width=15.32mm]{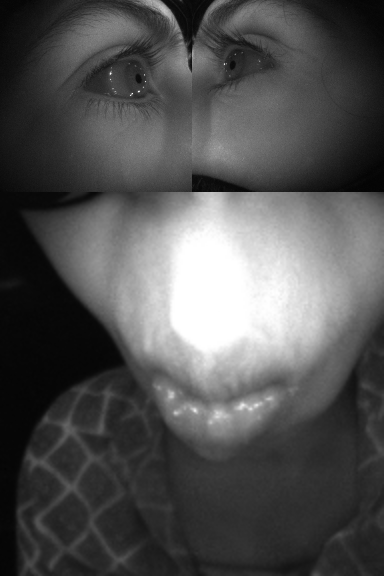} & 
 \includegraphics[trim=90 100 100 70,clip, width=14mm]{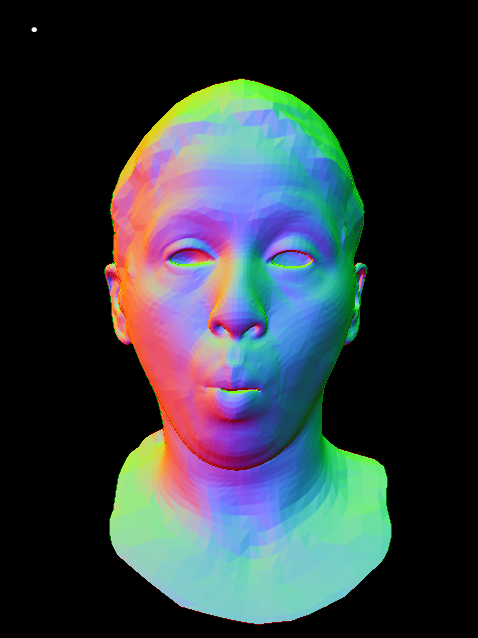} &
 \includegraphics[trim=90 100 100 70,clip, width=14mm]{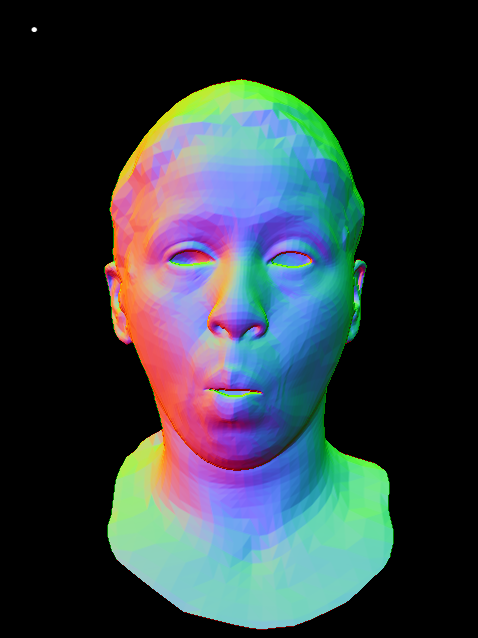} \\

 HMC & MIA & GT & & HMC & MIA & GT & & HMC & MIA & GT \\
 
\end{tabular}

\end{center}
\figvspaceB
\caption{The testing results for estimating the $3$D shapes for six untrained subjects and their ground truth  based on $11$-view results in~\cite{wei2019vr}.}
\label{fig:test_final_images_3D_shape}
\figvspace
\end{figure*}

\Paragraph{Total Loss:}The entire MIA network is trained end-to-end to optimize the networks' parameters by minimizing: 
\begin{equation}
\mathop{\text{minimize}}_{\psi, \gamma, \{\theta_i\}_{i=0}^K} \ \ \sum_{i=0}^K \  \textbf{L}^i_T + \lambda_\textbf{S}\textbf{L}_S^i,
\eqnvspace
\end{equation}
where $K$ is the number of subjects and $\lambda_\textbf{S}$ is the weight for the shape loss.

\SubSection{Augmentation} 
\label{sec:augmentation}
Data augmentation is a wildly practiced heuristic in many deep learning tasks. The main goal is to make the distribution of variations in training data more similar to those in the test set. Most common data augmentations techniques include scaling~\cite{simonyan2014very}, color augmentation~\cite{krizhevsky2012imagenet}, simple geometric transformations~\cite{sharif2014cnn}, and utilizing synthetic data~\cite{klaudiny2017real, mcdonagh2016synthetic}. However, a major source of variability in our task stems from headset factors, such as variations in camera placement and focus, as well as the slop of the headset relative to the face which varies during usage. These variations are not easily modeled using standard augmentation techniques that do not take the $3$D shape of the face into account. In this paper, we simulate headset-based variations by perturbing the 3D rotation and translation of the face shape in the training set, and use it to re-render augmented views of each HMC image on random backgrounds. Some examples are shown in Fig.~\ref{fig:augmentation}. As demonstrated in the experiments section below, this simple augmentation technique substantially improves the robustness of our method to real world variations.

\Section{Experimental Results}

This section reports experimental results and analysis on MIA. The first experiment shows how MIA can estimate accurate $3$D shapes directly from HMC images of untrained subjects. In the second experiment, we evaluate the quality of MIA's texture prediction for identities with pre-trained avatars under challenging testing scenarios. In the third experiment,  we show how MIA can incorporate new subjects with minimal training. In addition, we also present further analysis about what MIA learns prior and during adaptation.

\Paragraph{Data:}We used $120$ HMC captures of different subjects for training and $21$ HMC captures for testing. Training and testing HMC captures do not overlap. Each HMC capture is a $45$ minutes long video ($30$fps) of $11$-views HMC images, and contains $73$ peak expressions, two sets of continuous range-of-motion, recitation of $50$ sentences and $5$-$10$ minutes of conversion.  The HMC images are in the IR spectrum with a resolution of $480 \times 640$. During testing only $3$-views are available. For each subject, we have a pre-trained decoder to generate PS texture  for various expressions from arbitrary views. For more information, of how to build the PS decoder see~\cite{lombardi2018deep} and Eqn.~\ref{equ:equ_decoder}.

\Paragraph{Ground Truth:} We utilize the result of the method in~\cite{wei2019vr}, that solves for the correspondence between $11$-views HMC images and the CA parameters as the ground truth. Recall that the training data is captured with $11$-views to achieve more precise results in the correspondence between HMC and CA, while the testing data has only $3$-views.

\Paragraph{Baseline method:} We compare MIA with the person specific (PS) encoder in~\cite{wei2019vr}. The PS encoder is trained with one HMC capture ($3$-view images) and uses a CNN architecture with the same number of parameters as ours.

\Paragraph{Evaluation metrics:}We report the average Euclidean error for the eyes, mouth and face areas separately for both $3$D shape and the texture. The $3$D shape errors are measured in millimeters and the texture errors in raw intensity values (i.e. 0-255). We report the localized error metrics to analyze failure modes better. For example, the $3$D shape error in the eyes capture openness and blinking errors, while in the mouth, they capture deviations in lip shapes important for visual-speech. Similarly, texture error in the eyes is typically due to the errors in gaze direction, and in mouth, it corresponds to incorrect teeth and tongue estimation.

\begin{figure*}[!t]
\begin{center}
\setlength\tabcolsep{0.1pt}
\small
\begin{tabular}{cccccccccc}
 \includegraphics[width=14mm]{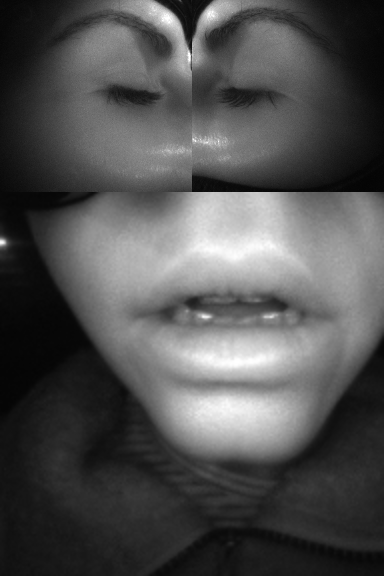} & 
 \includegraphics[width=14mm]{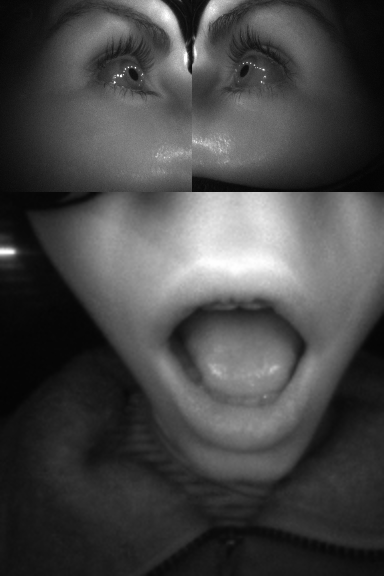} & 
 \includegraphics[width=14mm]{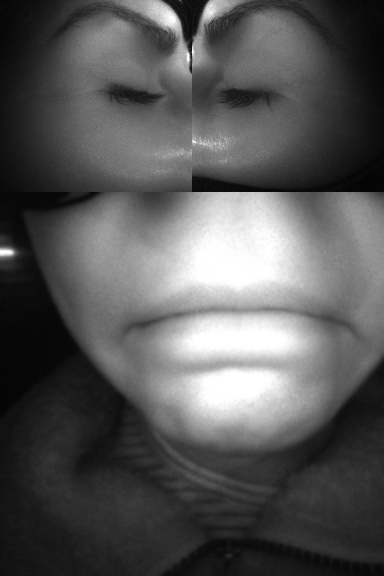} & 
 \includegraphics[width=14mm]{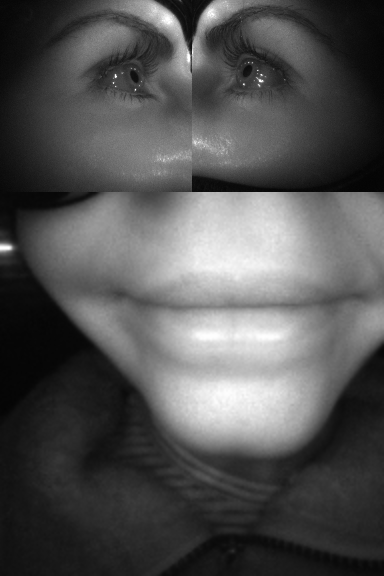} & \includegraphics[width=14mm]{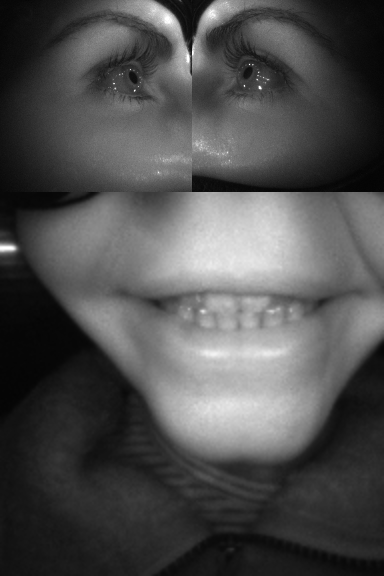} & 
 \includegraphics[width=14mm]{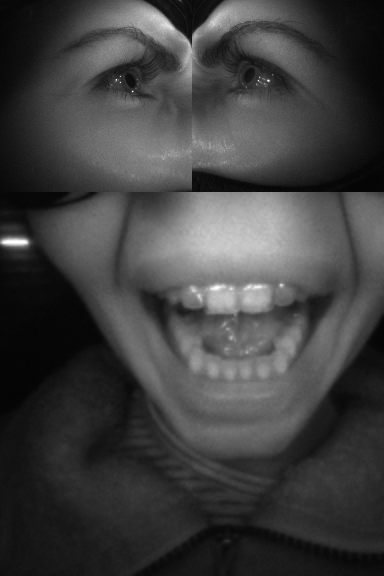} & 
 \includegraphics[width=14mm]{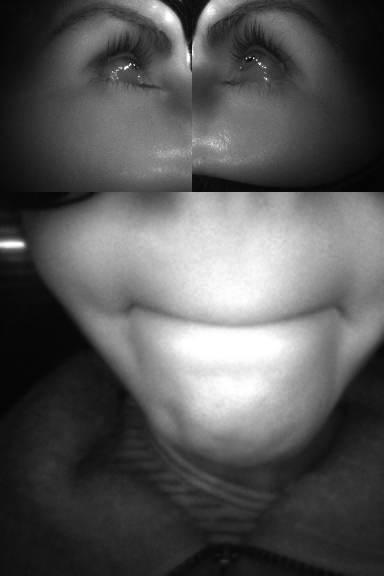} & 
 \includegraphics[width=14mm]{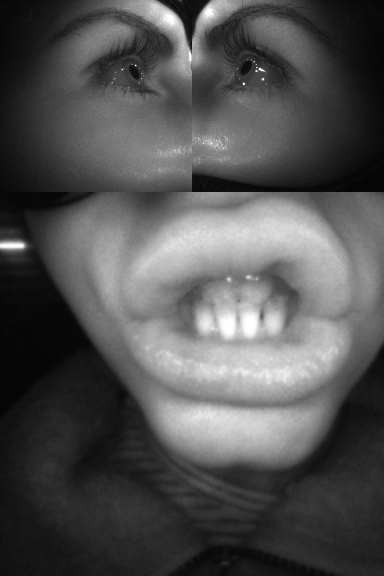} & 
 \includegraphics[width=14mm]{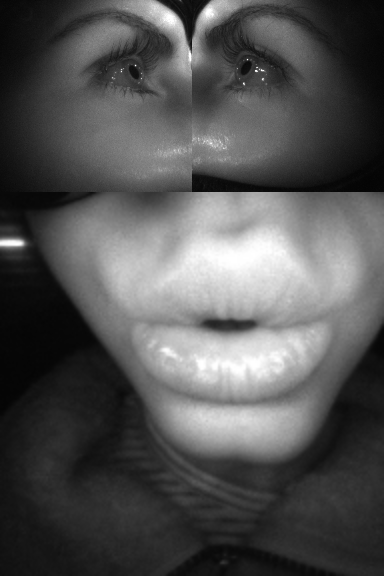} & 
 \includegraphics[width=14mm]{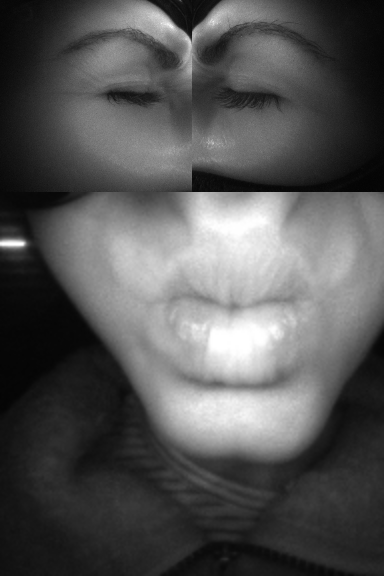} \\
  \includegraphics[trim=90 100 100 70,clip, width=14mm]{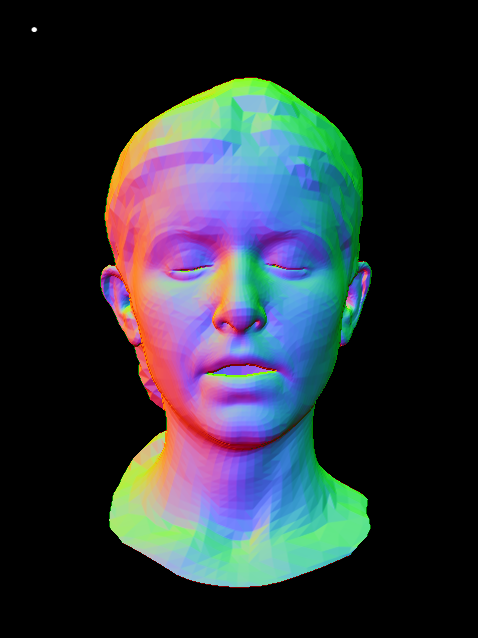} & 
 \includegraphics[trim=90 100 100 70,clip, width=14mm]{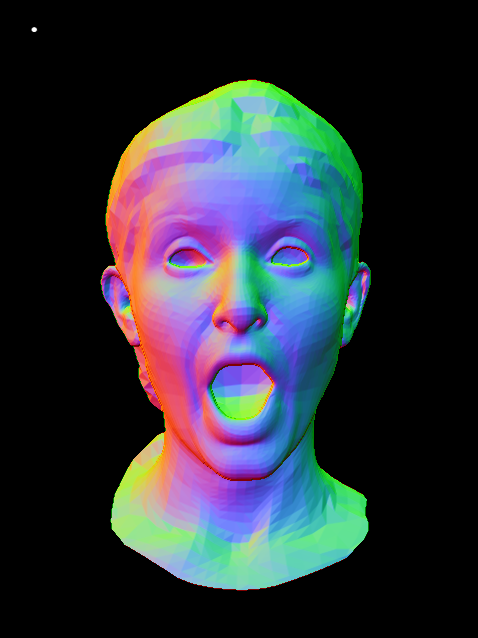} & 
 \includegraphics[trim=90 100 100 70,clip, width=14mm]{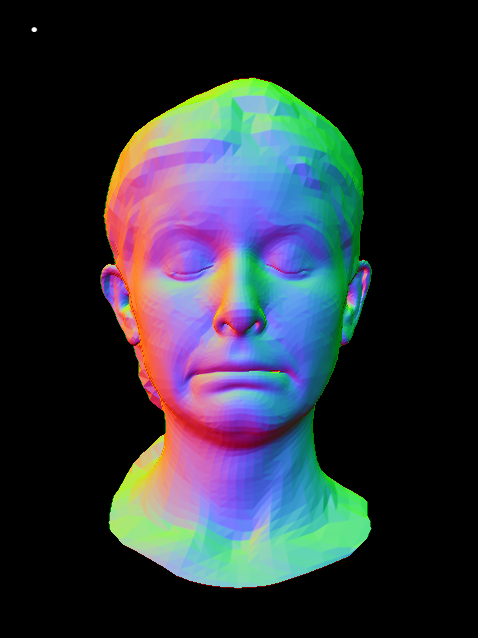} & 
 \includegraphics[trim=90 100 100 70,clip, width=14mm]{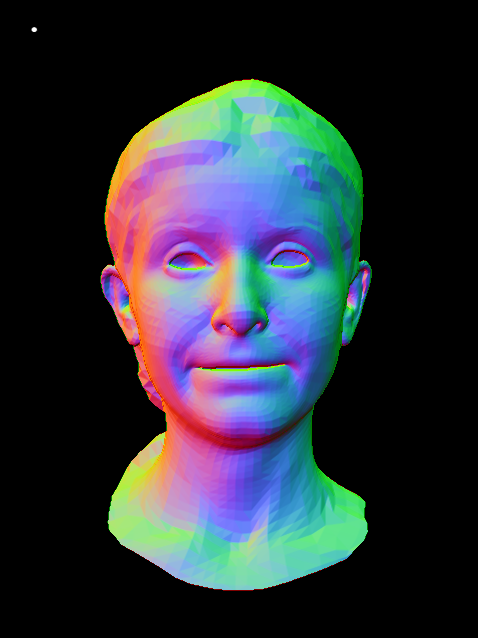} & 
 \includegraphics[trim=90 100 100 70,clip, width=14mm]{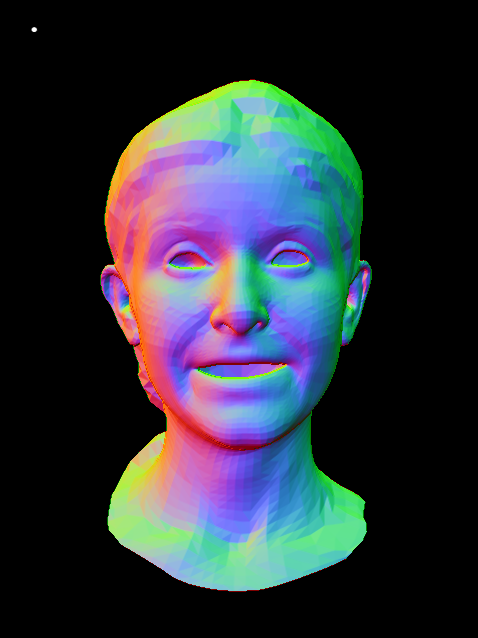} & 
 \includegraphics[trim=90 100 100 70,clip, width=14mm]{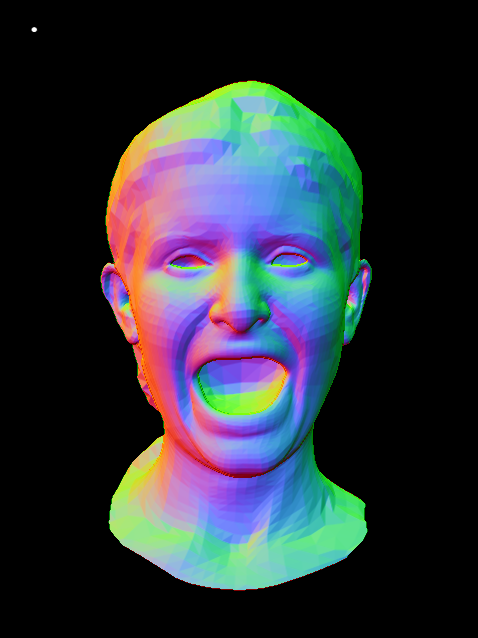} & 
 \includegraphics[trim=90 100 100 70,clip, width=14mm]{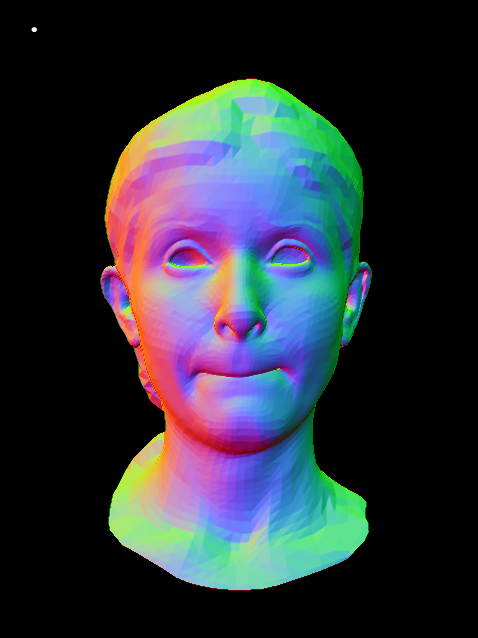} & 
 \includegraphics[trim=90 100 100 70,clip, width=14mm]{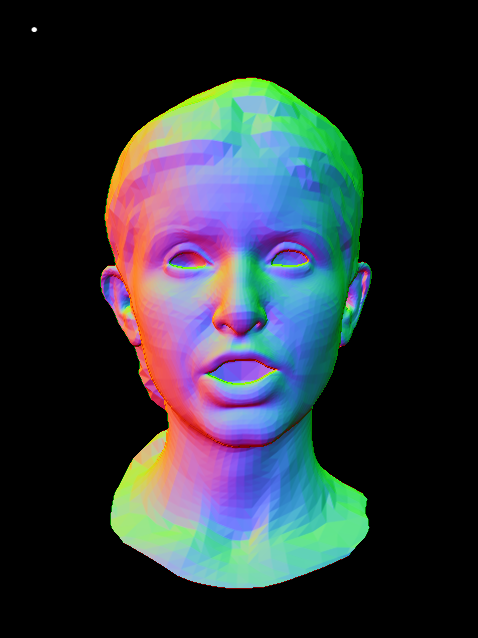} & 
 \includegraphics[trim=90 100 100 70,clip, width=14mm]{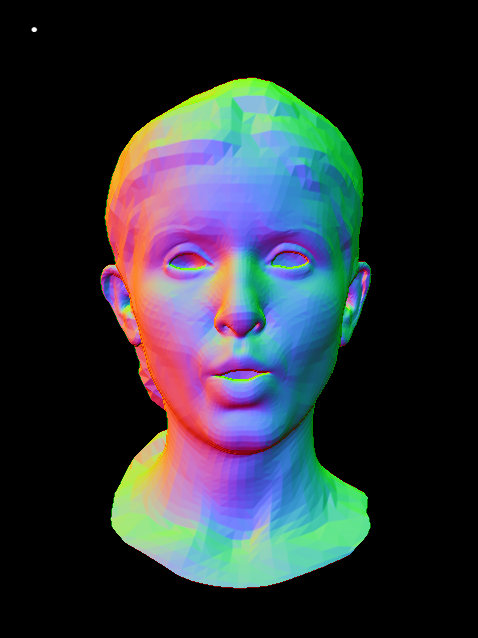} & 
 \includegraphics[trim=90 100 100 70,clip, width=14mm]{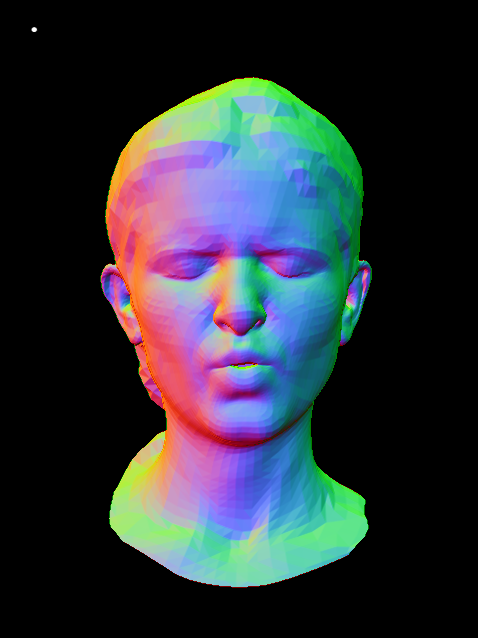} 

\end{tabular}
\end{center}
\figvspaceB
\caption{Test results to estimate the  $3$D shape from HMC images of an untrained subject for a wide range of expressions.}
\label{fig:test_shape_one_person_diff_exp}
\figvspace
\end{figure*}

\begin{table}[t]
\begin{center}
\footnotesize
\caption{Test results for $3$D shape estimation in untrained subjects. }
\vspace{-2mm}
\begin{tabular}{c| c c c}
\toprule
\multirow{2}{*}{Subject} & \multicolumn{3}{c}{$3$D Shape Error (mm)} \\\cline{2-4}
 & Face & Eyes & Mouth \\\hline
person $1$ & $1.68$ & $1.08$ & $2.90$\\ 
person $2$ & $1.51$ & $1.21$ & $2.32$ \\
person $3$ & $1.07$ & $0.74$ & $1.82$ \\
person $4$ & $1.84$ & $1.20$ & $2.92$ \\
person $5$ & $1.45$ & $0.89$ & $2.23$ \\
person $6$ & $1.47$ & $0.97$ & $2.56$ \\
person $7$ & $1.73$ & $1.20$ & $2.98$ \\
person $8$ & $1.57$ & $0.89$ & $2.52$ \\
person $9$ & $1.21$ & $0.91$ & $1.84$ \\
person $10$ & $1.58$ & $1.00$ & $2.68$ \\\hline \hline
\textbf{overall} & $1.51\pm0.23$ & $1.00\pm0.16$ & $2.47\pm0.42$ \\\hline
\end{tabular}
\label{tab:test_shape_error_untrained}
\end{center}
\tabvspace
\vspace{-2mm}
\end{table}

\begin{table*}[b]
\begin{center}
\vspace{-4mm}
\footnotesize
\caption{Testing results for a subject with multiple testing HMC captures with different variations. One HMC capture of the subject is inside the training set. The $3$D shape errors are in mm and the texture errors are in intensity.}
\resizebox{\linewidth}{!}{
\begin{tabular}{ c | c | c | c | c c c | c c c}
\toprule
Test & Sample  & \multirow{2}{*}{Variations} & \multirow{2}{*}{Method}& \multicolumn{3}{c}{$3$D Shape Error} &  \multicolumn{3}{c}{Texture Error}  \\ \cline{5-10}
Capture & Image &  &  & Face & Eyes & Mouth & Face & Eyes & Mouth \\ \hline
\multirow{2}{*}{1} & \multirow{2}{*}{Fig.~\ref{fig:test_samples}.(b)} & \multirow{2}{*}{Headset} & PS~\cite{wei2019vr}  & $0.85$ & $0.65$ & $1.33$ & $1.13$ & $1.93$ & $1.50$ \\ \cline{4-10} 
& & & MIA & $1.20$ & $0.85$ & $1.90$ & $1.33$ & $2.34$ & $1.79$ \\ \hline
\multirow{2}{*}{2}& \multirow{2}{*}{Fig.~\ref{fig:test_samples}.(c)} & Headset & PS~\cite{wei2019vr}  & $2.04$ & $0.77$ & $4.71$ & $1.84$ & $2.47$ & $3.51$ \\ \cline{4-10}
& & Facial appearance & MIA & $1.28$ & $0.79$ & $2.22$ & $1.49$ & $2.52$ & $2.00$ \\ \hline
\multirow{2}{*}{3}& \multirow{2}{*}{Fig.~\ref{fig:test_samples}.(d)} & Headset & PS~\cite{wei2019vr} & $1.90$ & $0.98$ & $3.68$ & $1.65$ & $2.73$ & $2.84$ \\ \cline{4-10} 
& & Facial appearance & MIA & $1.26$ & $0.86$ & $2.23$ & $1.32$ & $2.52$ & $2.03$ \\ \hline
\multirow{2}{*}{4}& \multirow{2}{*}{Fig.~\ref{fig:test_samples}.(e)} & Environment & PS~\cite{wei2019vr} & $2.21$ & $0.86$ & $4.92$ & $1.92$ & $2.33$ & $3.39$ \\ \cline{4-10} 
& & Facial appearance & MIA & $1.14$ & $0.73$ & $1.94$ & $1.45$ & $2.11$ & $2.05$ \\ \hline \cline{3-10}
\multicolumn{2}{c|}{} & \multirow{2}{*}{\textbf{overall}} & PS~\cite{wei2019vr} & $1.75\pm0.61$ & $0.81\pm0.13$ & $3.66\pm1.64$ & $1.63\pm0.35$ & $\textbf{2.36}\pm\textbf{0.33}$ & $2.81\pm0.92$ \\ \cline{4-10} 
\multicolumn{2}{c|}{} &  & MIA & $\textbf{1.22}\pm\textbf{0.06}$ & $\textbf{0.80}\pm\textbf{0.06}$ & $\textbf{2.07}\pm\textbf{0.17}$ & $\textbf{1.39}\pm\textbf{0.08}$ & $2.37\pm0.19$ & $\textbf{1.96}\pm\textbf{0.12}$ \\ \cline{3-10}
\end{tabular}}
\label{tab:test_1}
\end{center}
\tabvspace
\end{table*}

\Paragraph{Implementation details:}In training, we use the Adam optimizer, setting the batch size to $32$ and the initial learning rate to $1\mathrm{e}{-3}$. We decrease learning rate by $8\mathrm{e}{-1}$ after each $25$K iterations. In total, we train the encoder for $250$K iterations, and set both of $\lambda_\textbf{T}$ and $\lambda_\textbf{S}$ to $100$. We crop and resize the HMC images to $192\times192$ to focus on the face areas. 

The backbone network, $\textbf{B}_\psi$, consists of two residual networks~\cite{he2016deep}, one for eye images $\left[\textbf{H}^0,\textbf{H}^1\right]\in \mathbb{R}^{192\times192\times2}$ and another for the mouth $\textbf{H}^2\in \mathbb{R}^{192\times192}$. Each network consists of a Res-Net head module, five BottleNeck blocks and a $64$-way fully connected layer. Each BottleNeck block consists of ten convolutional layers with $3\times3$ and $1\times1$ filters. We add shortcut connections among the convolutional layers, and each layer is followed by ReLU~\cite{nair2010rectified} and Instance normalization~\cite{ulyanov2016instance} layers. To extract the final identity invariant features, we apply a global average pooling and a $64$-way fully connected layer to the  activations of the last BottleNeck block.
The architecture of the $3$D shape network, $\textbf{G}_\gamma$, consists of four fully connected layers where each one is followed by a leaky ReLU~\cite{xu2015empirical} layer with negative slope of $0.2$. We normalize the extracted features from the HMC images and the neutral $3$D shape, to account for their different domains, by employing group normalization~\cite{wu2018group} after concatenating the features. Finally, for the texture network $\textbf{F}_\theta$, we utilize the combination of a ReLU layer and a fully connected layer without bias.

\SubSection{Quantitative Evaluations}
This section quantifies the performance of MIA using three experiments: (1) driving the $3$D shape of untrained subjects. (2) robustness of shape and texture estimation for subjects with trained PS models. (3) generalization of learned features on new subjects. 

\Paragraph{Driving 3D Shape:} Inputs to the shape generation network, $\textbf{G}_\gamma$, are the HMC images and the corresponding identitiy's neutral $3$D shape. We train the network  with 120 subjects using the loss function in Eqn.~\ref{equ:loss_shape} as guidance. Fig.~\ref{fig:test_final_images_3D_shape} shows the estimated $3$D shape for extreme expression examples from six untrained subjects along with their ground truth. Our $3$D shape estimator captures subtle details in expressions necessary for inferring social signal. Table~\ref{tab:test_shape_error_untrained} shows the $3$D shape errors for face, eyes and mouth areas of the whole sequence for ten untrained subjects. The error is less than 2mm in the face/eyes and 3mm in the mouth.  Recall that MIA does not use any sample from the test subject other than the neutral shape and has never seen any HMC images for these subjects during training. Fig.~\ref{fig:test_shape_one_person_diff_exp} shows testing results of one untrained subject for a wide range of expressions. Note that PS~\cite{wei2019vr} is not able to estimate the $3$D shape for untrained subjects.

 Comparing Table~\ref{tab:test_shape_error_untrained} with PS's results for the $3$D shape error in Table~\ref{tab:test_1} (different capture), we find that MIA outperforms PS, despite PS having access to subject-specific HMC images, and their target shapes, during training. We suspect the reason for this is that MIA learns to marginalize the extrinsic variability of the problem (i.e. environment, headset) from the 120 subjects that is trained on, while the PS tends to overfit to the specific HMC capture session used for training. More comparative results can be found in the video in the supplementary material.

\Paragraph{Driving Full Avatars:}
In this experiment, we evaluate the ability of MIA to generate both shape and texture and its robustness against extrinsic factors such as headset, environment and facial appearance variations. Here, data for test subject is available during training, but from a different HMC capture. The selected subject was captured on five different dates; examples of the HMC images are shown in Fig.~\ref{fig:test_samples}. These samples show large appearance variations due to facial hair, pose changes in the headset slop, and camera assembly differences across headsets; it also contains background variation due to changing environment and overall lighting differences.  We use one HMC capture (Fig.~\ref{fig:test_samples}(a)) of the subject with $119$ HMC captures of other subjects for training, and test on the remaining four HMC captures of that subject. Table~\ref{tab:test_1} compares the testing errors of MIA against PS~\cite{wei2019vr}. On test capture $1$, which is very similar to the training capture, PS~\cite{wei2019vr} performs better than MIA. But, its performance declines significantly when testing on the other captures, where variations in environment and facial appearance are more extreme. Note that the overall errors for MIA, for all areas of $3$D shape and texture, are more stable and are similarly low across all test captures. The first two rows of Fig.~\ref{fig:test_final_images} shows visual comparison of methods on the test HMC captures, where a significant reduction in expressive detail is noticeable in results for PS~\cite{wei2019vr}. We refer the reader to the supplementary material for more results.

\Paragraph{Adaptation to New Identities:} We evaluated the generalization of MIA's feature extraction to new subjects on HMC captures of $6$ subjects that are not trained in MIA. Each of the 6 subjects has more than one HMC capture exhibiting variations in extrinsic factors. We used the pre-trained MIA network with $120$ subjects (excluding the test 6 subjects), and fix the shape generation network, $\mathbf{G}$, and backbone network, $\textbf{B}_\psi$. For each new subject, we trained a new small texture network $\textbf{F}_\theta$. During the testing on HMC captures with variations, we used the newly trained texture estimation branch for estimating the texture parameters, and decode both the texture and the $3$D shape by utilizing Eqn.~\ref{equ:equ_decoder}. Table~\ref{tab:test_second} shows the overall errors for $3$D shape and texture for different areas of $7$ testing HMC captures of the $6$ subjects. MIA achieves lower errors for all areas with smaller variability, demonstrating the effectiveness of the features extracted from the fixed backbone network. The last three rows of Fig.~\ref{fig:test_final_images} show visualizations of this case for.

\begin{table}[t]
\begin{center}
\setlength\tabcolsep{1.5pt}
\small
\caption{Testing results for training and testing on new subjects with pre-trained fixed backbone network.}
 \resizebox{\linewidth}{!}{
\begin{tabular}{ c|c|c|c| c | c | c}
\toprule
\multirow{2}{*}{Method}& \multicolumn{3}{c|}{$3$D Shape Error} &  \multicolumn{3}{|c}{Texture Error}  \\ \cline{2-7}
 & Face & Eyes & Mouth & Face & Eyes & Mouth \\ \hline
PS~\cite{wei2019vr} & $1.12\pm0.26$ & $0.74\pm0.11$ &	$1.98\pm0.65$ & $\textbf{2.22}\pm\textbf{0.61}$ &	$2.90\pm0.95$ & $2.77\pm0.83$\\
MIA & $\textbf{1.05}\pm\textbf{0.19}$ & $\textbf{0.74}\pm\textbf{0.09}$ &	$\textbf{1.71}\pm\textbf{0.37}$ & $2.22\pm0.62$ &	$\textbf{2.88}\pm\textbf{0.81}$ & $\textbf{2.65}\pm\textbf{0.78}$\\ \hline
\end{tabular}
}
\label{tab:test_second}
\end{center}
\tabvspace
\end{table}

\begin{figure}[b]
\begin{center}
\vspace{-4mm}
\setlength\tabcolsep{1.05pt}
\small
\begin{tabular}{ c c c c}
 \includegraphics[width=40mm]{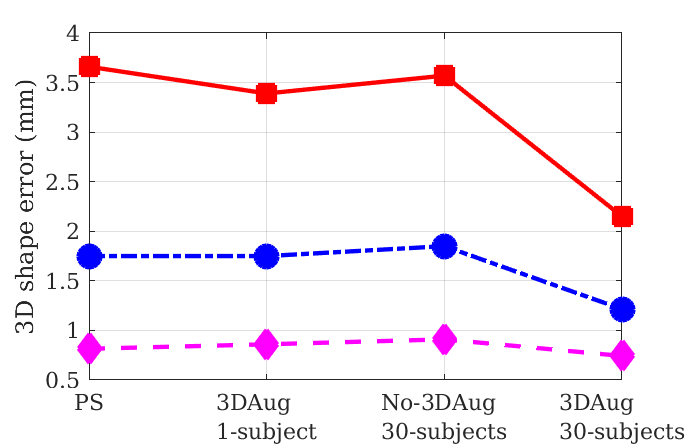} & \includegraphics[width=40mm]{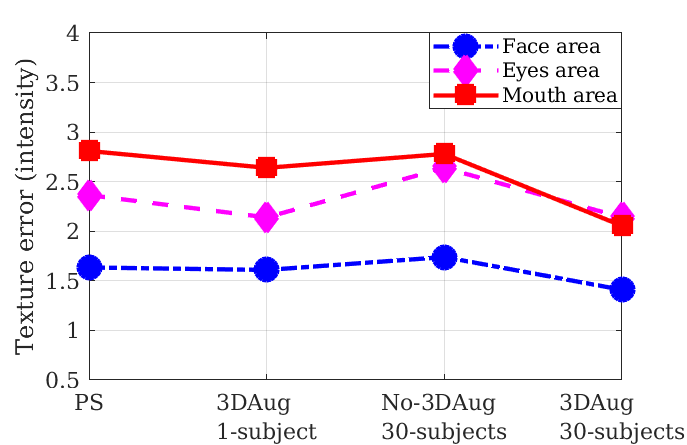}\\
\end{tabular}

\end{center}
\figvspaceB
\caption{The advantage of the $3$D augmentation layer, the errors drop significantly by using both of $3$D augmentation and MIA. }
\label{fig:test_ablation_3DAug}
\figvspace
\end{figure}

\begin{figure}[t]

\begin{center}
\setlength\tabcolsep{0.05pt}
\small
\includegraphics[width=45mm]{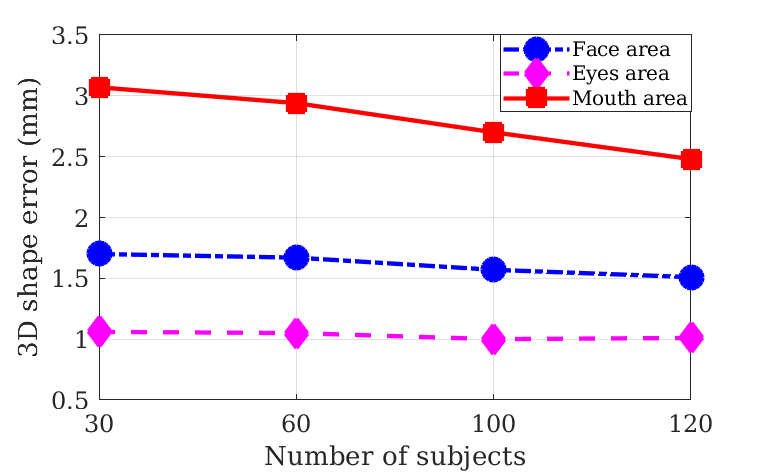}

\end{center}
\figvspaceB
\caption{The influence of number of training subjects. The shape errors are decreasing by increasing number of training subjects. } 
\label{fig:test_ablation_num_sub}
\figvspace
\end{figure}

\SubSection{Ablation Study}

\Paragraph{$3$D augmentation layer:}To analyze the advantage of using the $3$D augmentation layer, we compare the errors of the PS~\cite{wei2019vr} model, MIA with $3$D augmentation trained with $1$ subject, MIA without $3$D augmentation ($3$D Aug) trained with $30$ subjects, and MIA with $3$D Aug trained with $30$ subjects. Fig.~\ref{fig:test_ablation_3DAug} shows the average errors for the four test captures in Table~\ref{tab:test_1}. It shows that even using the $3$D Aug layer with $1$ subject reduces errors slightly in comparison to PS~\cite{wei2019vr}. However,  there is huge drop in errors by using the $3$D Aug layer with $30$ subjects. This reduction of error is more significant in the mouth area. It shows that the combination of MIA and $3$D Aug is effective. 

\Paragraph{Influence of number of subjects:}We evaluate the influence of number of training subjects in the performance of MIA during testing. We train MIA with $30, 60, 100$ and $120$ subjects, and test them on ten untrained subjects for estimating the $3$D shapes. Fig.~\ref{fig:test_ablation_num_sub} shows that by increasing number of training subjects the $3$D shape errors are decreasing, especially for the mouth area.

\begin{figure*}[t]
\setlength\tabcolsep{0.1pt}
\small
\vspace{2mm}
\begin{center}

\begin{tabular}{cccccccccc}
 \includegraphics[width=20.5mm]{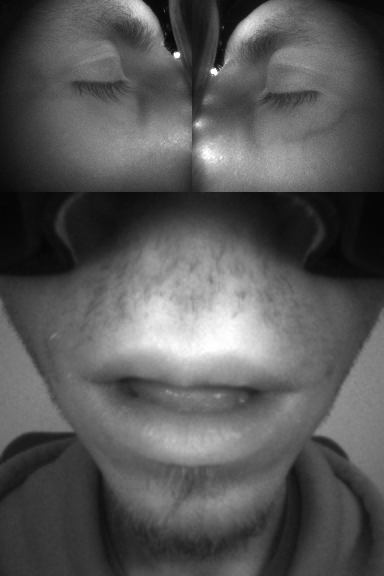} &
  \includegraphics[trim=90 100 100 70,clip, width=17mm]{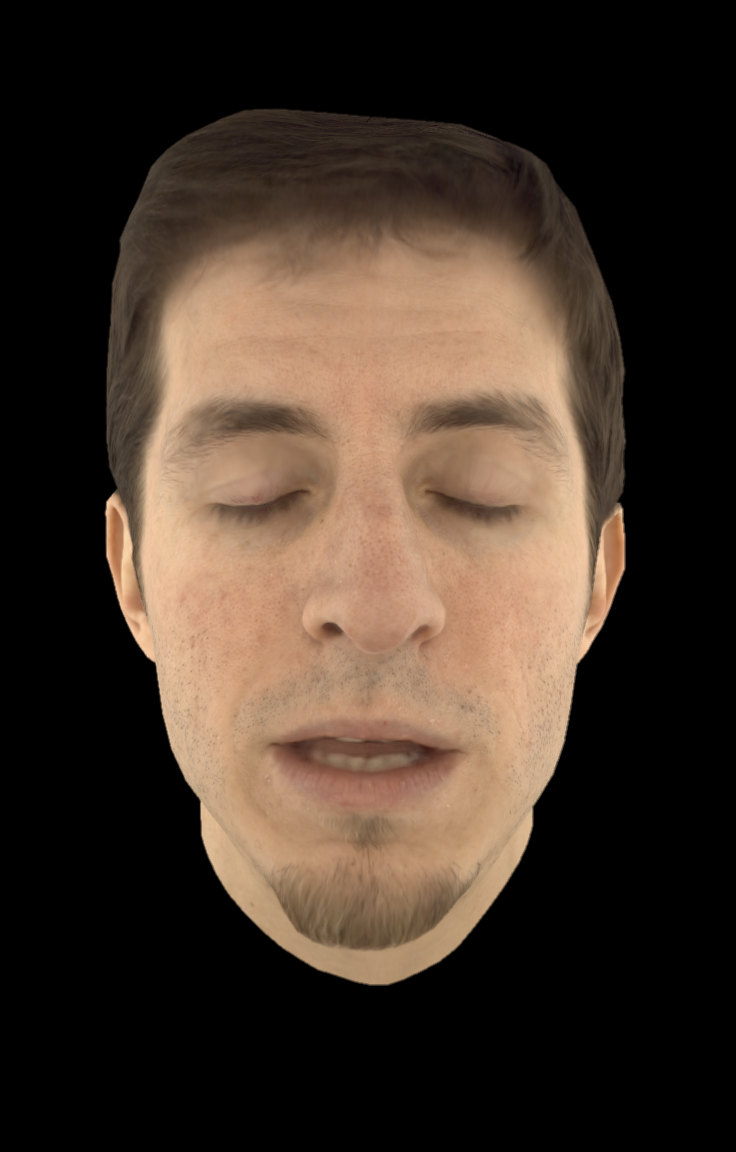} & 
 \includegraphics[trim=90 100 100 70,clip, width=17mm]{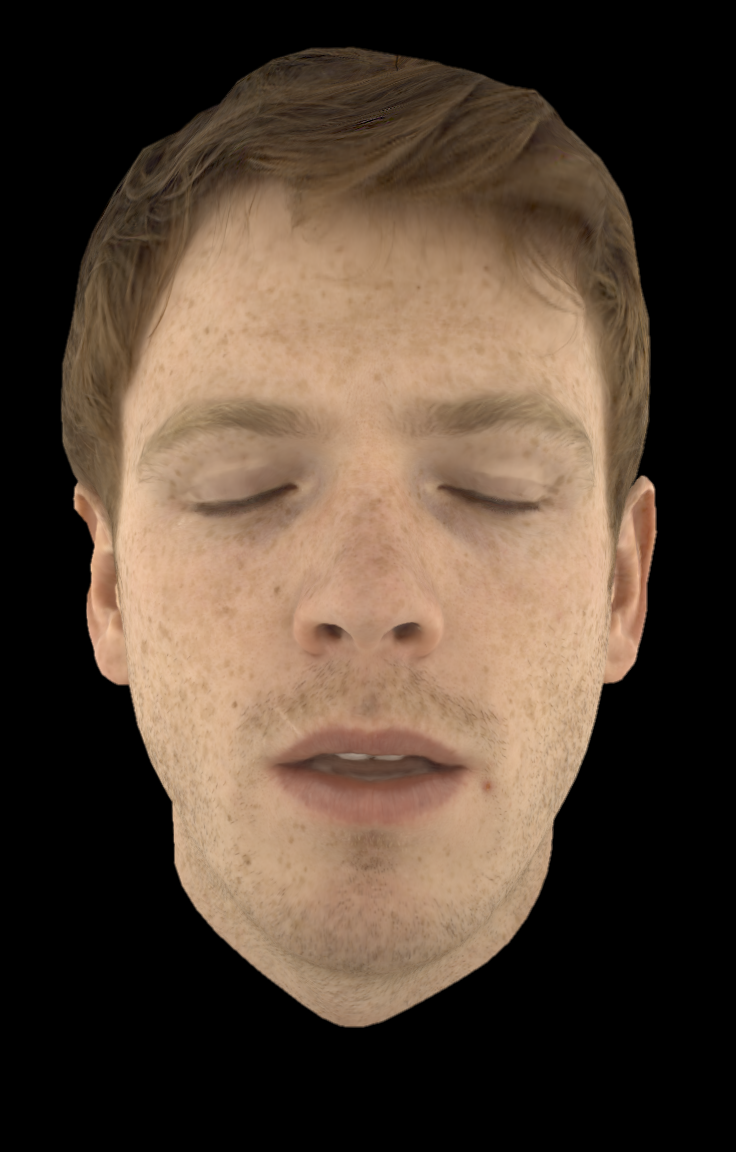} & 
 \includegraphics[trim=90 100 100 70,clip, width=17mm]{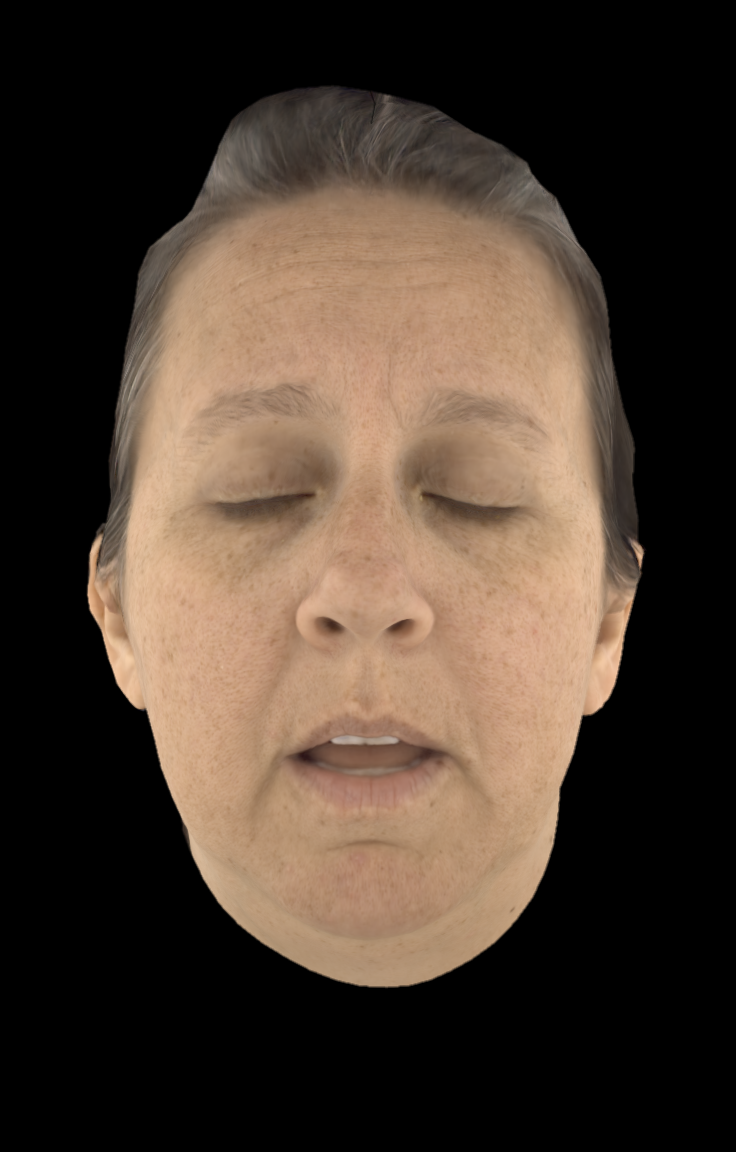} & 
 \includegraphics[trim=90 100 100 70,clip, width=17mm]{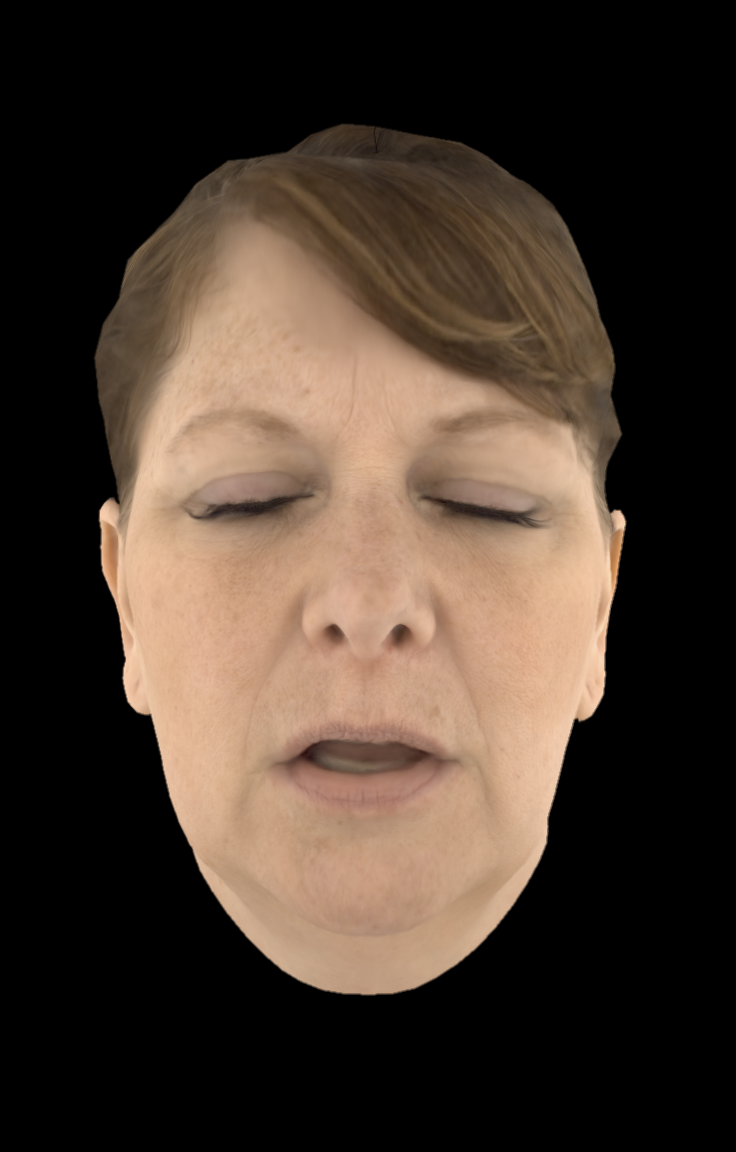} & 
 \includegraphics[trim=90 100 100 70,clip, width=17mm]{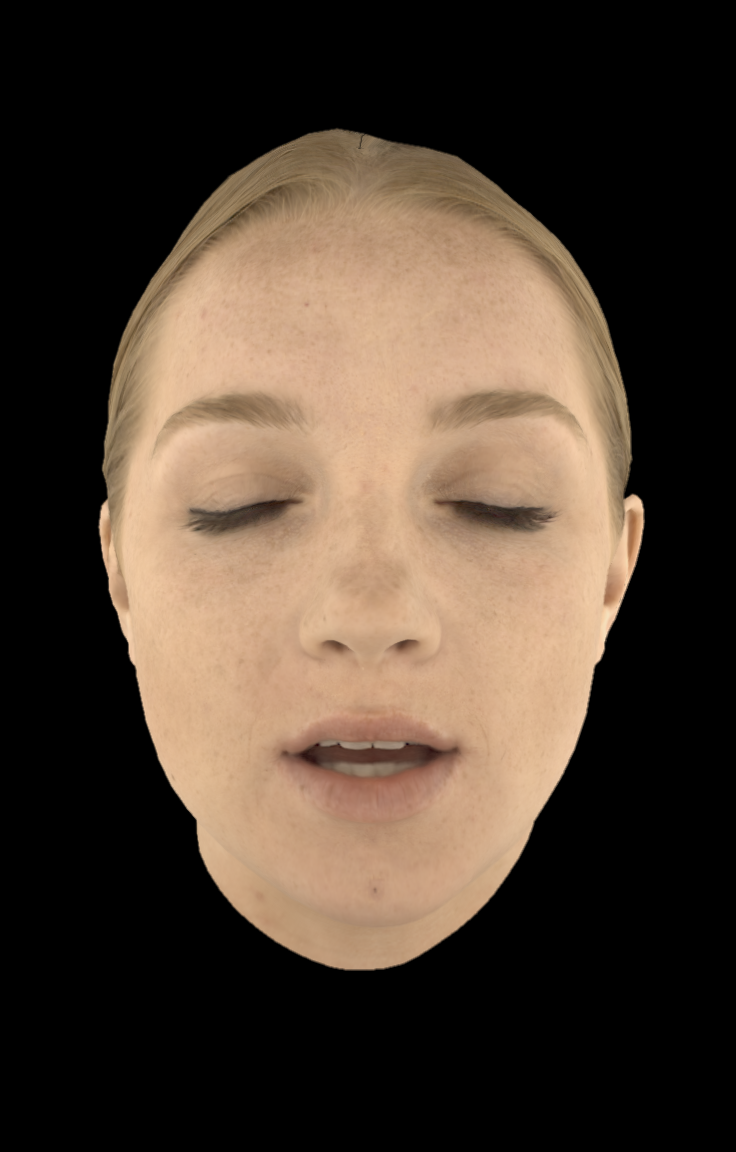} & 
 \includegraphics[trim=90 100 100 70,clip, width=17mm]{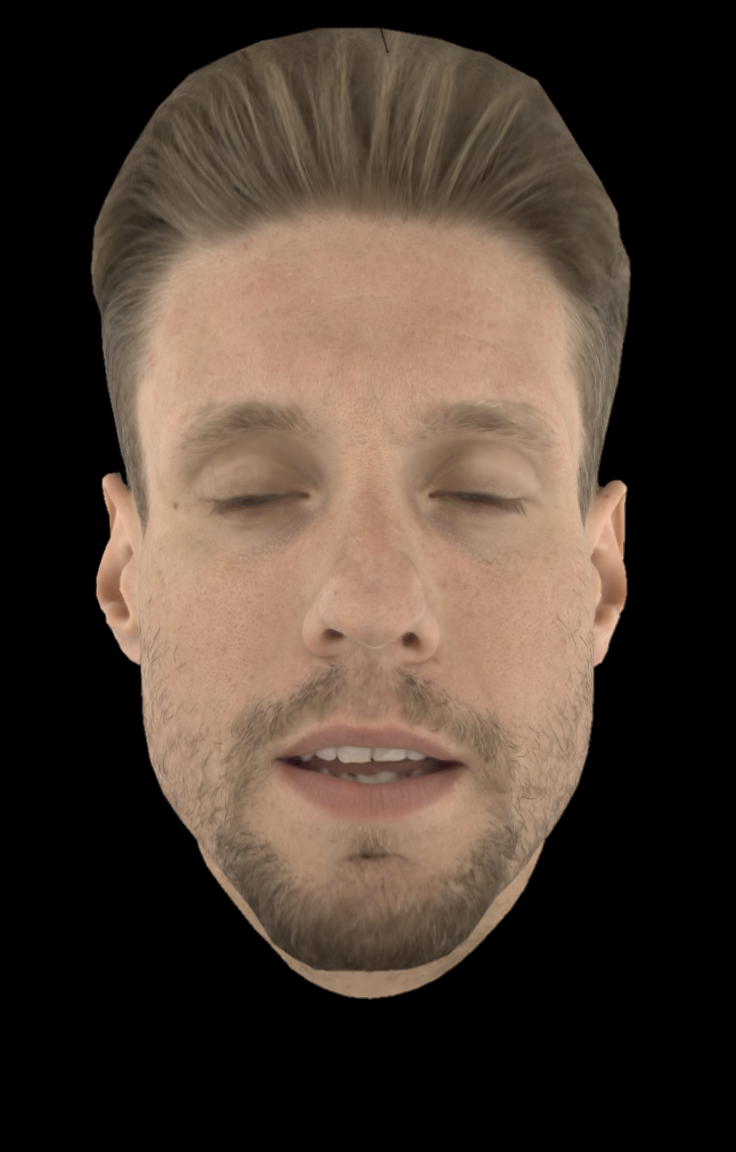} & 
 \includegraphics[trim=90 100 100 70,clip, width=17mm]{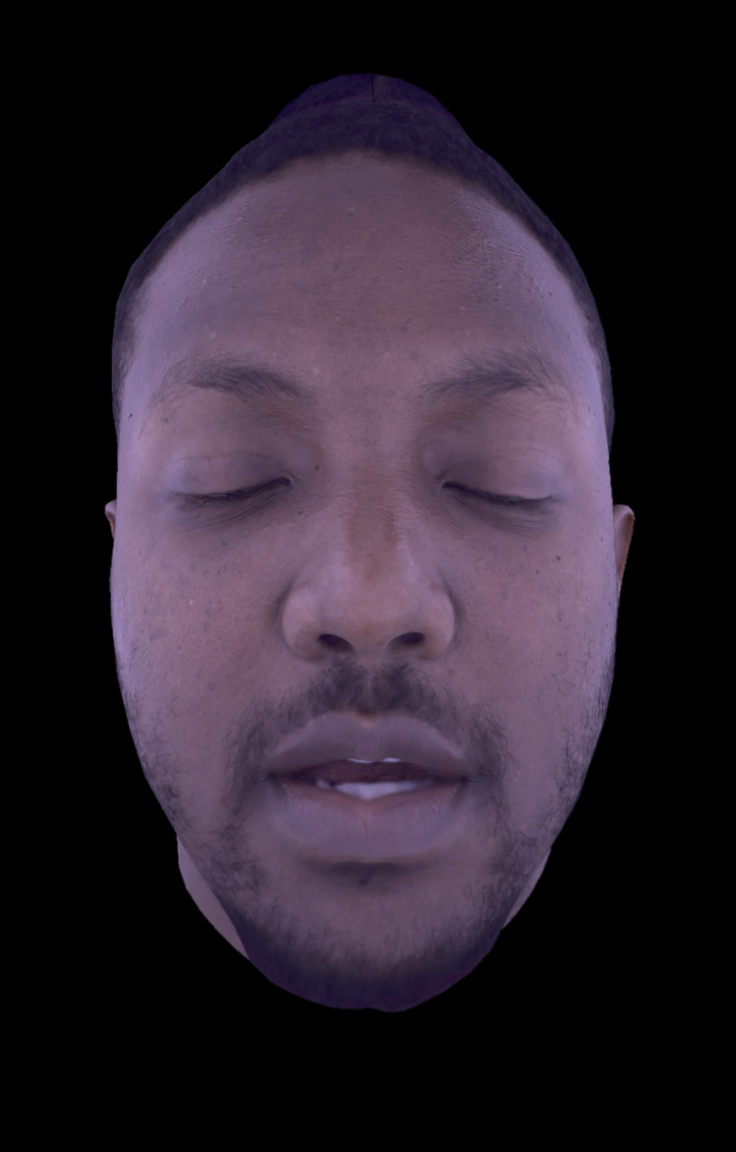} & 
 \includegraphics[trim=90 100 100 70,clip, width=17mm]{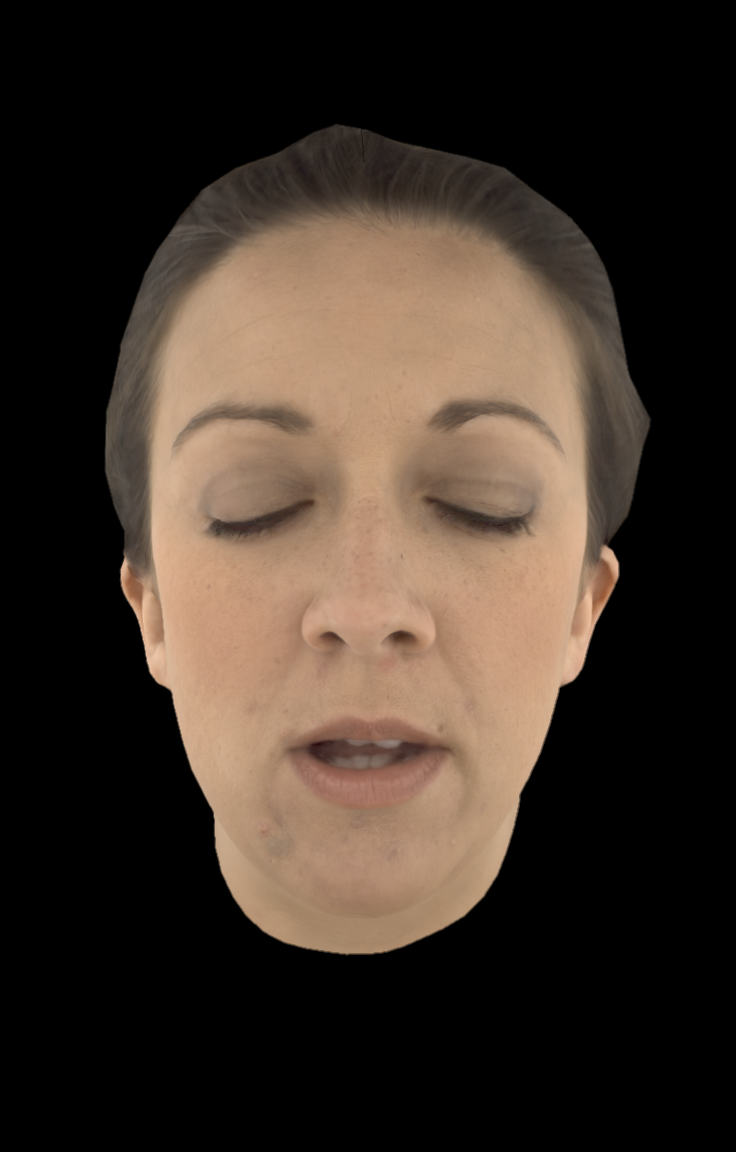} & 
 \includegraphics[trim=90 100 100 70,clip, width=17mm]{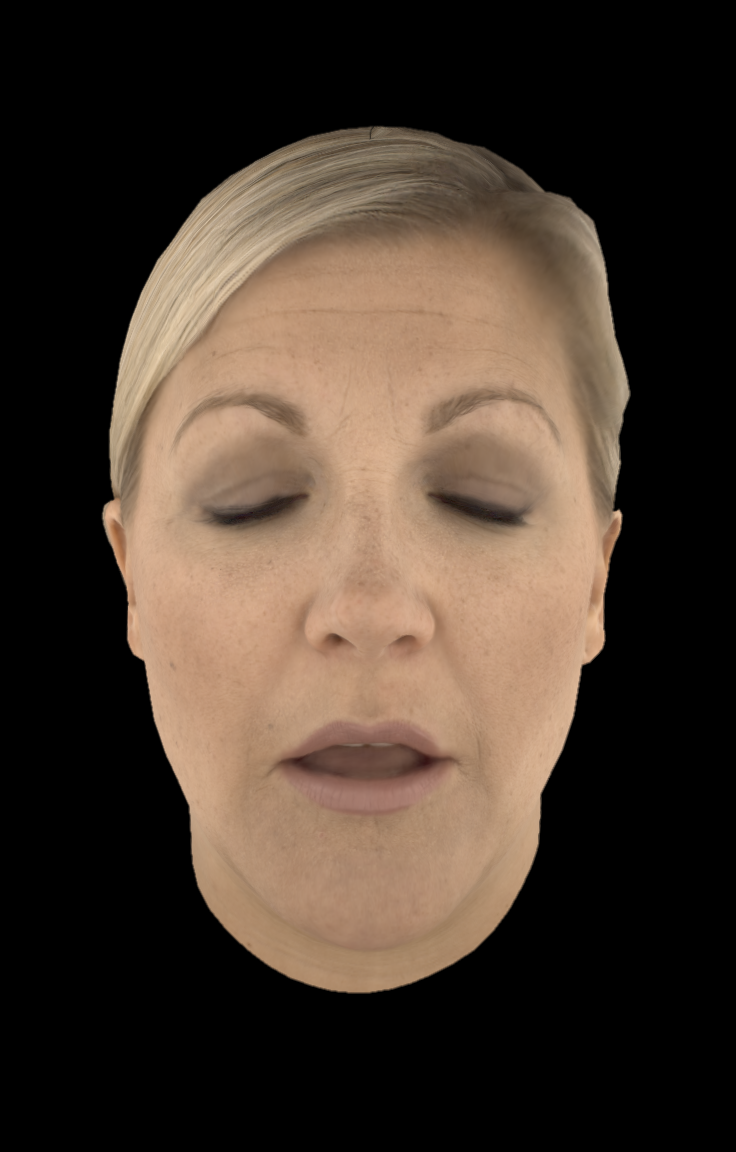} \\

\includegraphics[width=20.5mm]{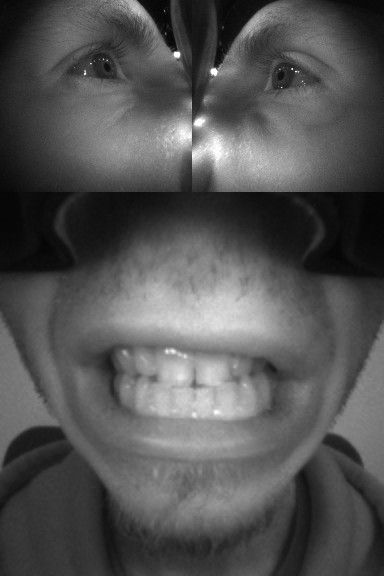} &
  \includegraphics[trim=90 100 100 70,clip, width=17mm]{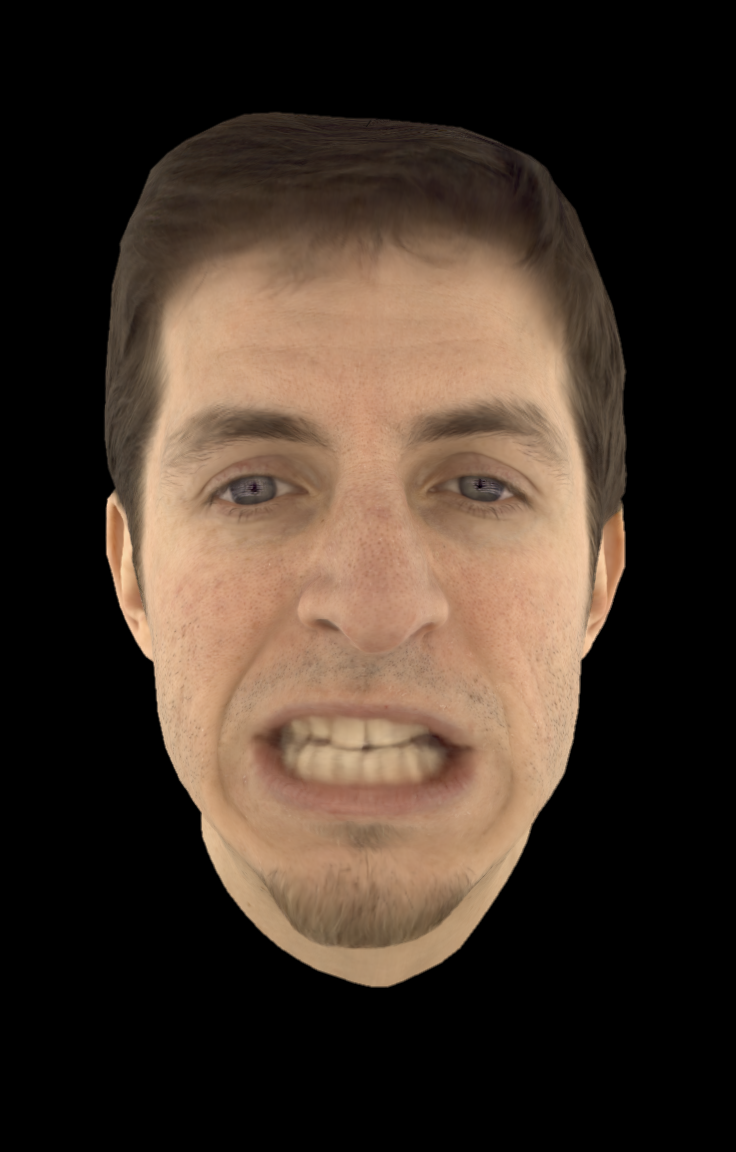} & 
 \includegraphics[trim=90 100 100 70,clip, width=17mm]{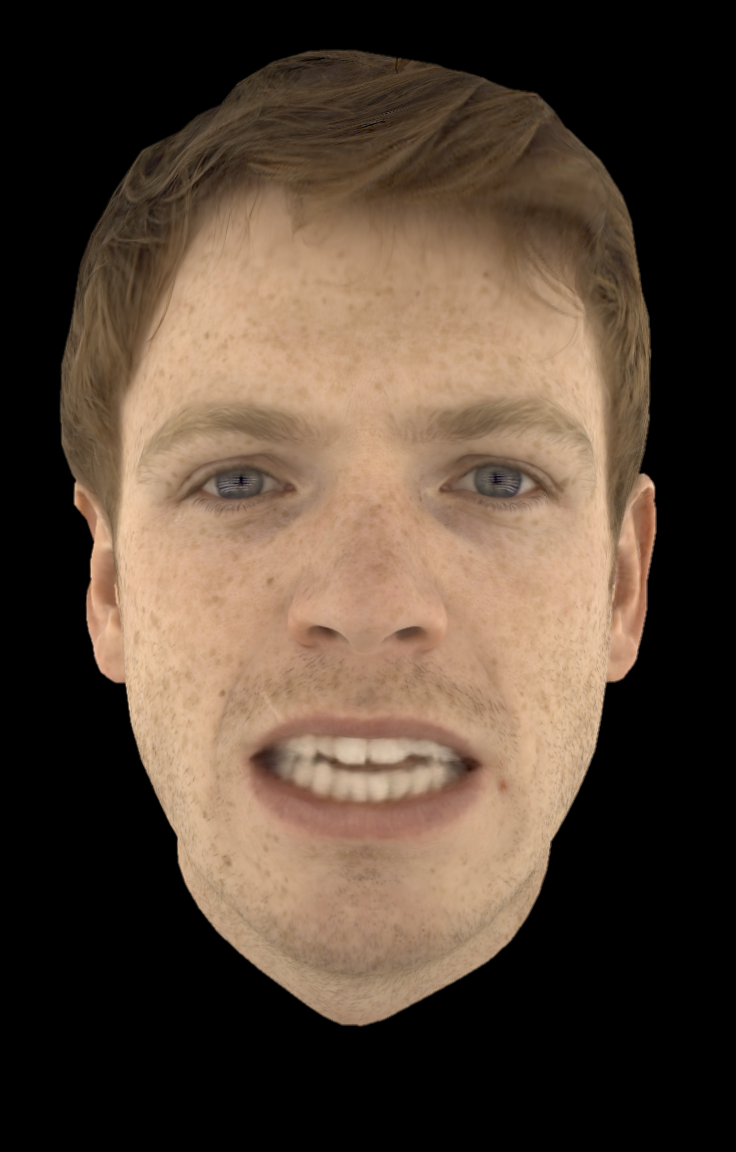} & 
 \includegraphics[trim=90 100 100 70,clip, width=17mm]{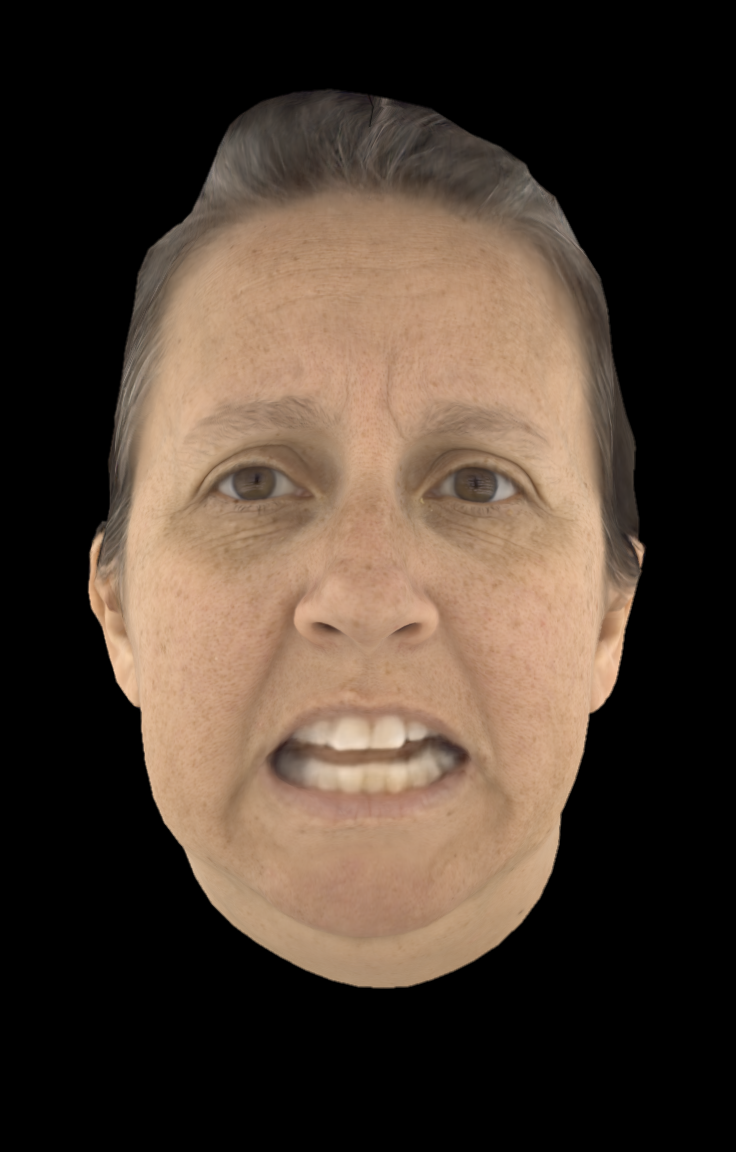} & 
 \includegraphics[trim=90 100 100 70,clip, width=17mm]{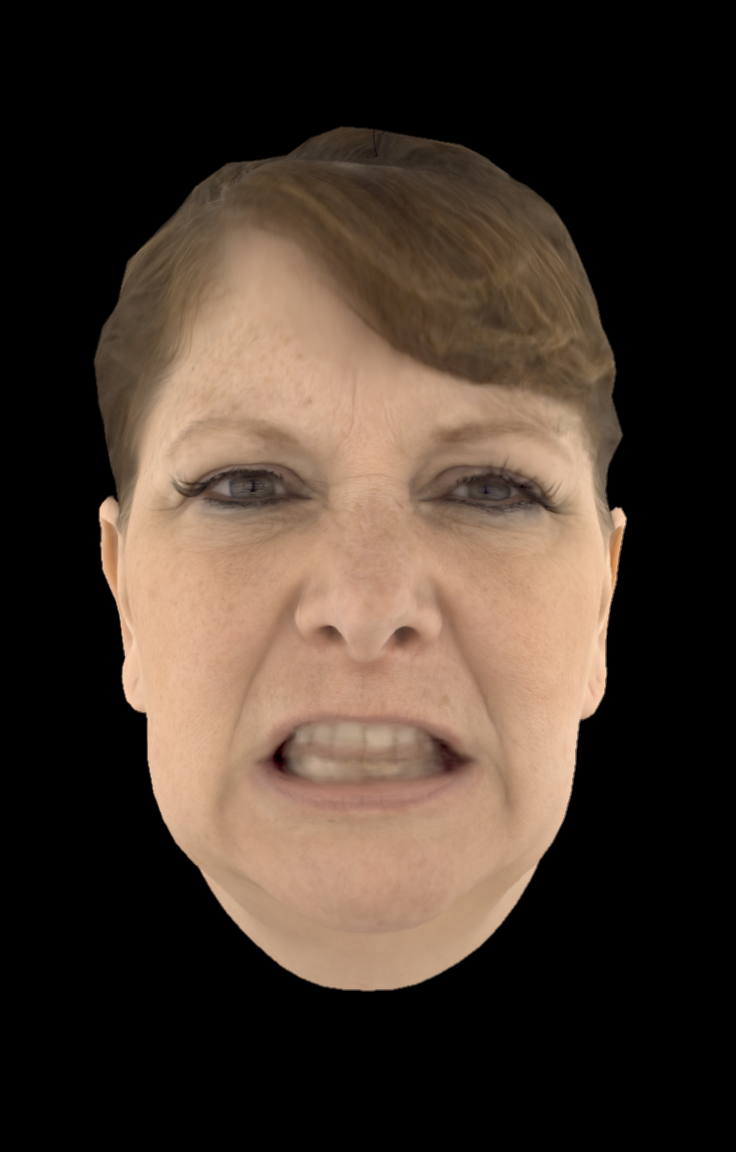} & 
 \includegraphics[trim=90 100 100 70,clip, width=17mm]{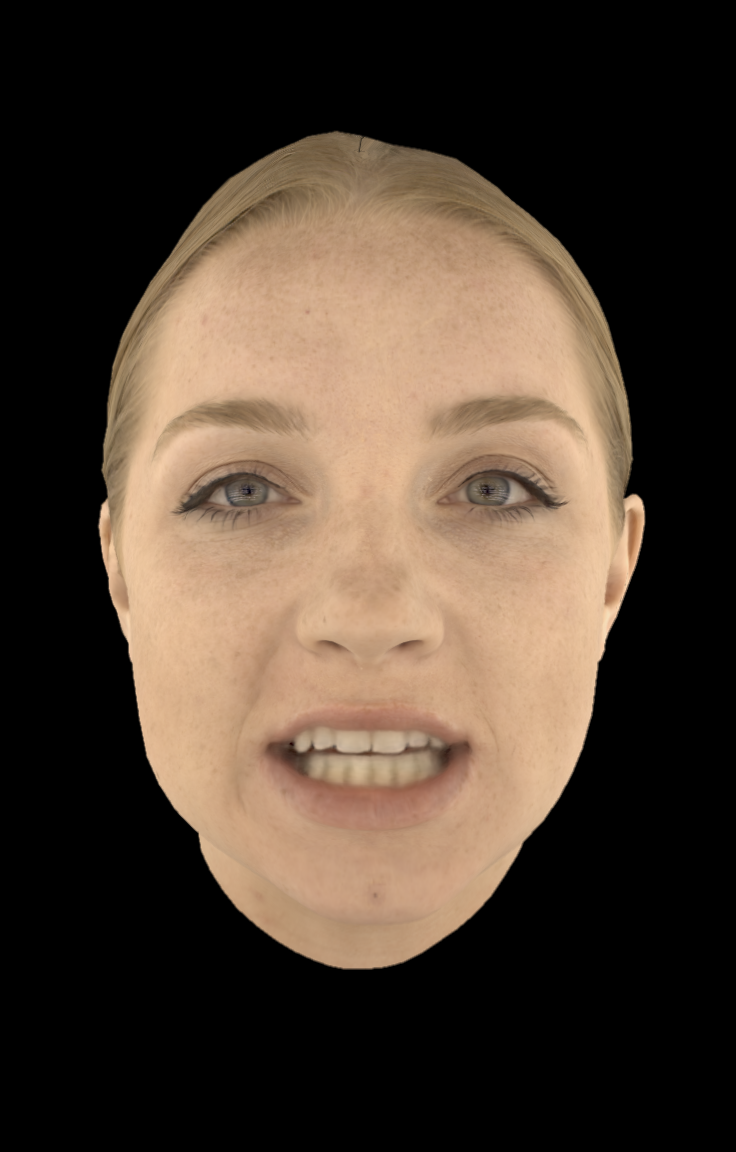} & 
 \includegraphics[trim=90 100 100 70,clip, width=17mm]{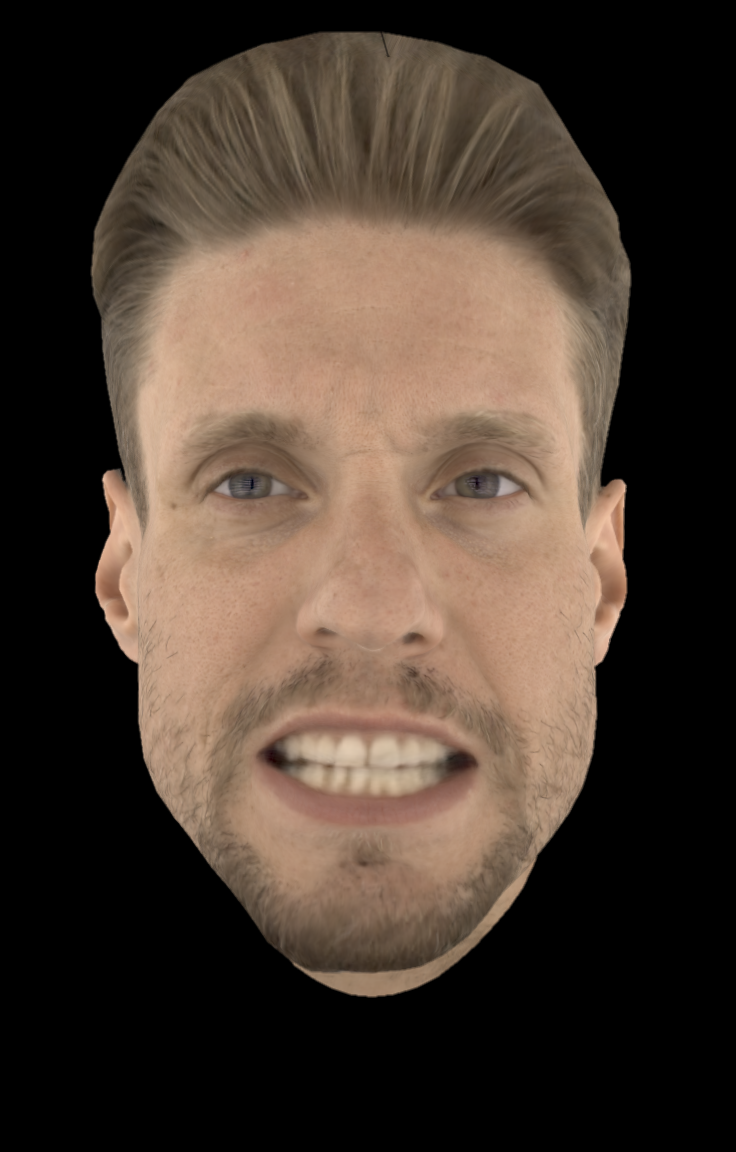} & 
 \includegraphics[trim=90 100 100 70,clip, width=17mm]{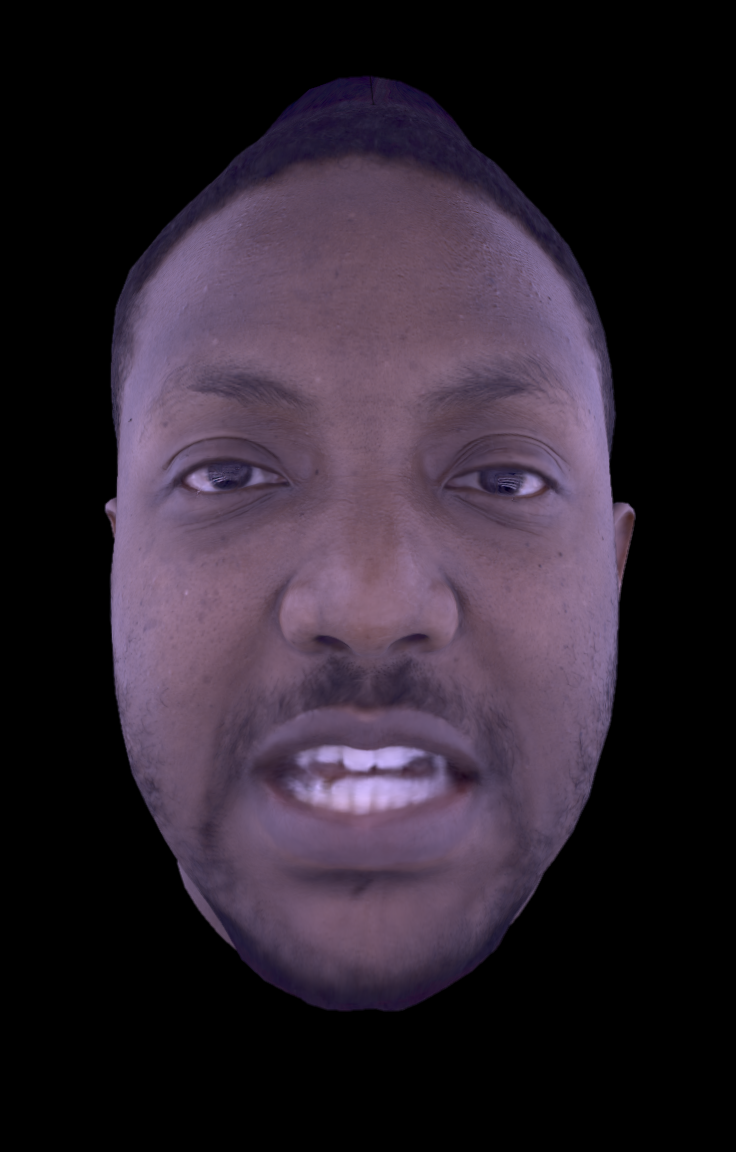} & 
 \includegraphics[trim=90 100 100 70,clip, width=17mm]{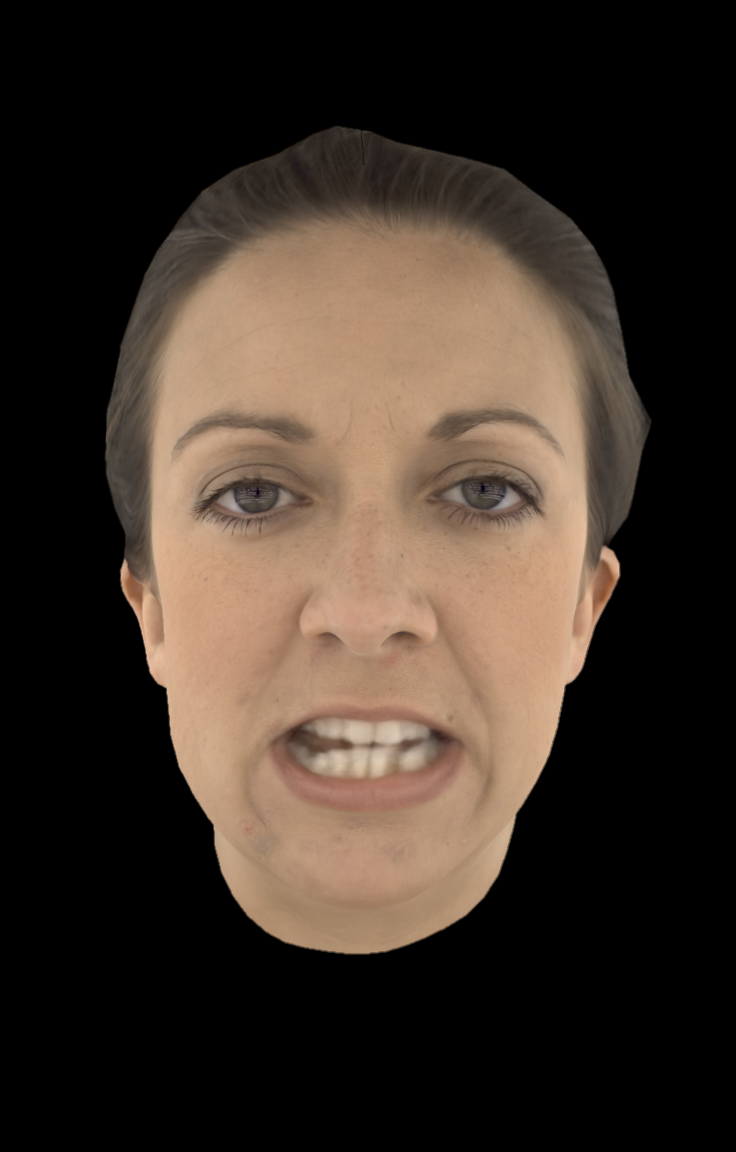} & 
 \includegraphics[trim=90 100 100 70,clip, width=17mm]{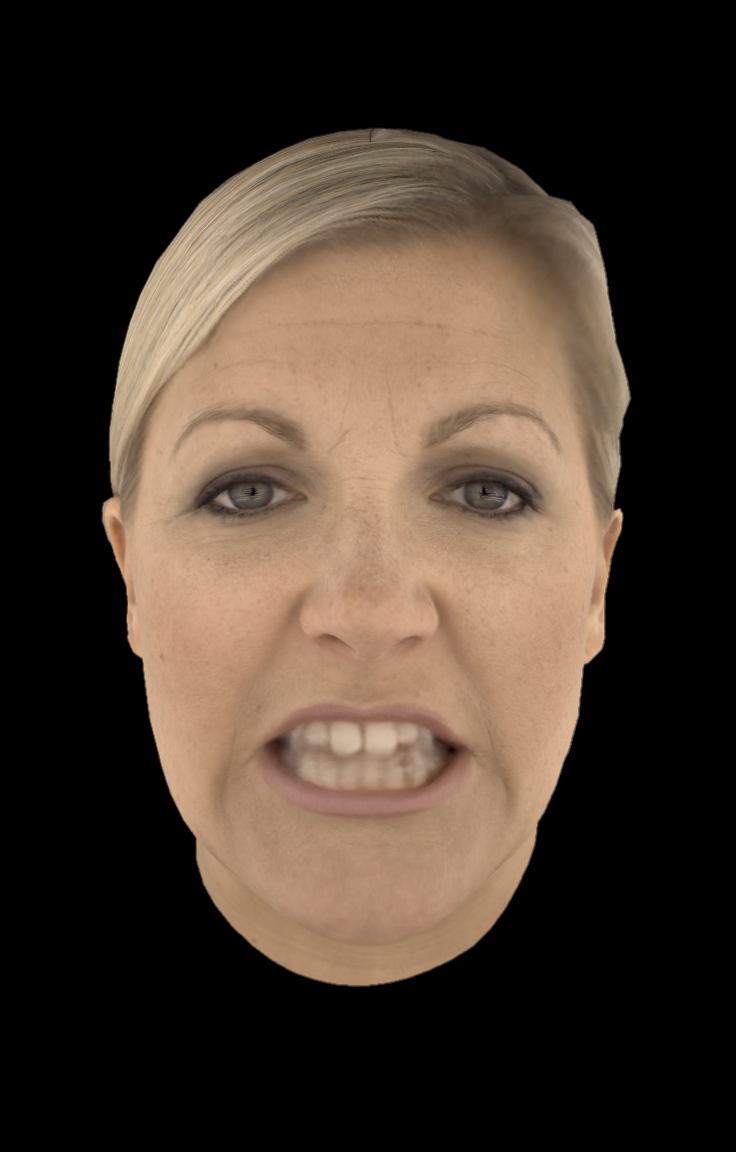} 
 
\end{tabular}
\end{center}
\figvspaceB
\caption{The examples of synchronized expressions of training subjects by using the same input HMC images. }
\label{fig:test_aligned_expressions}
\figvspace
\end{figure*}

\SubSection{ Unsupervised Expression Correspondence}
 The MIA implicitly learns to solve for correspondence across expressions in order to marginalize nuisance parameters (e.g.,  lighting). It naturally discovers that the best way to encode HMC multi-identity data is finding a latent space that only contains expression information. Fig.~\ref{fig:test_aligned_expressions} illustrates how MIA learns to solve for correspondence across expressions. The first column shows the input HMC images and the second column is the CA of the subject in the first column.  The remaining columns are the CAs of other subjects driven from the HMC images in first column, that is, the same extracted features from HMC images are utilized to estimate (by using the corresponding $\textbf{F}_\theta$) a new expression parameter (with the same facial expression meaning), {\bf z}, in the latent space of each of the remaining  subjects. As we can observe, MIA is able to {\em align} the expression across all of the subjects in an unsupervised manner,  and creating a common expression-only space. Please pay attention to the mouth area in the second row of Fig.~\ref{fig:test_aligned_expressions} that shows the same expression with different mouth interior.   

\begin{figure*}[t]
\begin{center}
\setlength\tabcolsep{1.0pt}
\small
\begin{tabular}{ccccccccc}

\includegraphics[width=17.5mm]{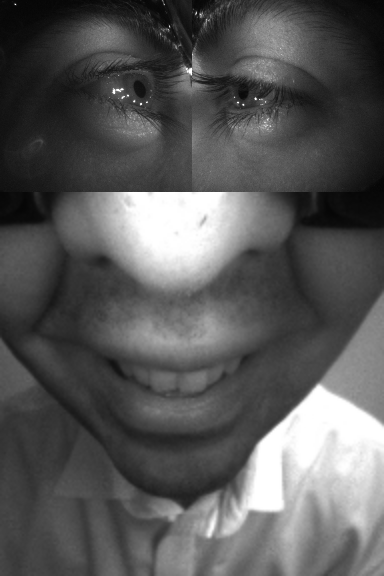} &
 \includegraphics[trim=90 100 100 100,clip, width=15mm]{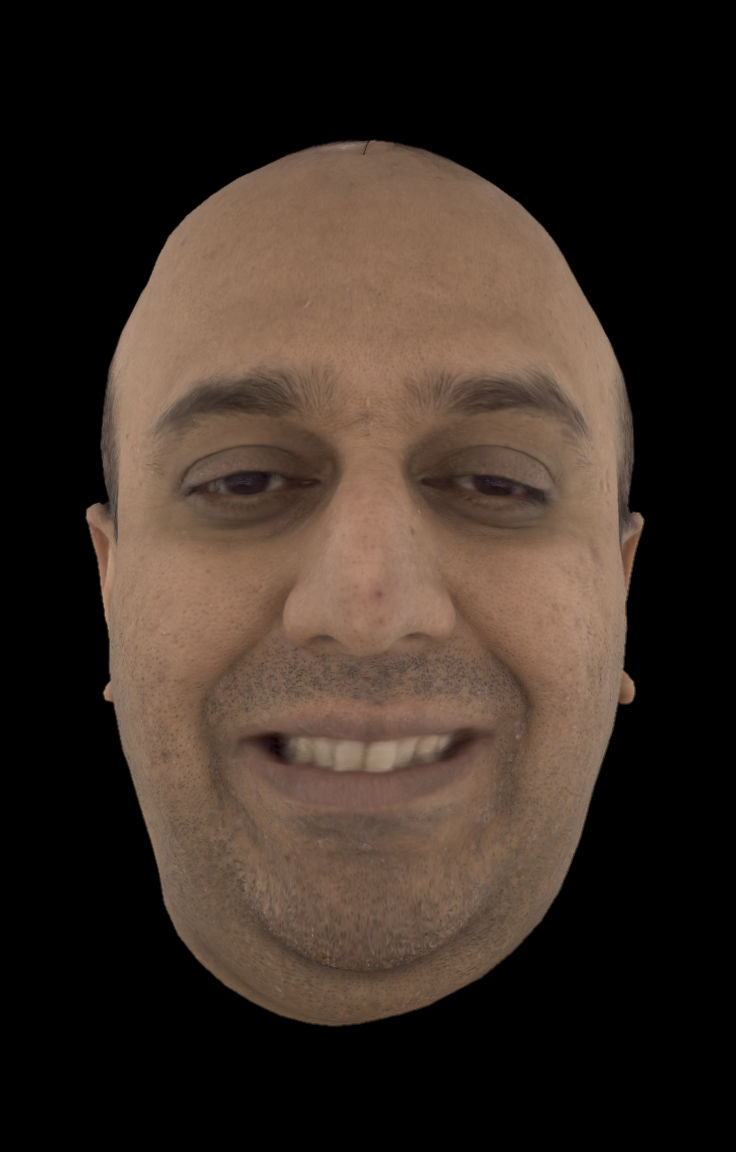} &
 \includegraphics[trim=90 100 100 100,clip, width=15mm]{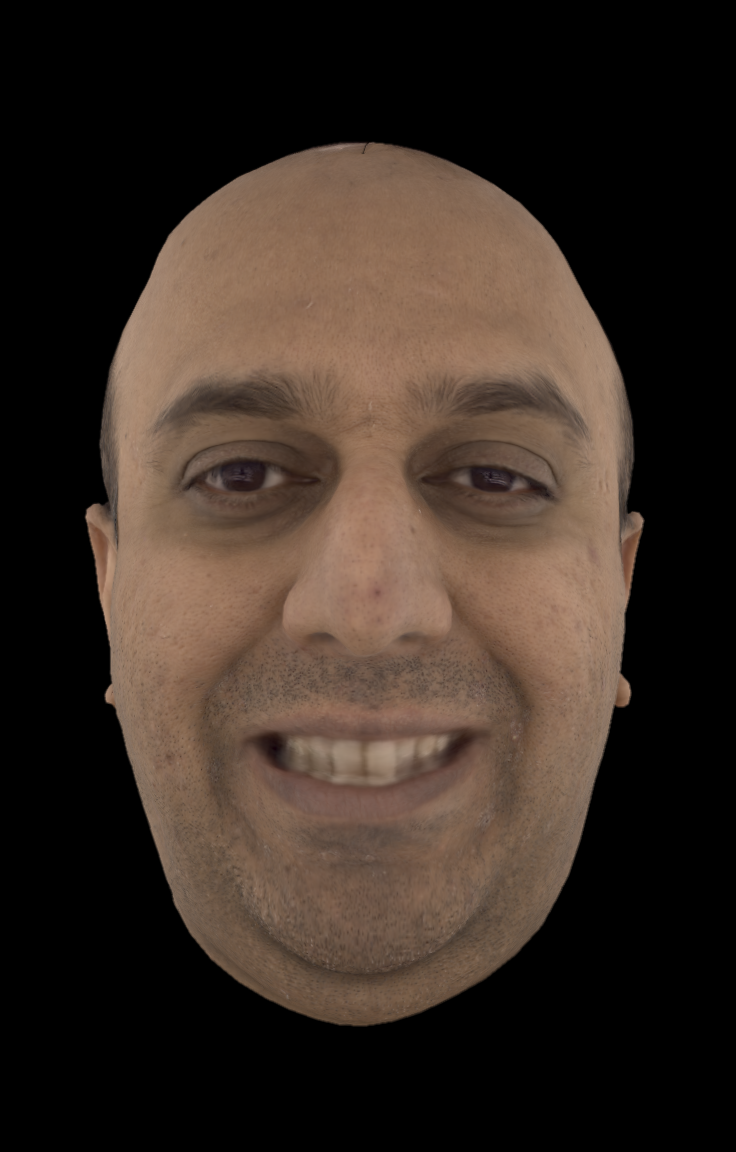} &
 \includegraphics[trim=90 100 100 100,clip, width=15mm]{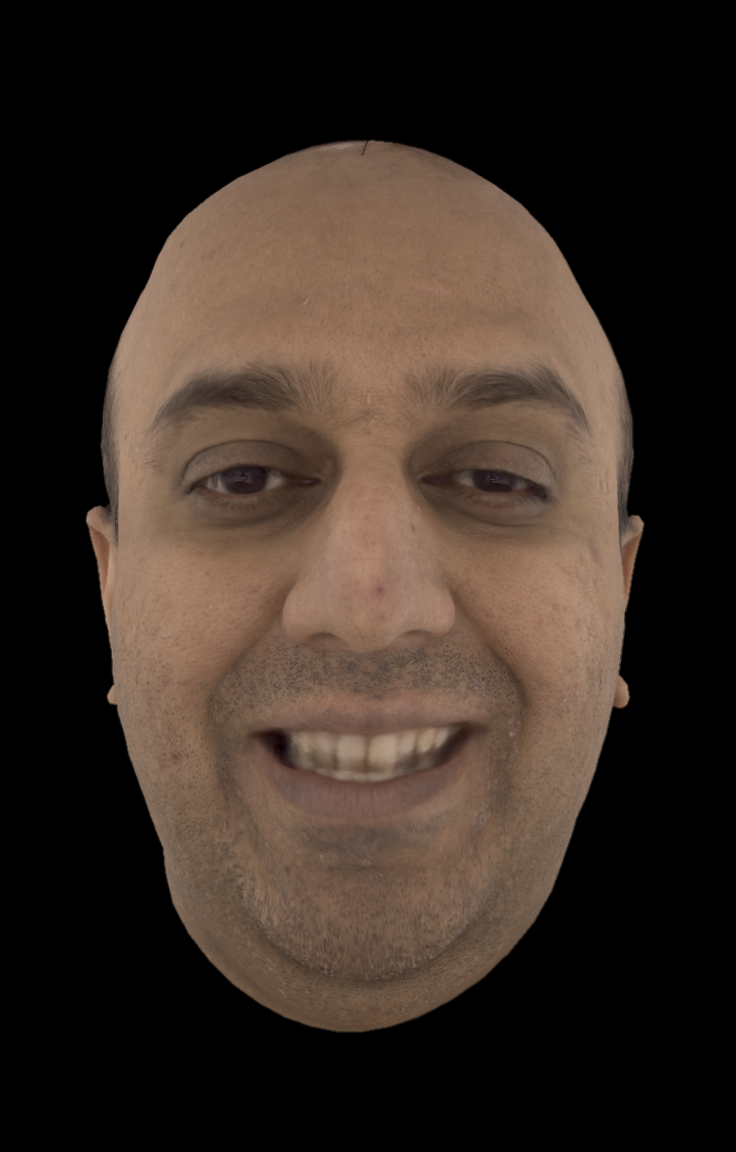} & \ \ \ \ &
 \includegraphics[width=17.5mm]{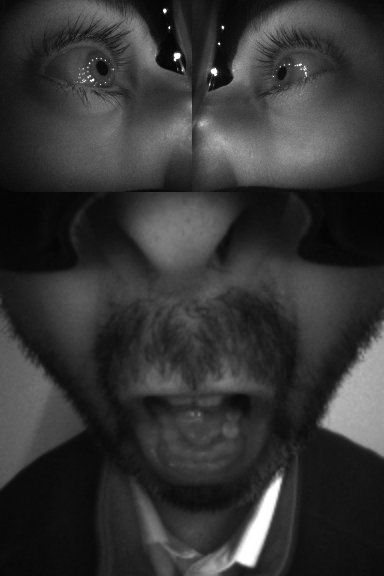} &
 \includegraphics[trim=90 100 100 100,clip, width=15mm]{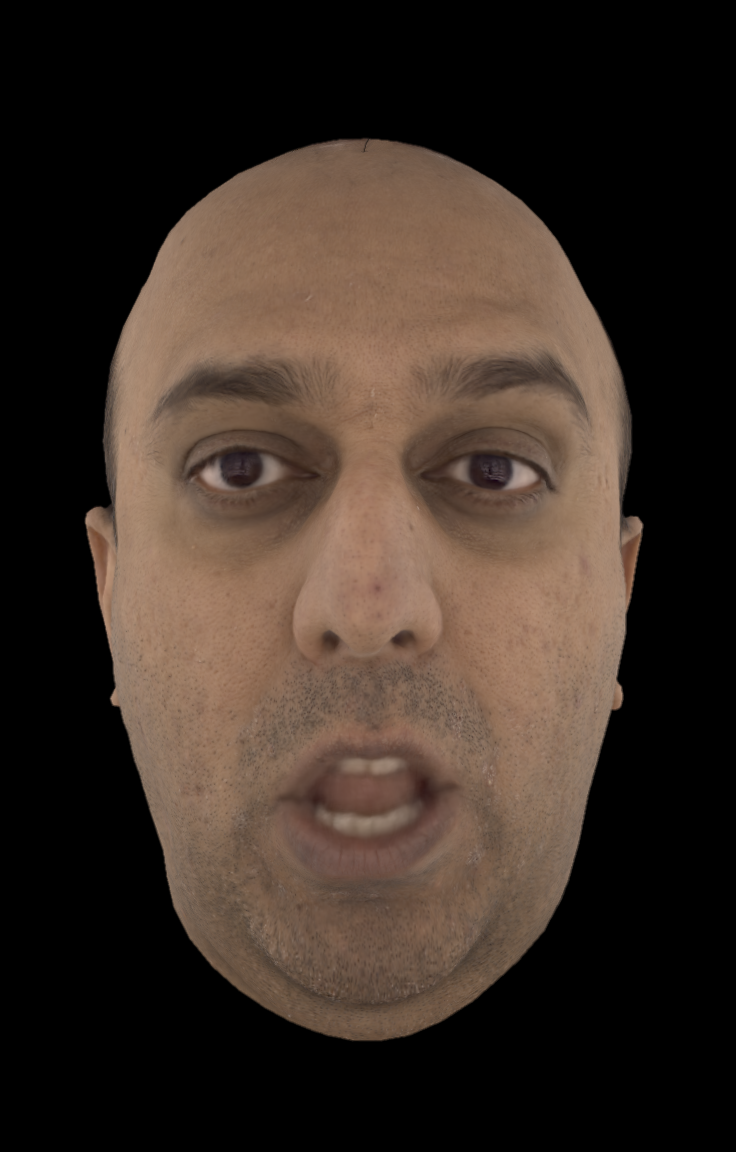} &
 \includegraphics[trim=90 100 100 100,clip, width=15mm]{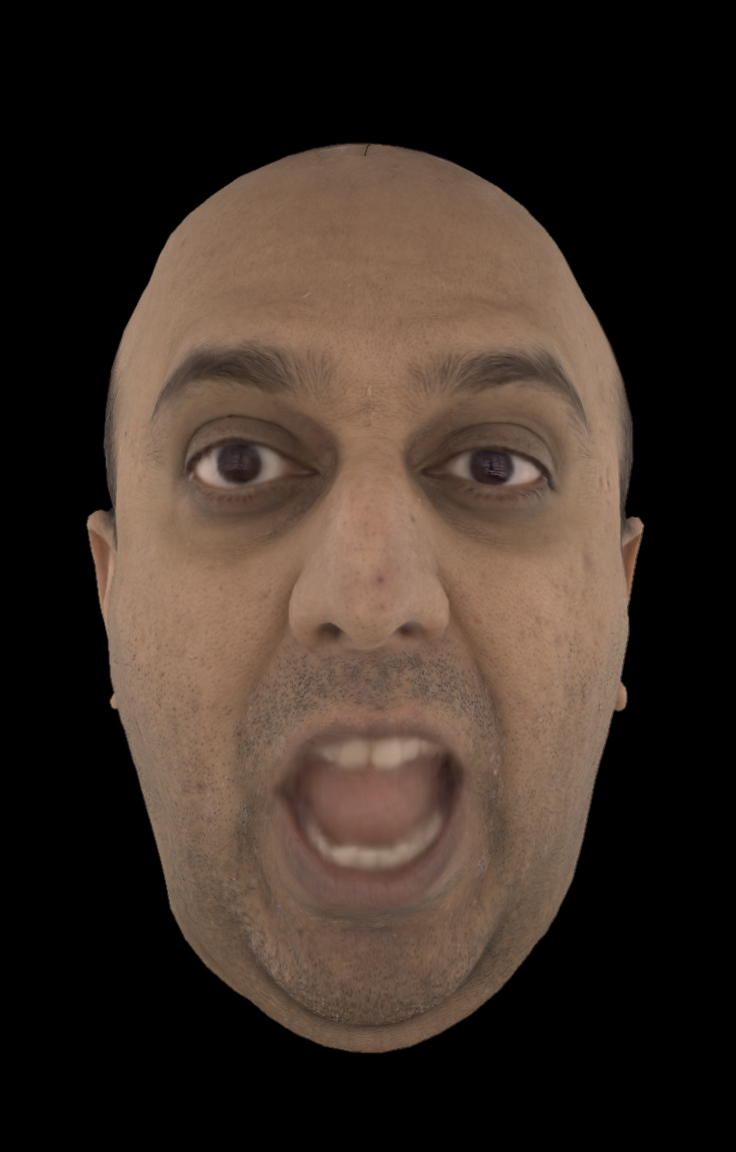} &
 \includegraphics[trim=90 100 100 100,clip, width=15mm]{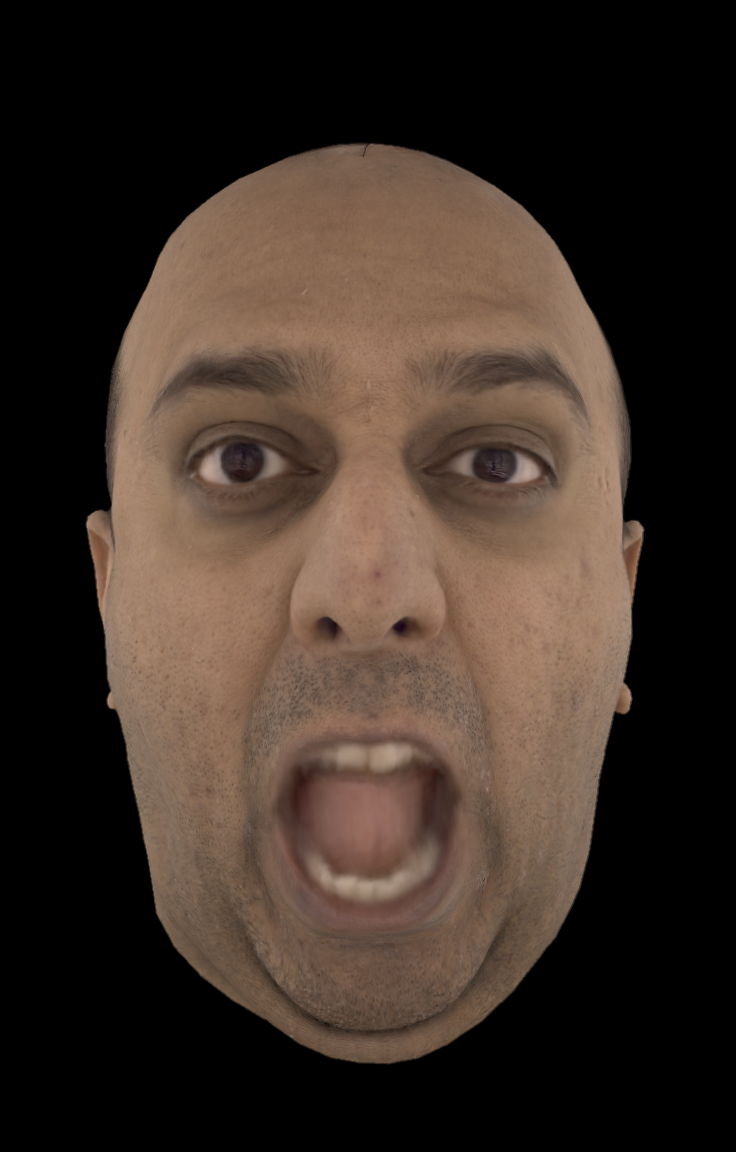}\\

 \includegraphics[width=17.5mm]{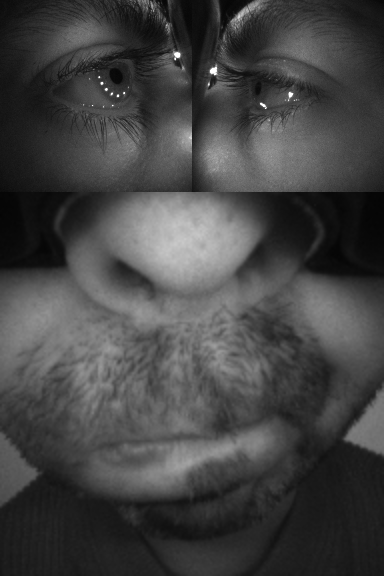} &
 \includegraphics[trim=90 100 100 100,clip, width=15mm]{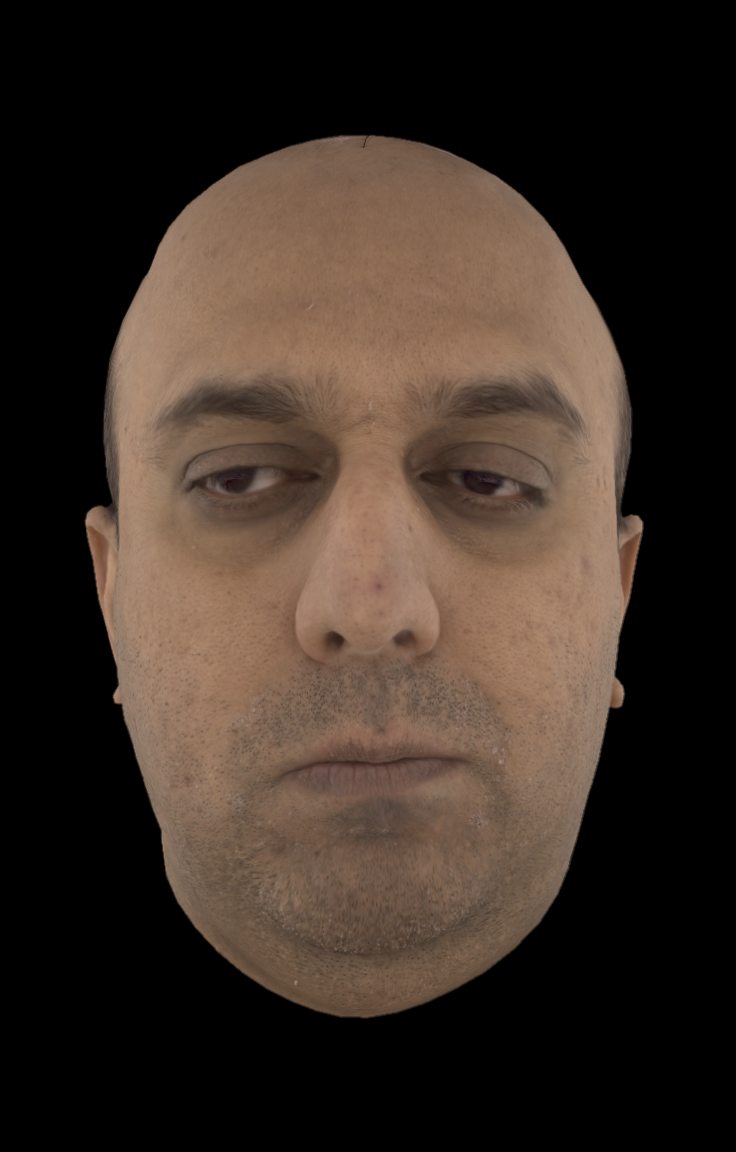} &
 \includegraphics[trim=90 100 100 100,clip, width=15mm]{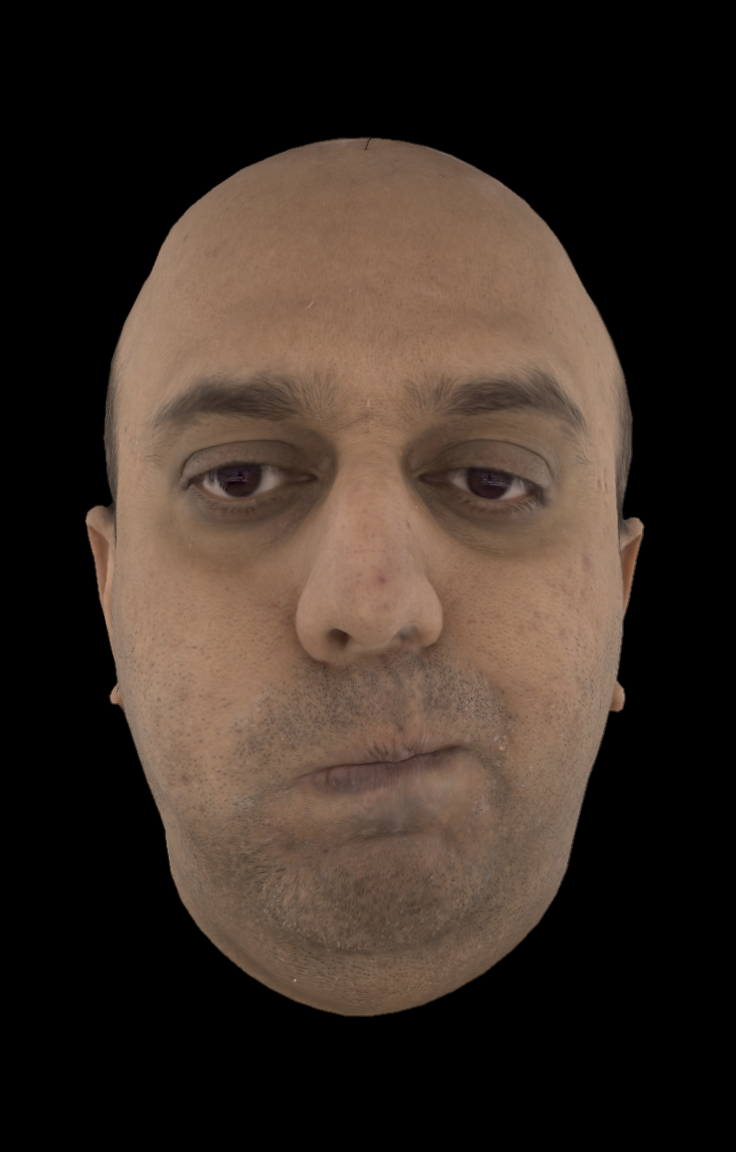} &
 \includegraphics[trim=90 100 100 100,clip, width=15mm]{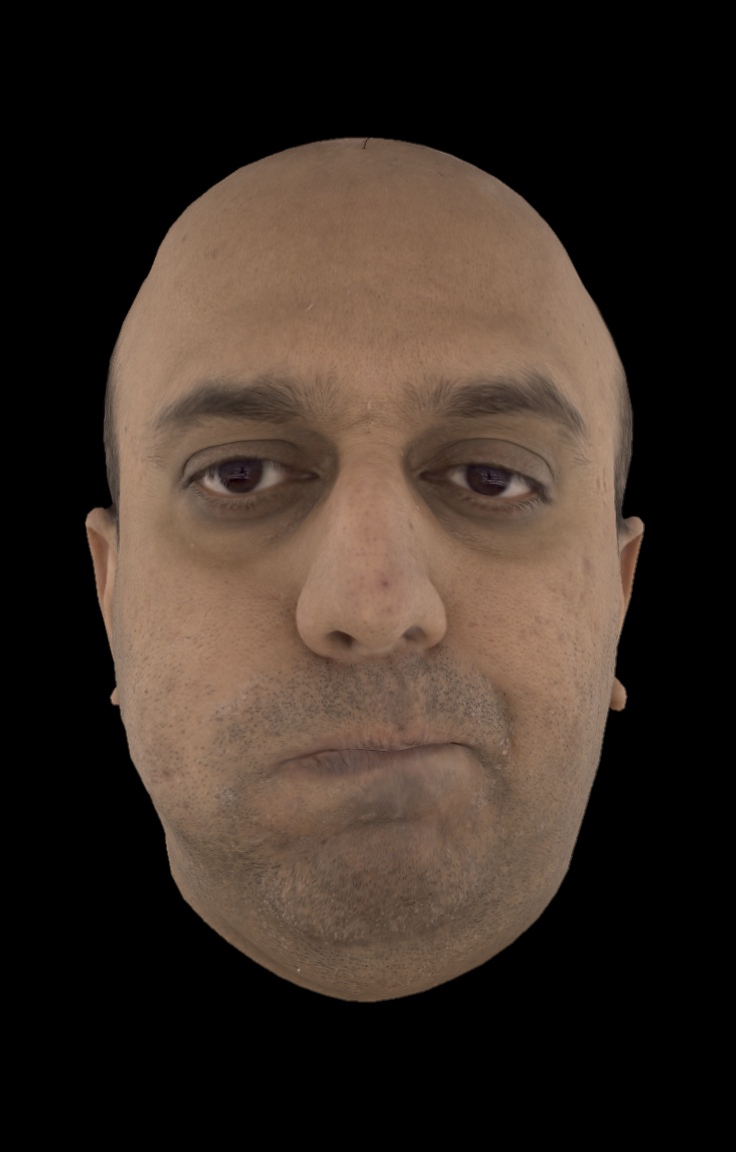} & \ \ \ \ &
 \includegraphics[width=17.5mm]{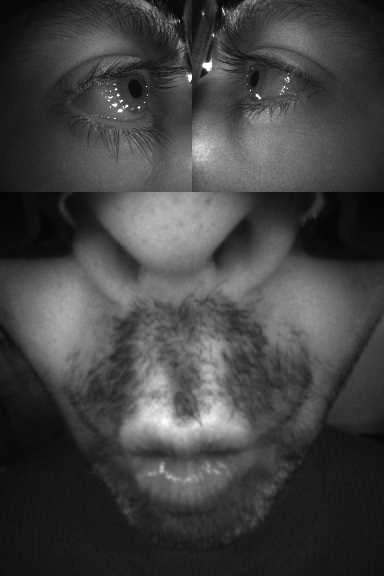} &
 \includegraphics[trim=90 100 100 100,clip, width=15mm]{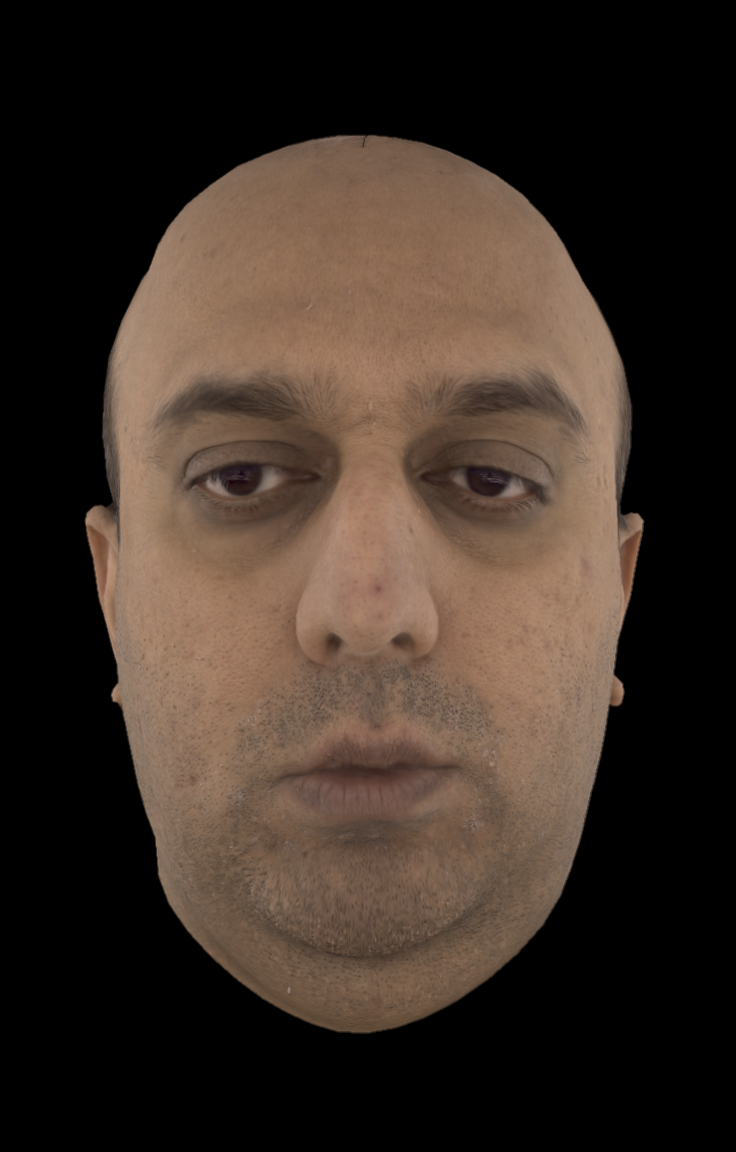} &
 \includegraphics[trim=90 100 100 100,clip, width=15mm]{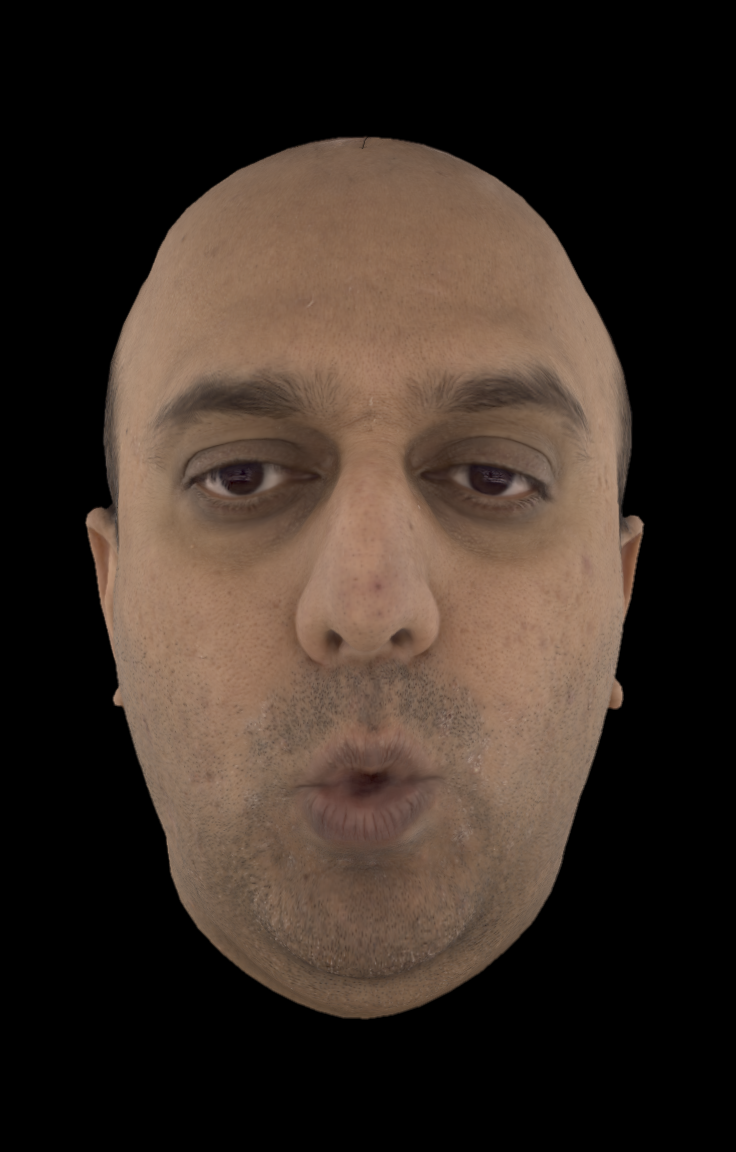} &
 \includegraphics[trim=90 100 100 100,clip, width=15mm]{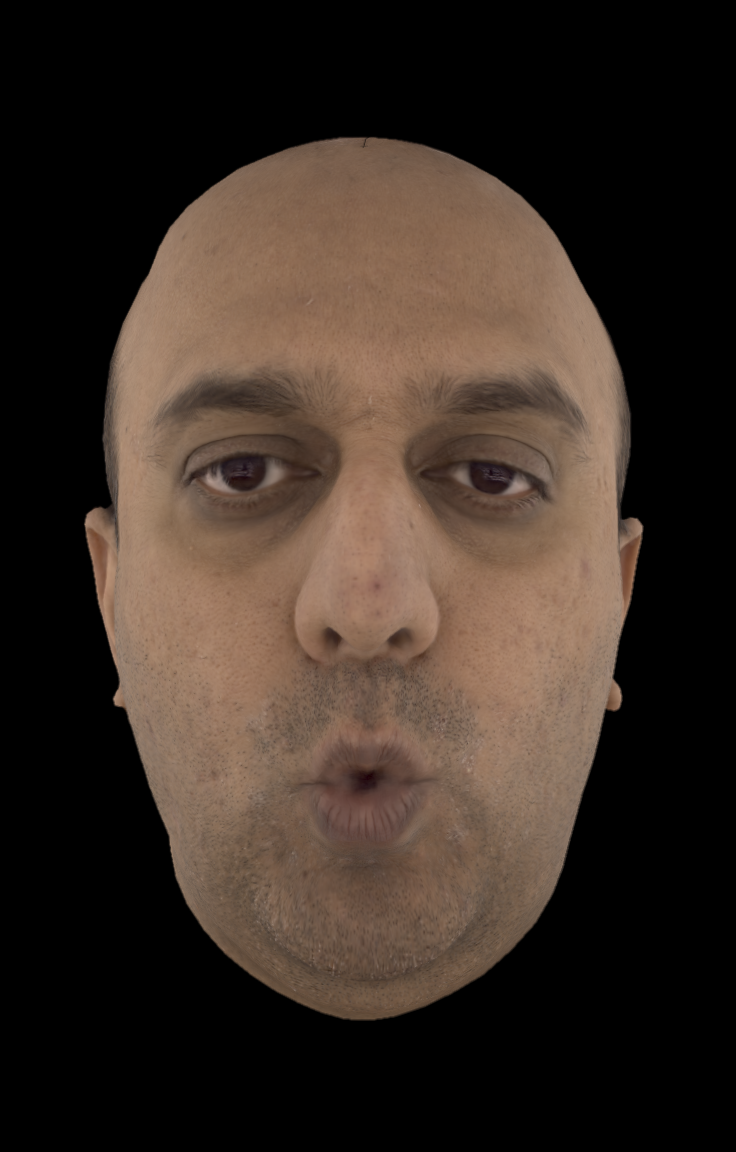}\\

 \toprule
 
 \includegraphics[width=17.5mm]{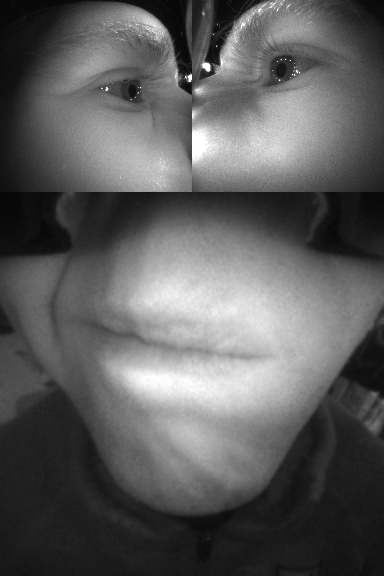} &
 \includegraphics[trim=90 100 100 100,clip, width=15mm]{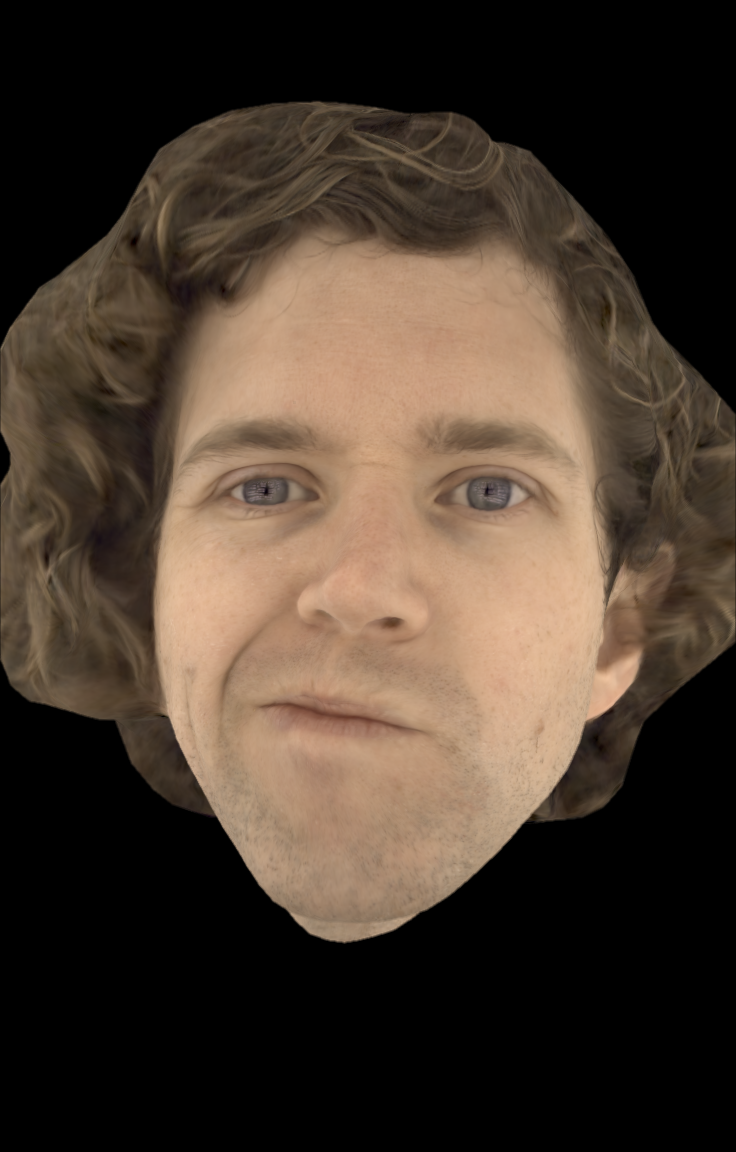} &
 \includegraphics[trim=90 100 100 100,clip, width=15mm]{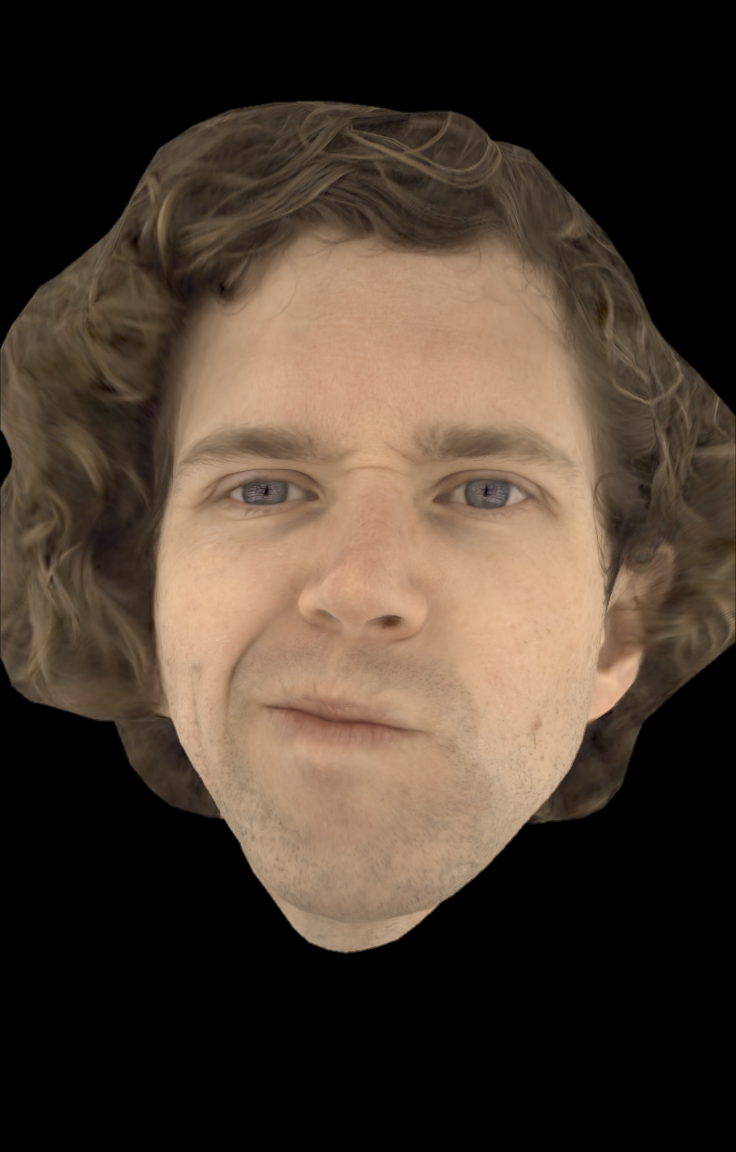} &
 \includegraphics[trim=90 100 100 100,clip, width=15mm]{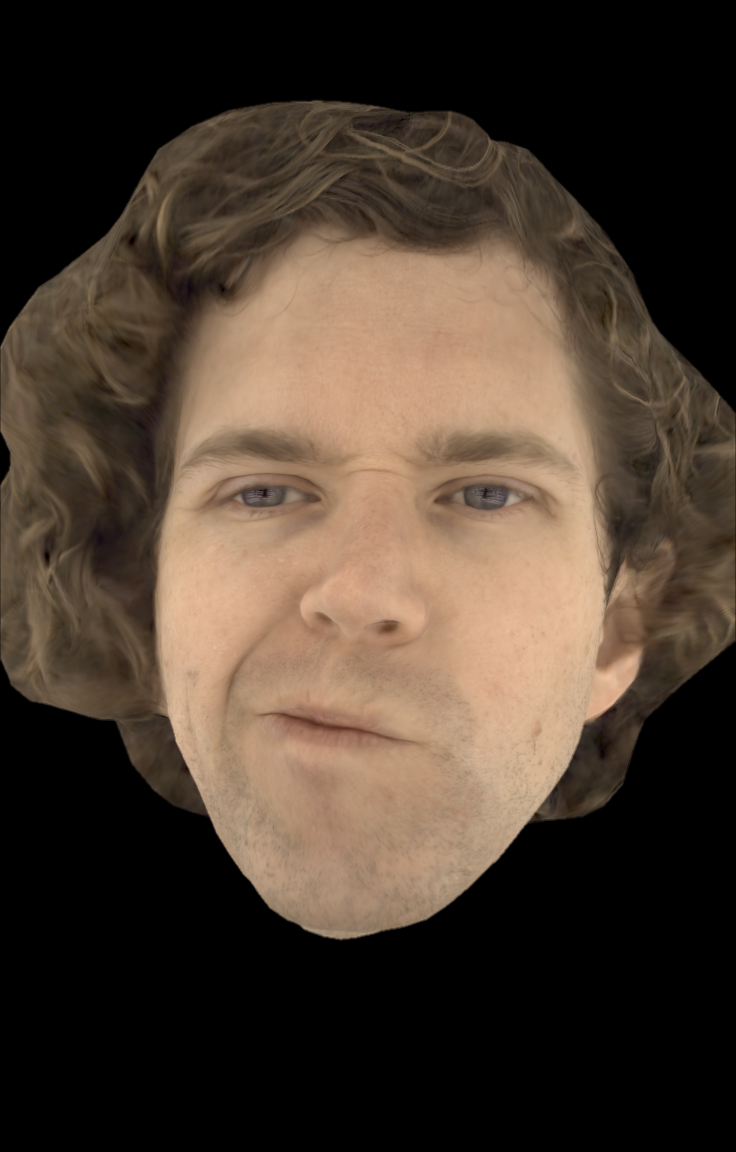} & \ \ \ \ &
 \includegraphics[width=17.5mm]{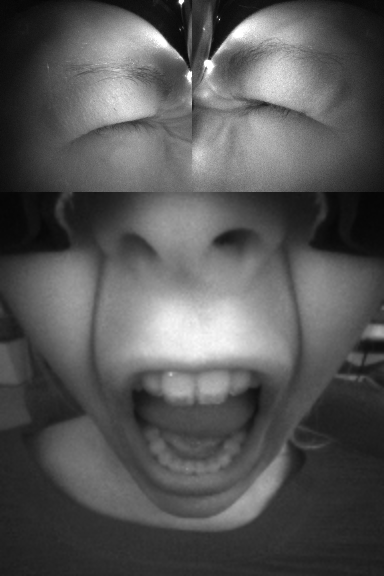} &
 \includegraphics[trim=90 100 100 100,clip, width=15mm]{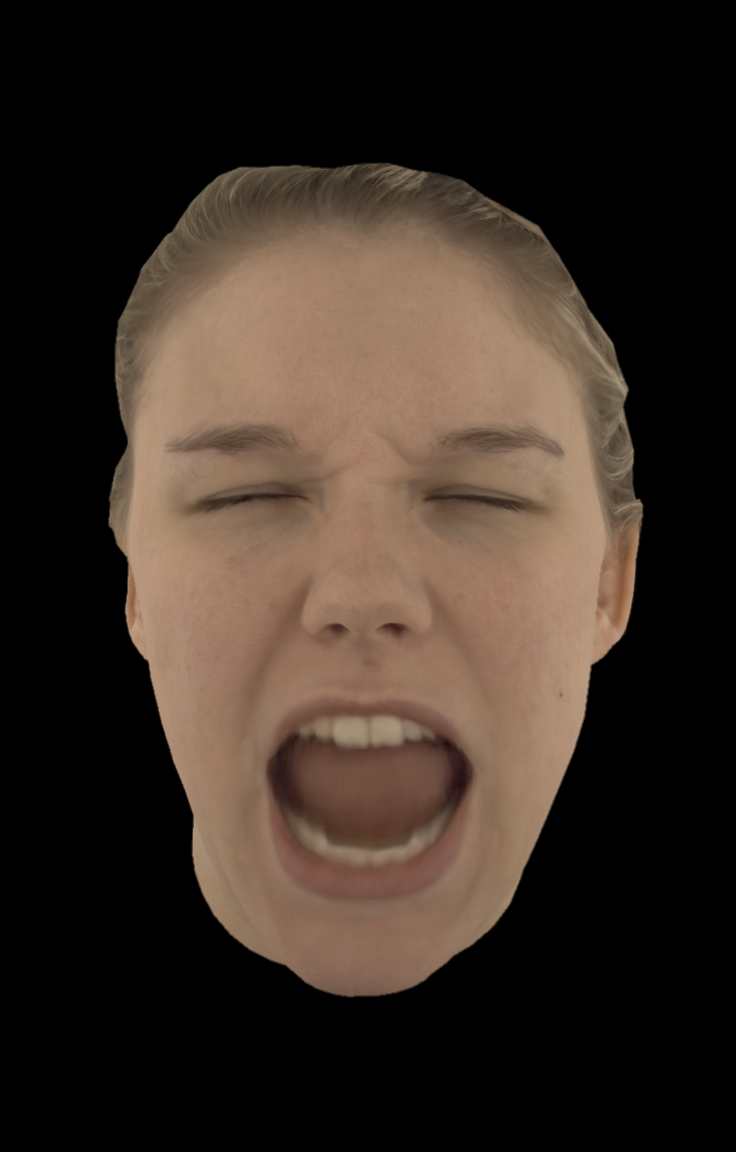} &
 \includegraphics[trim=90 100 100 100,clip, width=15mm]{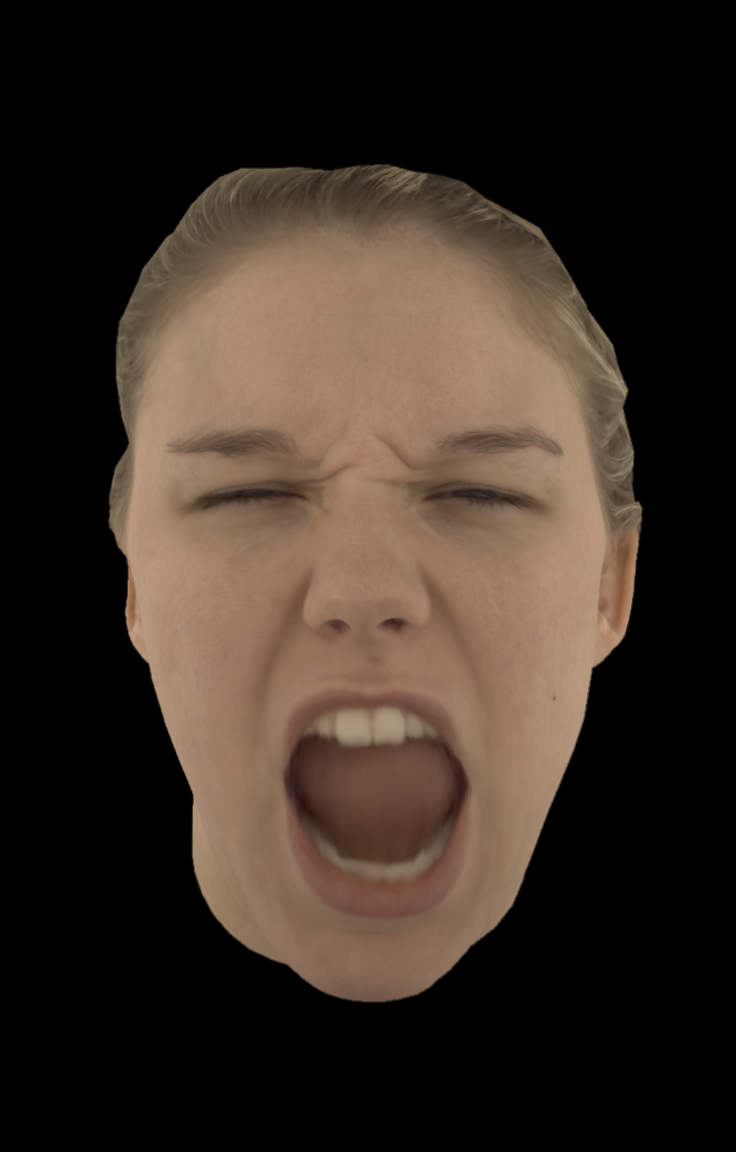} &
 \includegraphics[trim=90 100 100 100,clip, width=15mm]{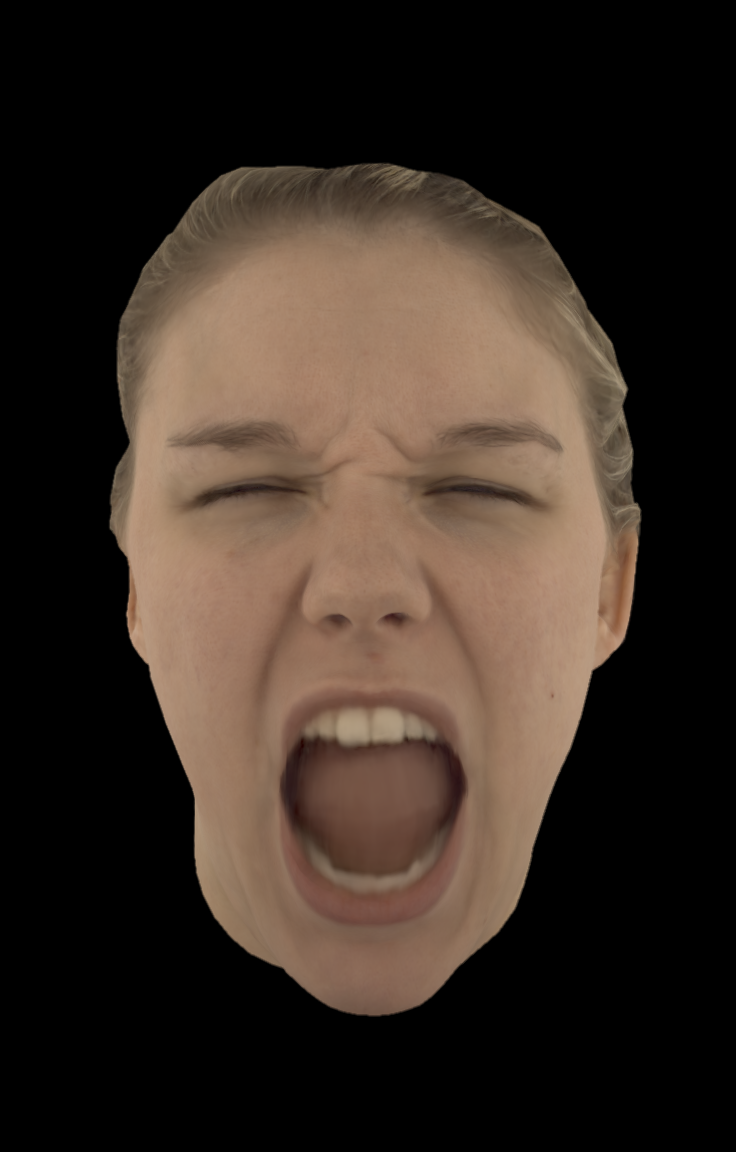}\\
 
 \includegraphics[width=17.5mm]{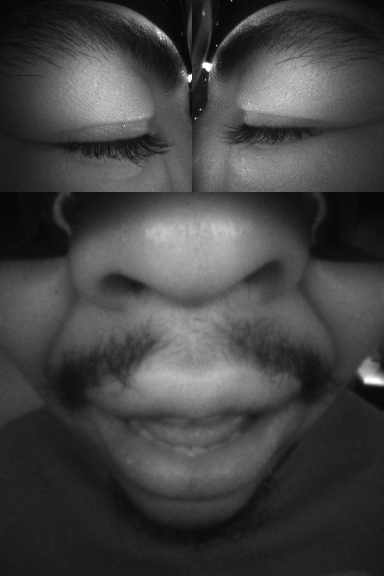} &
 \includegraphics[trim=90 100 100 100,clip, width=15mm]{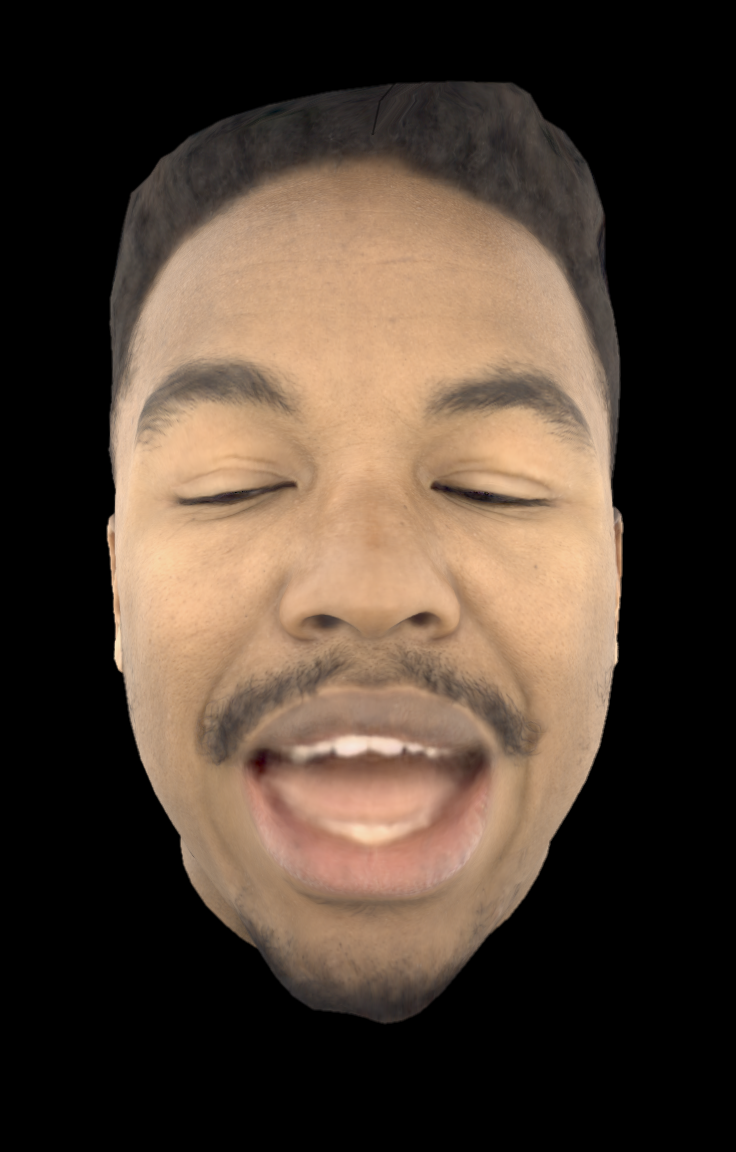} &
 \includegraphics[trim=90 100 100 100,clip, width=15mm]{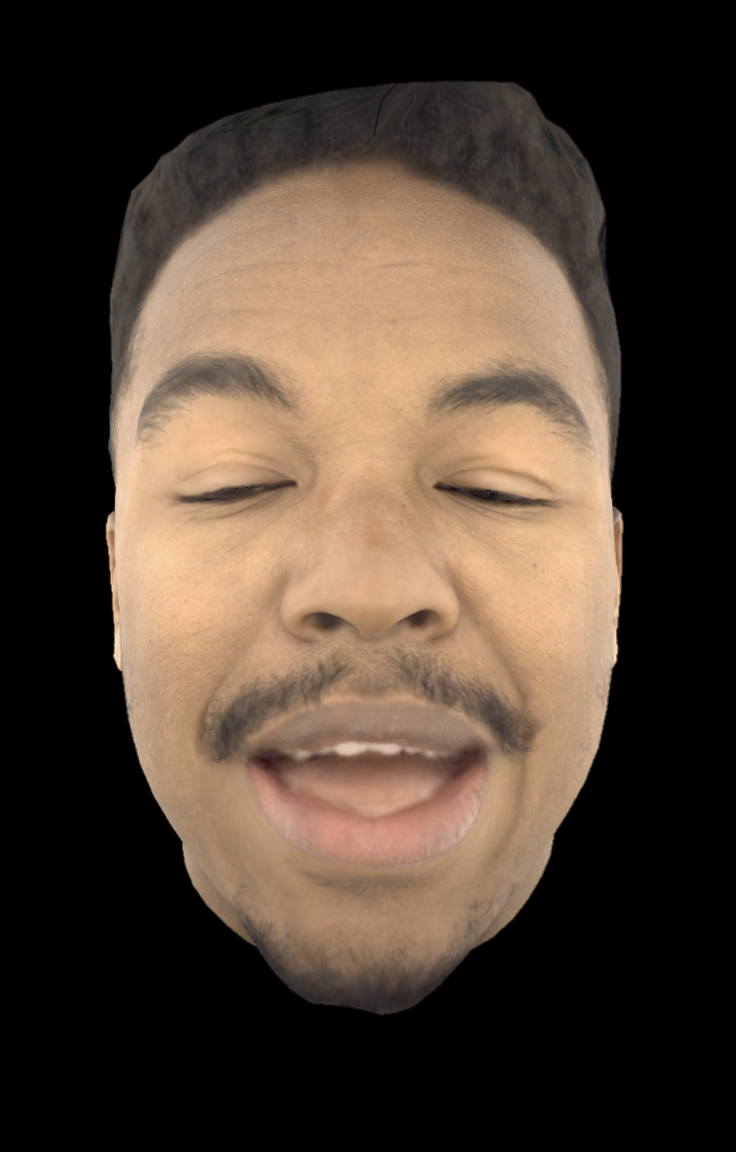} &
 \includegraphics[trim=90 100 100 100,clip, width=15mm]{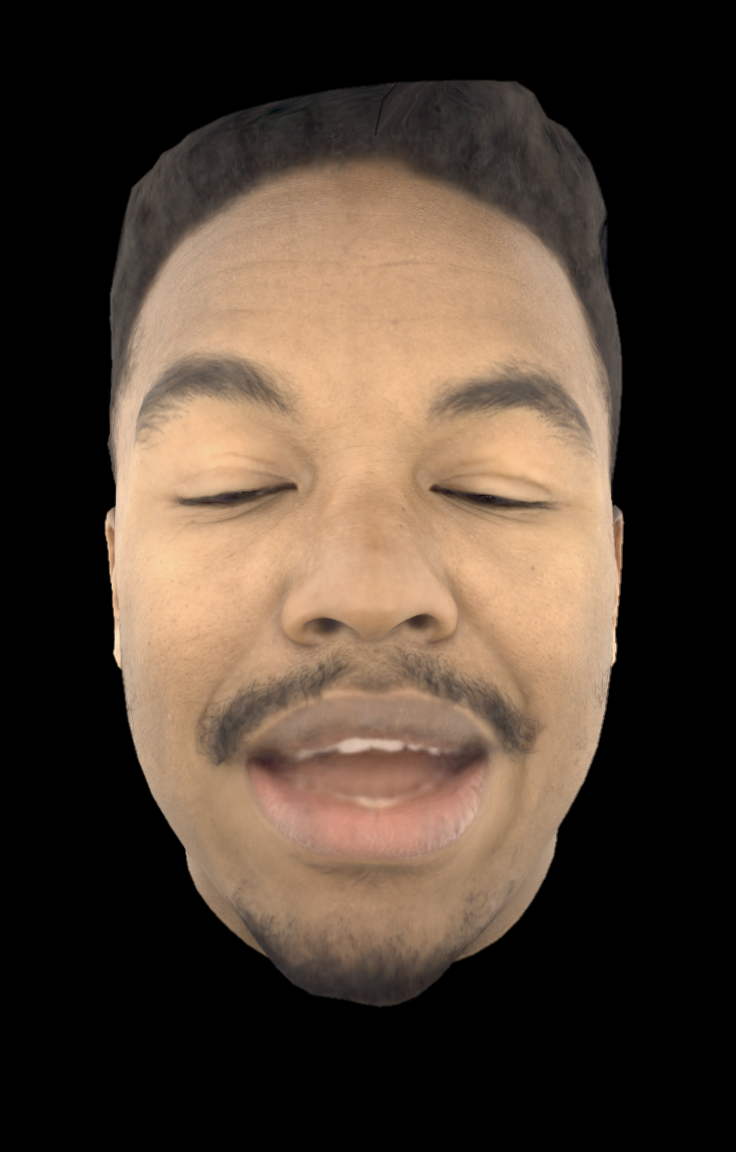} & \ \ \ \ &
 \includegraphics[width=17.5mm]{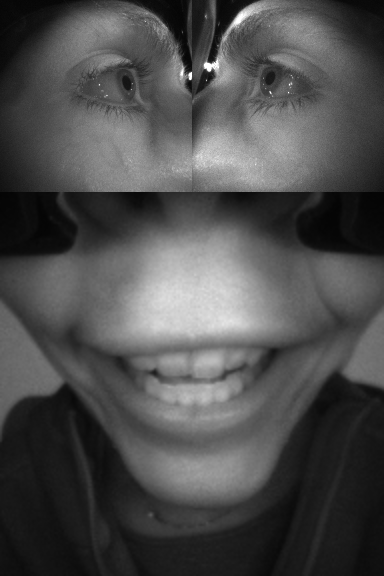} &
 \includegraphics[trim=90 100 100 100,clip, width=15mm]{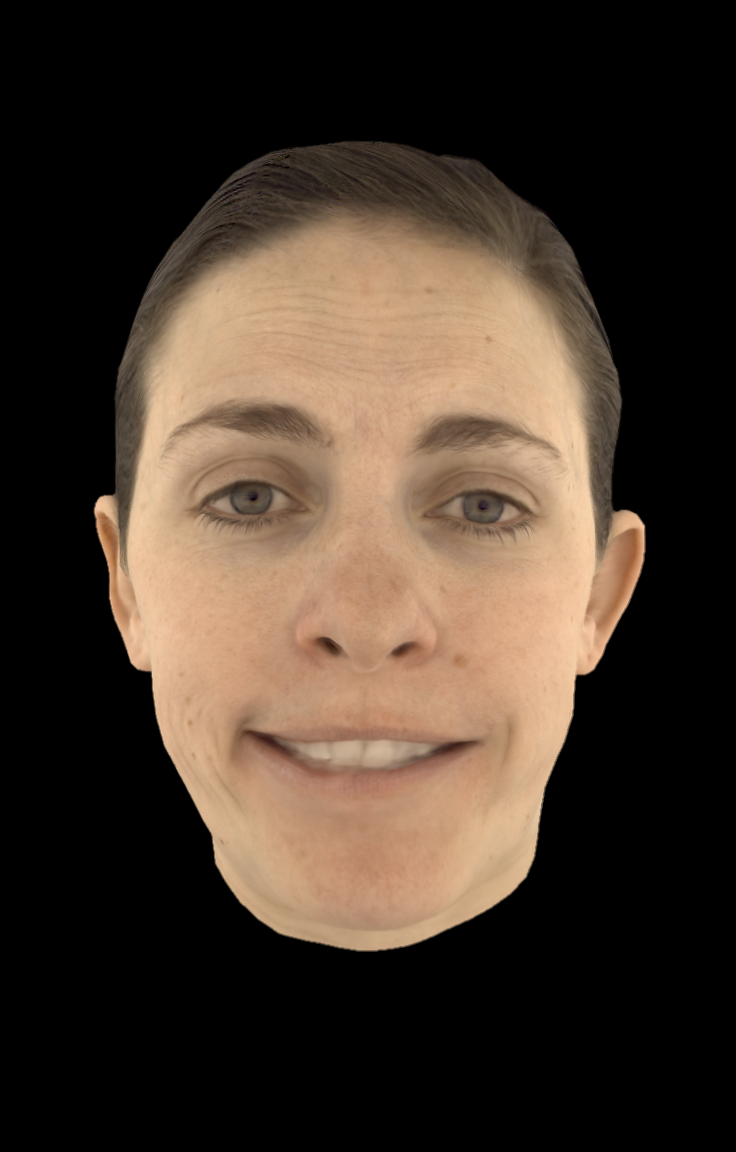} &
 \includegraphics[trim=90 100 100 100,clip, width=15mm]{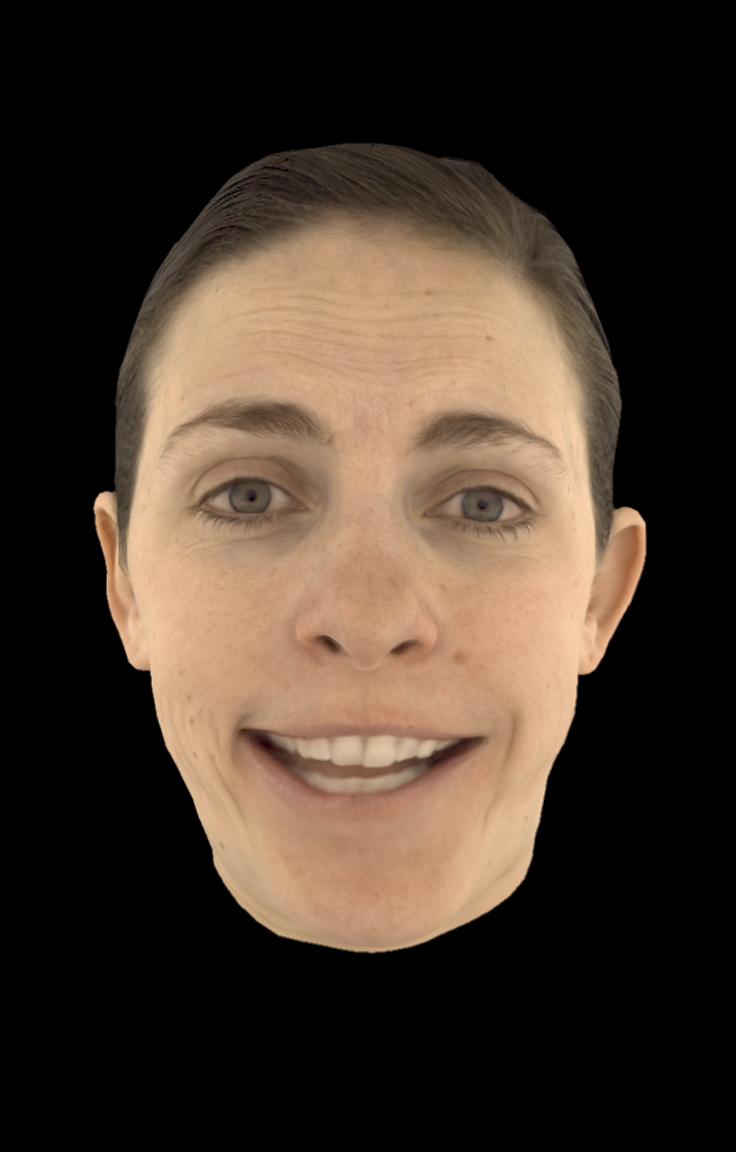} &
 \includegraphics[trim=90 100 100 100,clip, width=15mm]{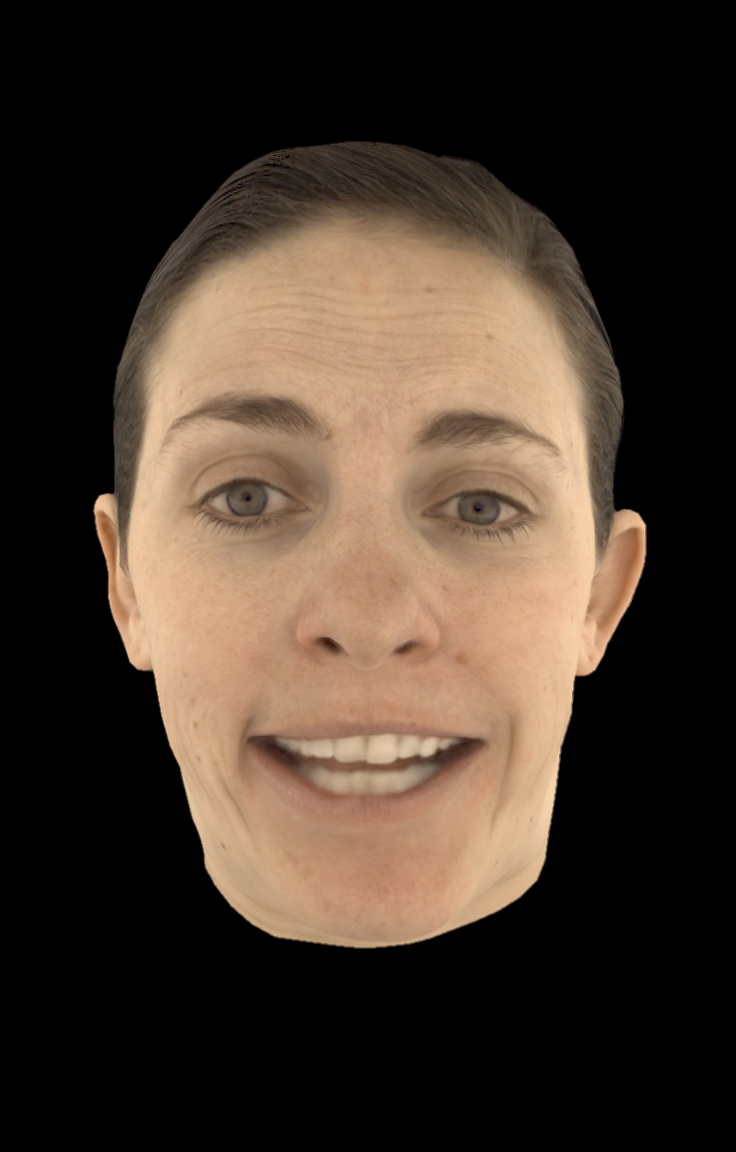}\\
 
 \includegraphics[width=17.5mm]{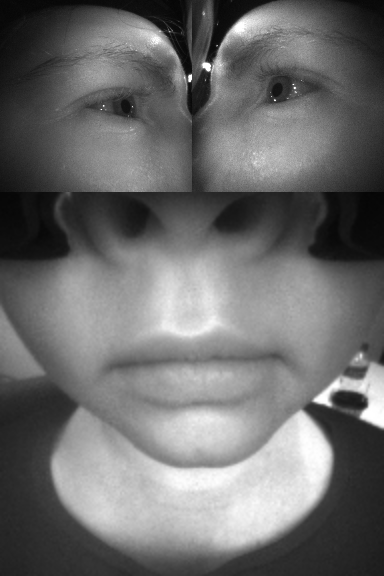} &
 \includegraphics[trim=90 100 100 100,clip, width=15mm]{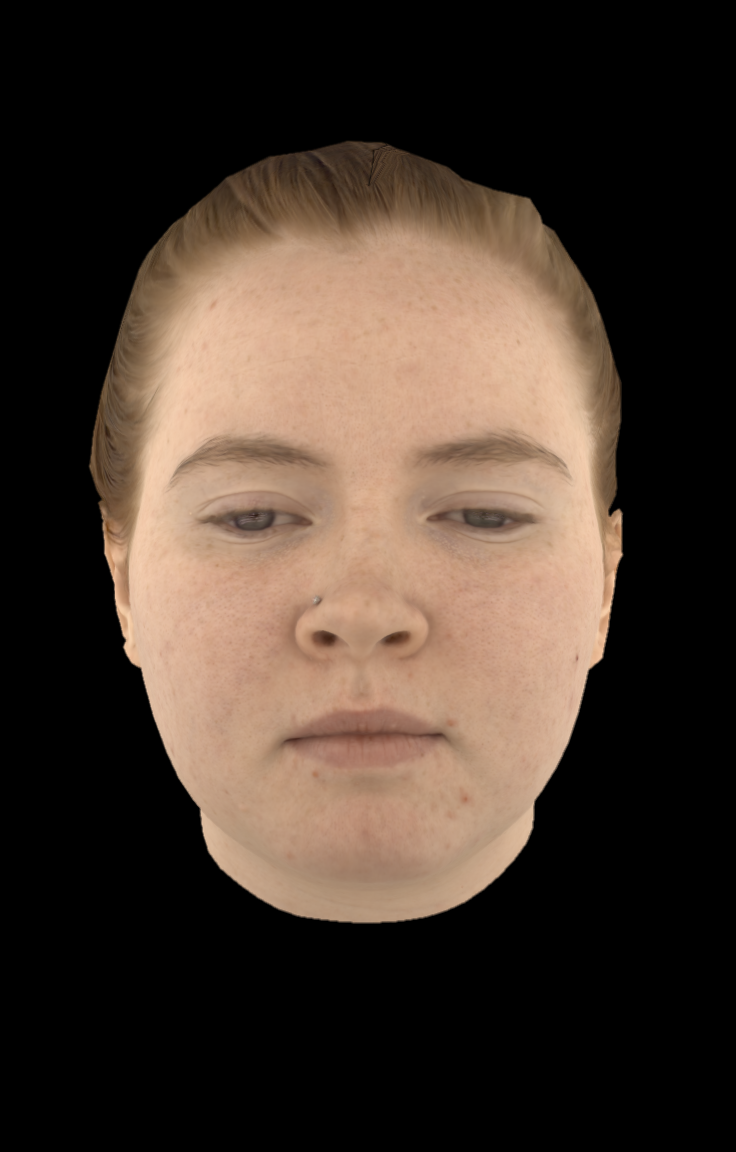} &
 \includegraphics[trim=90 100 100 100,clip, width=15mm]{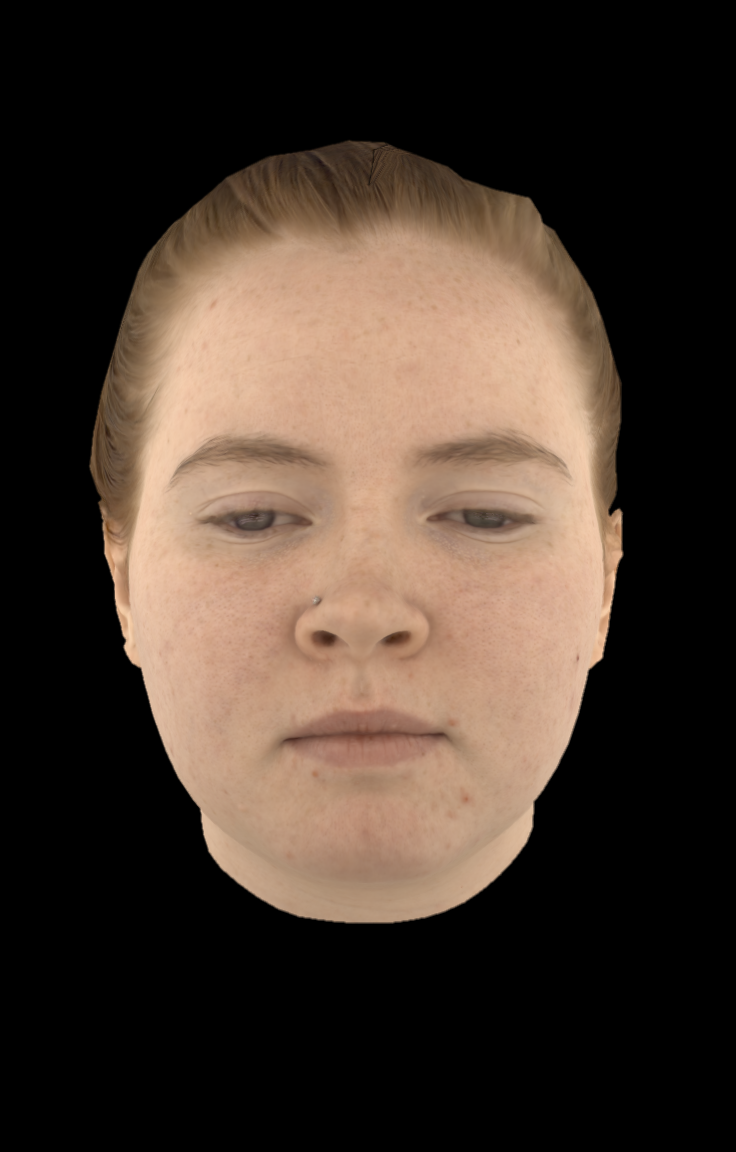} &
 \includegraphics[trim=90 100 100 100,clip, width=15mm]{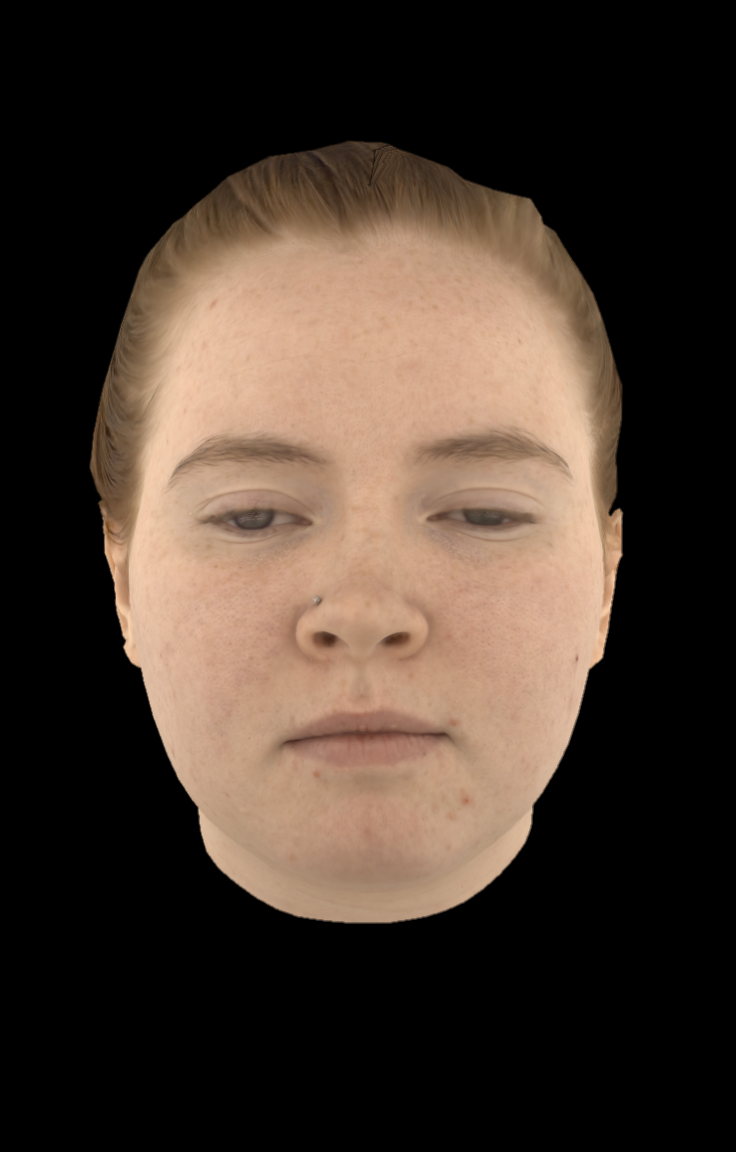} & \ \ \ \ &
 \includegraphics[width=17.5mm]{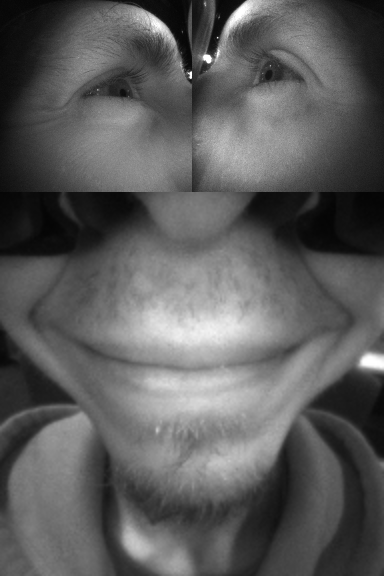} &
 \includegraphics[trim=90 100 100 100,clip, width=15mm]{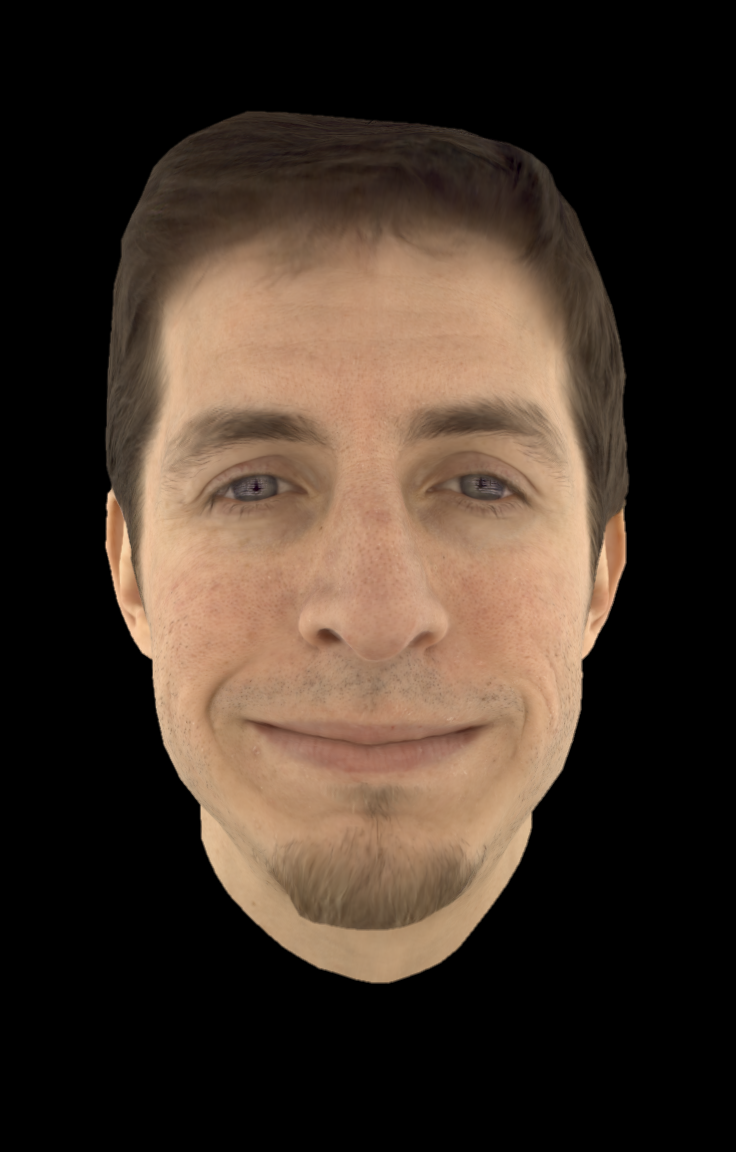} &
 \includegraphics[trim=90 100 100 100,clip, width=15mm]{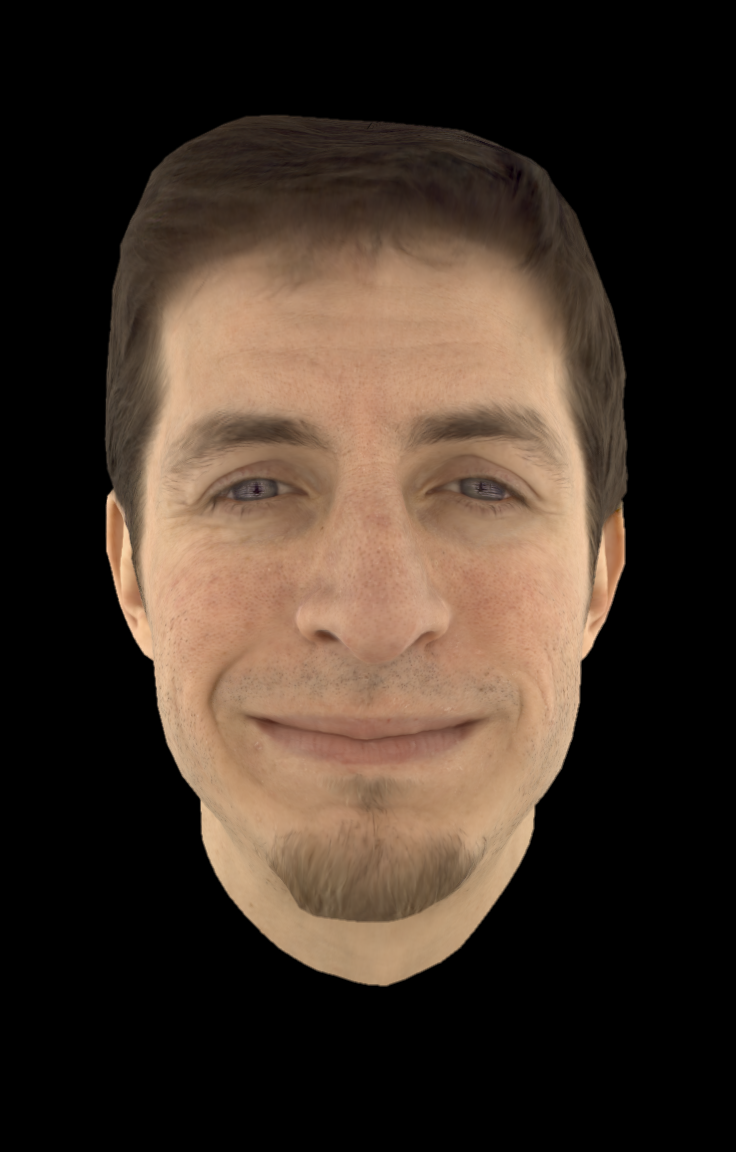} &
 \includegraphics[trim=90 100 100 100,clip, width=15mm]{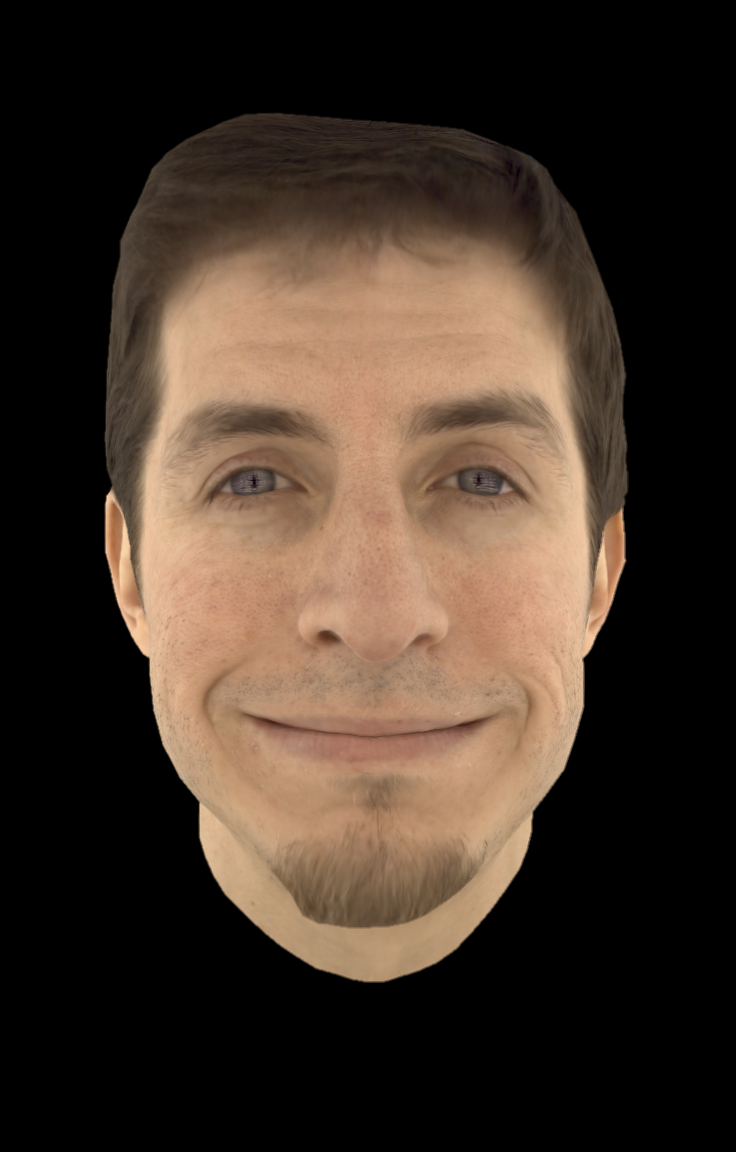}\\
 
 HMC & PS~\cite{wei2019vr} & MIA & GT & & HMC & PS~\cite{wei2019vr} & MIA & GT \\
\end{tabular}
\end{center}
\figvspaceB
\caption{The comparison of PS~\cite{wei2019vr} and MIA methods for animating the codec avatar from HMC images. The MIA can estimate more expressive and accurate expressions.}
\label{fig:test_final_images}
\figvspace
\end{figure*}

\vspace{-2mm}
\Section{Conclusion and Future Directions}

This paper proposes MIA to robustify and generalize existing PS methods for driving CAs. 
MIA learns to extract identity invariant features related to facial expression while marginalizing nuisance factors (headset, environment, facial expression) in an unsupervised manner. We show that MIA is able to drive the shape component in untrained subjects, and if the PS texture decoder is available, with a minimum training, MIA can drive CAs for new subjects. 
For future directions, first, we will design new loss function based on the closeness of 3D surfaces to model the lips and eyes closure. Second, we will work on texture-conditional decoders to make the texture part of the method generalizable for new subjects without pre-trained decoders.

\Section{Acknowledgement}

We thank "Baris Gecer" for early implementation of the 3D augmentation approach in Section~\ref{sec:augmentation}, and help brainstorming ideas during his internship at Facebook in the summer of 2019.
By error, his name was first omitted from the CVPR 2022 publication of this work.

\clearpage

{\small
\bibliographystyle{ieee_fullname}
\bibliography{egbib}
}

\end{document}